\documentclass{article}

     \usepackage[final,nonatbib]{neurips_2022}

\usepackage[utf8]{inputenc} 
\usepackage[T1]{fontenc}    
\usepackage{hyperref}       
\usepackage{cite}
\usepackage{url}            
\usepackage{booktabs}       
\usepackage{amsfonts}       
\usepackage{nicefrac}       
\usepackage{microtype}      
\usepackage{xcolor}         

\usepackage{soul} 
\usepackage{color, xcolor} 
\usepackage{amsmath}
\usepackage{amssymb}
\usepackage{amsthm}

\DeclareMathOperator*{\argmin}{arg\,min}

\newtheorem{theorem}{Theorem}
\newtheorem{proposition}{Proposition}
\newtheorem{definition}{Definition}

\newenvironment{proofsketch}{%
  \proof}{\endproof}

\usepackage{caption}
\usepackage{graphicx,subcaption}
\usepackage{enumitem, kantlipsum}
\usepackage{wrapfig}
\usepackage{array}
\usepackage{tabularx}
\title{Unsupervised Learning for Combinatorial Optimization with Principled Objective Relaxation}

\author{
  Haoyu Wang \\
  Purdue University \\
  \texttt{wang5272@purdue.edu} \\
   \And
  Nan Wu\\
  UCSB \\
  \texttt{nanwu@ucsb.edu} \\
  \And
  Hang Yang\\
  Georgia Tech. \\
  \texttt{hyang628@gatech.edu} \\
  \And 
  Cong Hao \\
   Georgia Tech. \\
  \texttt{callie.hao@gatech.edu} \\
  \And
  Pan Li\\ 
  Purdue University\\
  \texttt{panli@purdue.edu} \\
}

\begin{document}

\maketitle

\sethlcolor{pink}
\vspace{-3mm}
\begin{abstract}
\vspace{-1mm}
  Using machine learning to solve combinatorial optimization (CO) problems is challenging, especially when the data is unlabeled. This work proposes an unsupervised learning framework for CO problems. Our framework follows a standard relaxation-plus-rounding approach and adopts neural networks to parameterize the relaxed solutions so that simple back-propagation can train the model end-to-end. Our key contribution is the observation that if the relaxed objective satisfies entry-wise concavity, a low optimization loss guarantees the quality of the final integral solutions. This observation significantly broadens the applicability of the previous framework inspired by Erd\H{o}s' probabilistic method~\cite{karalias2020erdos}. In particular, this observation can guide the design of objective models in  applications where the objectives are not given explicitly  while requiring being modeled in prior. We evaluate our framework by solving a synthetic graph optimization problem, and two real-world applications including resource allocation in circuit design and approximate computing. Our framework\footnote{Our code and the datasets are available at: \url{https://github.com/Graph-COM/CO_ProxyDesign}} largely outperforms the baselines based on na\"{i}ve relaxation, reinforcement learning, and Gumbel-softmax tricks.         

\end{abstract}
\vspace{-2mm}
\section{Introduction}
\vspace{-1mm}
Combinatorial optimization (CO) with the goal of finding the optimal solution from a discrete space is a fundamental problem in many scientific and engineering applications~\cite{papadimitriou1998combinatorial,naseri2020application,crama1997combinatorial}. Most CO problems are NP-complete. Traditional methods efficient in practice often either depend on heuristics or produce approximation solutions. Designing these approaches requires considerable insights into the problem. Recently, machine learning has paved a new way to develop CO algorithms, which asks to use neural networks (NNs) to extract heuristics from the data~\cite{hopfield1985neural,smith1999neural,vinyals2015pointer}. Several learning for CO (LCO) approaches have therefore been developed, providing solvers for the problems including SAT ~\cite{selsam2018learning,amizadeh2018learning,yolcu2019learning}, mixed integer linear programming~\cite{khalil2016learning,gasse2019exact,delarue2020reinforcement}, vertex covering~\cite{khalil2017learning,li2018combinatorial} and routing problems~\cite{bello2016neural,chen2019learning,kool2018attention,joshi2019efficient,hudson2021graph,ma2021learning,kwon2021matrix,kim2021learning}.    


Another promising way to use machine learning techniques is to learn proxies of the CO objectives whose evaluation could be expensive and time-consuming~\cite{wu2022highlevel,mendis2019ithemal,zhang2019circuit,vasudevan2021learning}. For example, to optimize hardware/system design, evaluating the objectives such as computation latency, power efficiency~\cite{mishra2018caloree},  and resource consumption~\cite{renda2020difftune,wu2021ironman,mirhoseini2021graph} requires running complex simulators. Also, in molecule design, evaluating the objective properties such as protein fluorescence or DNA binding affinity needs either costly simulations or living experiments~\cite{chen1991enzyme,angermueller2019model,gomez2018automatic}. In these cases, proxies of the objectives can be learned first to reduce the evaluation cost~\cite{mirhoseini2021graph}, and then optimize these proxies to solve the design problems. Later, we name the CO problem with proxy objectives as Proxy-based CO (PCO) problems. Moreover, we claim that learning for PCO is often of greater significance than learning for traditional CO problems because commercial CO solvers such as Gurobi cannot applied in PCO due to the missing closed-form objectives and the lack of heuristics for such proxy objectives.  Generic solvers such as simulated annealing~\cite{bertsimas1993simulated} may be applied while they are extremely slow. 


\begin{figure}[t]
    \centering
    \includegraphics[width = \textwidth]{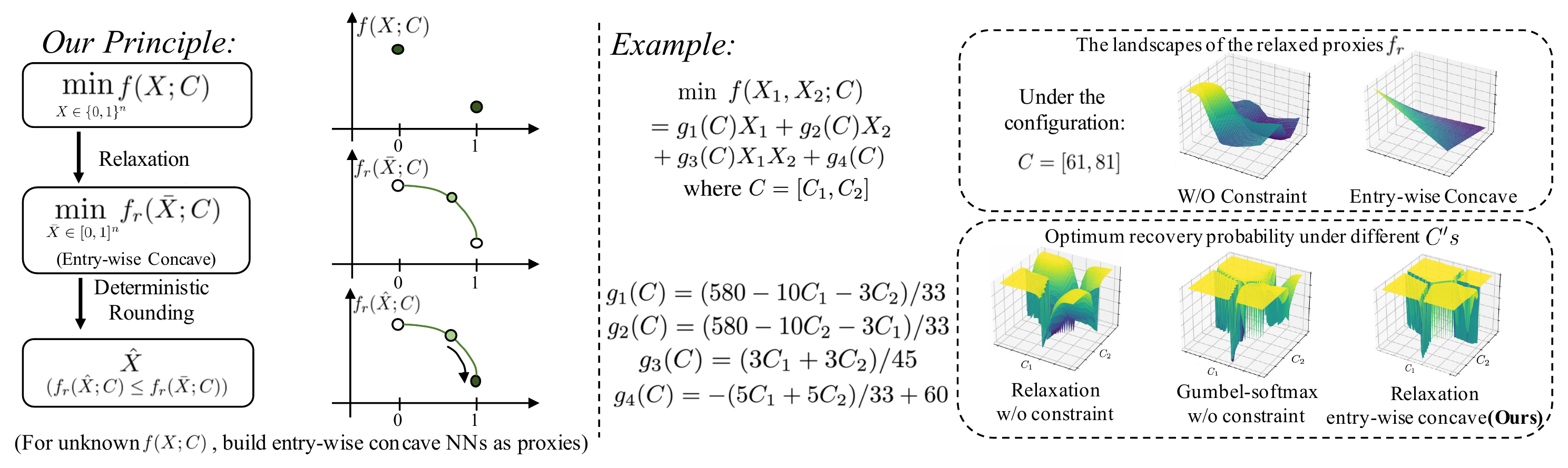}
    \vspace{-5mm}
    \small{\caption{The entry-wise concave relaxation-and-rounding principle (for the case without constraints as an illustration) and an example. Consider an optimization objective $f(X;C)$, $X \in \{0,1\}^n$. Here, $C$ is the problem configuration such as an attributed graph in a graph optimization problem. We relax the objective to an \emph{entry-wise concave} $f_r(\bar{X};C)$, $\bar{X} \in [0,1]^n$. The soft solution by minimizing $f_r(\bar{X};C)$ will be rounded to a discrete solution (in Def.~\ref{def:round}) with performance guarantee. When $f(X;C)$ is not explicitly given, we will learn its relaxed proxies $f_r$ by a NN. This corresponds to a PCO problem. In the toy example on the right, we first learn different relaxed proxies $f_r$'s with or without the entry-wise concave constraint. We compare their landscapes in the top-right figure. We further optimize both proxies and round the obtained soft solutions to integral solutions. The bottom-right figure shows the optimum-recovery probabilities of different methods under different $C$'s. 
    }
    \label{fig:toy_example}}
    \vspace{-3mm}
\end{figure}

In this work, we propose an unsupervised LCO framework. Our findings are applied to general CO problems while exhibiting extraordinary promise for PCO problems. Unsupervised LCO has recently attracted great attentions~\cite{amizadeh2018learning,toenshoff2019run,yao2019experimental,karalias2020erdos,hudson2021graph} due to its several great advantages. Traditional supervised learning often gets criticized for the dependence on huge amounts of labeled data~\cite{yehuda2020s}. Reinforcement learning (RL) on the other hand suffers from notoriously unstable training~\cite{mnih2015human}. In contrast, unsupervised LCO is superior in its faster training, good generalization, and strong capability of dealing with large-scale problems~\cite{karalias2020erdos}. Moreover, 
unsupervised learning has never been systematically investigated for PCO problems. Previous works for PCO problems, e.g., hardware design~\cite{mirhoseini2021graph,wu2021ironman}, were all based on RL. A systematic unsupervised learning framework for PCO problems is still under development.   

\begin{figure}[t]
    \centering
    \includegraphics[width = \textwidth]{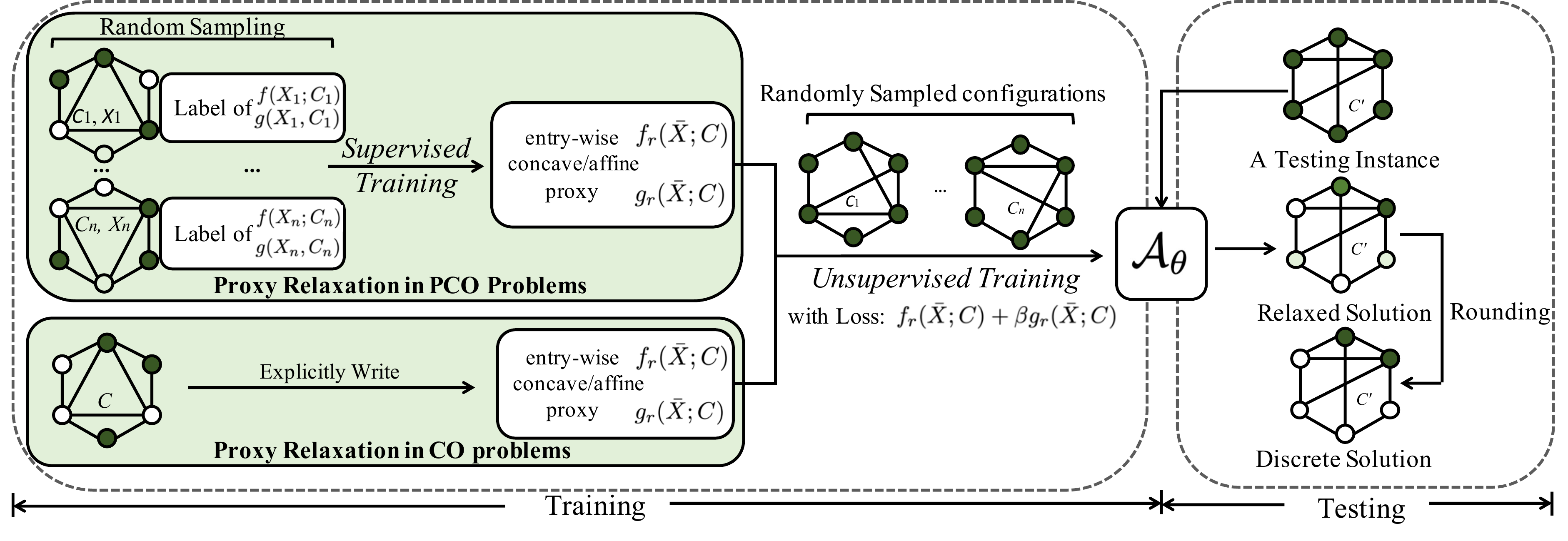}
    \vspace{-5mm}
    \small{\caption{The pipeline of empirical evaluation in Sec.~\ref{sec:eva1} and Sec.~\ref{sec:proxy} on our relaxation-and-rounding principle. For a PCO task whose cost function or constraints are unknown, we first build NNs with the entry-wise concave (CON) / affine (AFF) structure to learn as their proxies ($f_r(\bar{X};C),g_r(\bar{X};C)$) via supervised learning. For traditional CO tasks, we follow our principle and explicitly write their entry-wise affine costs and constraints relaxation. Then, we learn the algorithm $\mathcal{A}_{\theta}$ to optimize the relaxed loss function $f_r(\bar{X};C)+\beta g_r(\bar{X};C)$ in an unsupervised manner. After training, with any unseen testing instance $C'$, we run our $\mathcal{A}_{\theta}$ to infer the relaxed soft solution $\bar{X}$, and then round the soft solution into discrete solution $X$ with performance guarantee. }
    \label{fig:pipeline}}
    \vspace{-3mm}
\end{figure}

Our framework follows a relaxation-plus-rounding approach. We optimize a carefully-designed continuous relaxation of the cost model (penalized with constraints if any) and obtain a soft solution. Then, we decode the soft assignment to have the final discrete solution. This follows a common approach to design an approximation algorithm~\cite{gandhi2006dependent,byrka2013steiner}. However, the soft assignment here is given by a NN model optimized based on the historical (unlabeled) data via gradient descent. Note that learning from historical data is expected to facilitate the model understanding the data distribution, which helps with extracting heuristics, avoiding local minima and achieving fast inference. An illustration of our principle with a toy-example is shown in Fig.~\ref{fig:toy_example}. We also provide the  pipeline to empirically evaluate our relaxation-plus-rounding principle in Fig.~\ref{fig:pipeline}.

Our method shares a similar spirit with~\cite{karalias2020erdos} while making the following significant contributions. We abandon the probabilistic guarantee in~\cite{karalias2020erdos}, because it is hard to use for general CO objectives, especially those based on proxies. Instead, we design a deterministic objective relaxation principle that gives performance guarantee. We prove that if the objective relaxation is entry-wise concave w.r.t. the binary optimization variables, a low-cost soft solution plus deterministic sequential decoding guarantees generating a valid and low-cost integral solution. 
This principle significantly broadens the applicability of this unsupervised learning framework. In particular, it guides the design of model architectures to learn the objectives in PCO problems. We also justify the wide applicability of the entry-wise concave principle in both theory and practice.


We evaluate our framework over three PCO applications including feature-based edge covering \& node matching problems, and two real-world applications, including imprecise functional unit assignment in approximate computing (AxC)~\cite{han2013approximate, mittal2016survey, venkatesan2011macaco, ma2021workload, li2015joint} and resource allocation in circuit design~\cite{wu2021ironman, wu2022highlevel}. In all three applications, our framework achieves  a significant performance boost compared to previous RL-based approaches and relaxed gradient approaches based on the Gumbel-softmax trick~\cite{bengio2013estimating,jang2016categorical,maddison2016concrete}. The datasets for these applications are also provided in our github repository. 

\vspace{-2mm}
\subsection{Further Discussion on Related Works}
\vspace{-1mm}
Most previous LCO approaches are based on RL~\cite{bello2016neural,chen2019learning,kwon2020pomo,kwon2021matrix,khalil2017learning,delarue2020reinforcement,wang2021bi,kool2018attention,nandwani2021neural} or supervised learning~\cite{khalil2016learning,gasse2019exact,yehuda2020s}, as these two frameworks do not hold any special requirements on the formulation of CO problems. However, they often suffer from the issues of training instability and subpar generalization. Previous works on unsupervised learning for CO have studied satisfaction problems~\cite{amizadeh2018learning,toenshoff2019run,duan2022augment}. 
Applying these approaches to general CO problems requires problem reductions. Others have considered max-cut~\cite{yao2019experimental} and TSP problems~\cite{hudson2021graph}, while these works depend on carefully selected problem-specific objectives. The work most relevant to ours is~\cite{karalias2020erdos} and we give detailed comparison in Sec.~\ref{sec:method}. 
Note that all previous works on unsupervised learning for CO do not apply to PCO as they need an explicit objective to manipulate. For PCO problems, previous studies focus on how to learn more generalizable proxies of the objectives, such as via Bayesian learning~\cite{kumar2020model,brookes2019conditioning} and adverserial training~\cite{trabucco2021conservative,kumar2021data}. Once proxies are learned, direct objective relaxation plus gradient descent~\cite{trabucco2021conservative}, or RL~\cite{mirhoseini2021graph,wu2021ironman}, or  general MIP solvers with reformulations of the objectives~\cite{anderson2020strong,papalexopoulos2022constrained} have been often adopted. Studying  generalization of proxies is out of the scope of this work while we conjecture that our suggested entry-wise concave proxies seem smoother than those without constraints (See Fig.~\ref{fig:toy_example}) and thus have the potential to achieve better generalization.     

\vspace{-2mm}
\section{Preliminaries and Problem Formulation} \label{sec:prelim}
\vspace{-2mm}
In this section, we define several useful concepts and notations.

\textbf{Combinatorial Optimization (CO).} 
\label{def_CO}
Let $C\in \mathcal{C}$ denote a data-based configuration such as a graph with weighted edges. Let $\Omega$ be a finite set of all feasible combinatorial objects and each object has a binary vector embedding $X=(X_i)_{1\leq i\leq n}\in\{0,1\}^n$. For example, in the node matching problem, each entry of $X$ corresponds to an edge to denote whether this edge is selected or not. Note that such binary embeddings are applicable even when the choice is not naturally binary: Choosing at most one element from a tuple $(1,2,3)$ can be represented as a 3-dim binary vector $(X_1, X_2, X_3)$ with the constraint $X_1+X_2+X_3\leq 1$. W.l.o.g, we assume an algebraic form of the feasible set $\Omega\triangleq\{X\in \{0,1\}^n: g(X;C)< 1\}$ where $g(X;C)\geq 0$ for all $X\in \{0,1\}^n$ \footnote{Normalization $(g(\cdot;C) - g_{\min})/(g_{\min}^+ - g_{\min})$ where $g_{\min}^+ = \min_{X\in\{0,1\}^n\backslash \Omega} g(X;C)$ and $g_{\min} = \min_{X\in\{0,1\}^n} g(X;C)$ always satisfies the property. 
$g_{\min}^+,\,g_{\min}$ often can be easily estimated in practice.
}. 
For notational simplicity, we only consider one inequality constraint while our later discussion in Sec.~\ref{sec:method} and our case studies in Sec.~\ref{sec:proxy} may contain multiple inequalities. Given a configuration $C$ and a constraint $\Omega$, a combinatorial optimization (CO) is to minimize a cost $f(\cdot;C)$ by solving 
\begin{align} \label{eq:PCO}
    \min_{X\in\{0,1\}^n} f(X;C), \quad \text{s.t.}\quad g(X;C) < 1.
\end{align}


\textbf{Proxy-based CO (PCO).} In the many applications, the cost or the constraint may not be cheaply evaluated. 
Some proxies of the cost $f$ or the constraint $g$ often need to be learned from the historical data. 
With some abuse of notations, we interchangably use $f$ ($g$, resp.) to denote the objective (the constraint, resp.) and its proxy.

\textbf{Learning for CO/PCO (LCO).} A LCO problem is to learn an algorithm $\mathcal{A}_{\theta}(\cdot):\mathcal{C}\rightarrow \{0,1\}^n$, say a neural network (NN) parameterized by $\theta$ to solve CO or PCO problems. Given a configuration $C\in\mathcal{C}$, we expect $\mathcal{A}_{\theta}$ to (a) generate a valid solution $\hat{X}= \mathcal{A}_{\theta}(C)\in\Omega$ and (b) minimize $f(\hat{X}; C)$. 

There are different approaches to learn $\mathcal{A}_{\theta}$. Our focus is  unsupervised learning approaches where given a configuration $C$, \emph{no ground-truth solution $X^*$ is used during the training.} $\theta$ can only be optimized just based on the knowledge of the cost and the constraint, or their proxies. Note that in this work, for PCO problems, the proxies are assumed to be trained based on ground-truth values of $f(X;C)$ given different $(X,C)$ pairs, which is supervised. Unsupervised learning in this work refers to the way to learn $\mathcal{A}_{\theta}(C)$.


\textbf{Erd\H{o}s' Probabilistic Method (EPM).} The EPM has recently been brought for LCO~\cite{karalias2020erdos}. Specifically, The EPM formulates $\mathcal{A}_{\theta}(C)$ as a randomized algorithm that essentially gives a probabilistic distribution over the solution space $ \{0,1\}^n$, which solves the optimization problem: 
\begin{align}
    \min_{\theta}\quad \mathbb{E}\left[l(X;C)\right] ,\;\text{where}\; l(X;C)\triangleq f(X; C) + \beta 1_{g(X;C)\geq 1},\,X\sim \mathcal{A}_{\theta}(C)\;\text{and $\beta>0$ .} \label{eq:EPM} 
\end{align}
Karalias \& Loukas proved that with $\beta > \max_{X\in\Omega} f(X;C)$ and a small expected loss $\mathbb{E}\left[l(X,C)\right] < \beta$, sampling a sufficiently large number of $\hat{X}\sim \mathcal{A}_{\theta}(C)$ guarantees the existance of a feasible $\hat{X}\in\Omega$ that achieves the cost $f(\hat{X};C)\leq \mathbb{E}\left[l(X,C)\right]$~\cite{karalias2020erdos}. Although this guarantee makes EPM intriguing, applying EPM in practice is non-trivial. We will explain the challenge in Sec.~\ref{sec:mot}, which inspires our solutions and further guides the objective design for general CO and PCO problems. 

\vspace{-1mm}
\section{The Relaxation Principle for Unsupervised LCO} \label{sec:method}
\vspace{-2mm}

In this section, we start with the practical issues of EPM. Then, we introduce our solutions by proposing a relaxation principle of the objectives, which gives performance guarantee for general practical unsupervised LCO. 
\vspace{-1mm}
\subsection{Motivation: The Practical Issues of EPM} \label{sec:mot}
\vspace{-1mm}
Applying EPM in practice has two fundamental difficulties. First, optimizing $\theta$ in Eq.\eqref{eq:EPM} is generally hard as the gradient $\frac{d X}{d \theta}$ (note here each entry of $X$ is binary) may not exist so the chain rule cannot be used. We discuss the potential solutions to this problem in Sec.~\ref{sec:others}. Second, EPM needs to sample a large number of $X\sim \mathcal{A}_{\theta}(C)$ for evaluation to achieve the performance guarantee in~\cite{karalias2020erdos}. This is not acceptable where the evaluation per sample is time-consuming and expensive. 

So, in practice, Karalias \& Loukas consider a deterministic method. They view $\mathcal{A}_{\theta}(C)\in[0,1]^n$ as the parameters of Bernouli distributions to generate the entries of $X$ so $\mathbb{E}[X]=\mathcal{A}_{\theta}(C)$. First, they optimize $\min_{\theta} l(\mathcal{A}_{\theta}(C),C)$ instead of $\min_{\theta} \mathbb{E}[l(X,C)]$, and then, sequentially round the probability $\mathcal{A}_{\theta}(C)$ to discrete $X\in\{0,1\}^n$ by comparing conditional expectations, e.g., $\mathbb{E}[l(X,C)|X_1=0]$ v.s. $\mathbb{E}[l(X,C)|X_1=1]$ to decide $X_1$. However, such conditional expectations cannot be efficiently computed unless one uses Monte-Carlo sampling or $l$ has special structures as used in the two case studies in~\cite{karalias2020erdos}, i.e., max-clique and graph-partition problems. However, what special structures are needed has not been defined,  
which blocks the applicability of this framework to general CO problems, especially for the PCO problems where the objectives $l$ are learned as models.

\vspace{-1mm}
\subsection{Our Approach: Relaxation plus Rounding, and Performance Guarantee}
\vspace{-1mm}
Our solution does not use the probabilistic modeling but directly adopts a relaxation-plus-rounding approach. 
We optimize a relaxation of the objective $l_r$ and obtain a soft solution $\bar{X}\in [0,1]^n$. Then, we deterministically round the entries in $\bar{X}$ to a solution $X$ in the discrete space $\{0,1\}^n$. Note that throughout the paper, we use $\bar{X}$ to denote a soft solution and $X$ to denote a discrete solution. The question is whether the obtained solution may still achieve the guarantee as EPM does. Our key observation is that such success essentially depends on how to relax the objective $l$. 



Therefore, our first contribution beyond~\cite{karalias2020erdos} is to propose the principle (Def.~\ref{def:cond}) to relax general costs and constraints. With this principle, the unsupervised LCO framework can deterministically yield valid and low-cost solutions (Thm.~\ref{thm:main}) as the EPM guarantees, and is applied to any objective $l$. 

First, we introduce the pipeline. Consider a relaxation of a deterministic upper bound of Eq.\eqref{eq:EPM}:
\begin{align}\label{eq:relax}
        \min_{\theta}\; l_{r}(\theta;C) \triangleq  f_r(\bar{X}; C) + \beta g_r(\bar{X};C) ,\, \text{where}\;\bar{X} = \mathcal{A}_{\theta}(C)\in [0,1]^n\;\text{, $\beta>0$.}
\end{align}
Here $f_r(\cdot; C):[0,1]^n\rightarrow \mathbb{R}$ is the relaxation of $f(\cdot; C)$, which satisfies $f_r(X;C)=f(X;C)$ for $X\in\{0,1\}^n$. The relation between the constraint $g$ and its relaxation $g_r$ is similar, i.e., $g_r(X;C)=g(X;C)$ for $X\in\{0,1\}^n$. Here, we also use the fact that $g_r(X;C)$ provides a natural upper bound $1_{g(X;C) \geq 1}\leq g_r(X;C)$ for $X\in\{0,1\}^n$ given the normalization of $g(X;C)$ adopted in Sec.~\ref{sec:prelim}. 

Now, suppose the parameter $\theta$ gets optimized so that $l_{r}(\theta;C)$ is small. Further, we adopt the sequential rounding in Def.~\ref{def:round} to adjust the continuous solution $\bar{X}=\mathcal{A}_{\theta}(C)$ to discrete solution $X$. 

\begin{definition}[Rounding] \label{def:round} Given a continuous vector $\bar{X} \in [0,1]^n$ and an arbitrary order of the entries, w.o.l.g., $i=1,2,...,n$, round $\bar{X}_i$ into $0$ or $1$ and fix all the other variables un-changed. Set  $X_i = \argmin_{j=0,1} f_r(X_1,...,X_{i-1},j, \bar{X}_{i+1},...,\bar{X}_n;C) + \beta g_r(X_1,...,X_{i-1},j, \bar{X}_{i+1},...,\bar{X}_n;C)$, replace $\bar{X}_i$ with $X_i$ and repeat the above procedure until all the variables become discrete.
\end{definition}

Note that our rounding procedure does not need to evaluate any conditional expectations $\mathbb{E}[l(X;C)|X_1]$ which EPM in~\cite{karalias2020erdos} requires. Instead, we ask both relaxations $f_r$ and $g_r$ to satisfy the principle in Def.~\ref{def:cond}. With this principle, the pipeline allows achieving a valid and low-cost solution $X$, as proved in Theorem~\ref{thm:main}. We leave the proof in Appendix~\ref{sec:proof-main}.


\begin{definition}[The Entry-wise Concave Principle]\label{def:cond} For any $C\in\mathcal{C}$, $h_r(\cdot;C):[0,1]^n\rightarrow \mathbb{R}$ is entry-wise concave if for any $\gamma\in[0,1]$ and any $\bar{X}, \bar{X}'\in[0,1]^n$ that are only different in one entry,
\begin{align*}
    \gamma h_r(\bar{X};C) + (1-\gamma) h_r(\bar{X}';C) \leq  h_r(\gamma\bar{X}+(1-\gamma)\bar{X}';C). 
\end{align*}
\end{definition}\vspace{-1mm}

Note that entry-wise concavity is much weaker than concavity. For example, the function $h_r(\bar{X}_1,\bar{X}_2)=-\text{Relu}(\bar{X}_1\bar{X}_2)$, $\bar{X}_1,\bar{X}_2\in \mathbb{R}$ is entry-wise concave but not concave.

\begin{theorem}[Performance Guarantee] 
\label{thm:main}Let $\beta > \max_{X\in\Omega} f(X;C)$ and $\min_{X\in\Omega}f(X;C)\geq0$ in Eq.\eqref{eq:relax}. Suppose the relaxed cost $f_r$ and constraint $g_r$ are entry-wise concave, and the learned parameter $\theta$ achieves $l_r(\theta;C)< \beta$. Then, rounding (Def.~\ref{def:round}) the relaxed solution $\bar{X}=\mathcal{A}_{\theta}(C)$ generates a valid discrete solution $X\in\Omega$ such that $f(X;C)<l_r(\theta;C)$.
\end{theorem}



When there are multiple constraints $g^{(j)}(X;C) < 1$ for $j=1,2,...$, we may use relaxation $\beta\sum_j g_r^{(j)}(X;C)$ as the penalty term in Eq.\eqref{eq:relax}, where $g_r^{(j)}$ is a relaxation of $g^{(j)}$. It can be shown that if $\sum_j g_r^{(j)}$ satisfies the condition of entry-wise concavity, the guarantee of Thm.~\ref{thm:main} still applies. 

\vspace{-1mm}
\subsection{The Wide Applicability of Entry-wise Concave Relaxations}
\vspace{-1mm}
We have introduced the entry-wise concave principle to relax the objective to associate our framework with performance guarantee. The question is how widely applicable this principle could be. 

Actually, every function with binary inputs can be relaxed as an entry-wise affine (also called multi-linear) function with the exactly same values at the discrete inputs, as shown in  Theorem~\ref{thm:app}. Note that entry-wise affinity is a special case of entry-wise concavity. In Sec.~\ref{sec:proxy}, we will provide the design of NN architecture (for PCO) and math derivation (for CO) that guarantee formulating an entry-wise concave function. Note that the objectives for max-clique and graph-partition problems used in~\cite{karalias2020erdos} are essentially entry-wise affine.

\begin{theorem}[Wide Applicability] \label{thm:app}
For any binary-input function $h(\cdot): \{0,1\}^n\rightarrow \mathbb{R},$ there exists a  relaxation $h_r(\cdot): [0,1]^n\rightarrow \mathbb{R}$ such that (a) $h_r(X)=h(X)$ for $X\in \{0,1\}^n$ and (b) $h_r$ is entry-wise affine, i.e., for any $\gamma\in[0,1]$ and any $\bar{X}, \bar{X}'\in[0,1]^n$ that are only different in one entry,
\begin{align*}
    \gamma h_r(\bar{X}) + (1-\gamma) h_r(\bar{X}') =  h_r(\gamma\bar{X}+(1-\gamma)\bar{X}'). 
\end{align*} 
\end{theorem}
\begin{proofsketch}
Set $h_r(\bar{X}) = \sum_{X\in \{0,1\}^n} h(X) \prod_{j=1}^n \bar{X}_j^{X_j}(1-\bar{X}_j)^{(1-X_j)}$, which satisfies (a) and (b). Note that we suppose that $\bar{X}_j^0=1$ for any $\bar{X}_j\in[0,1]$. The detailed proof is in Appendix~\ref{sec:proof_thm:app}.
\end{proofsketch}
\vspace{-2mm}
Although Theorem~\ref{thm:app} shows the existence of entry-wise affine relaxations, the constructed representation in the proof depends on higher-order moments of the input entries, which make it hard to be implemented by a model, say a NN architecture. However, we claim that  using entry-wise concave functions is able to implicitly generate higher-order moments via representations based on low-order moments. 
For example, when $n=2$, we could use the composition of $-\text{Relu}(\cdot)$ and affine operators (only 1st-order moments) to achieve universal representation (See Prop.~\ref{proposition_concave} and the proof in Appendix~\ref{sec:proof-prop}).  For general $n$, we leave as a future study. 

\begin{proposition}
\label{proposition_concave}
For any binary-input function $h(X_1,X_2)$, there exists parameters $\{w_{ij}\}$ such that an entry-wise concave function $h_r(\bar{X}_1, \bar{X}_2) = w_{00} - \sum_{i=1}^3 \text{Relu}(w_{i1}\bar{X}_1+w_{i2}\bar{X}_2+w_{i0})$ satisfies $h_r(X_1, X_2) = h(X_1, X_2)$ for any $X_1,X_2\in\{0,1\}$.
\end{proposition}

\vspace{-1mm}
\subsection{Discussion: Methods to Directly Optimize the Randomized Objective in EPM Eq.\eqref{eq:EPM}}\label{sec:others}
\vspace{-1mm}

The na\"{i}ve way to optimize the randomized objective in Eq.\eqref{eq:EPM} without worrying about the specific form of the objective $l$ is based on the policy gradient in RL  via the logarithmic trick, i.e., estimating the gradient $\frac{dl}{d \theta}$ via $(f(X;C)+\beta 1_{g(X;C)\geq 1})\log \mathbb{P}(X)$ by sampling $X\sim \mathcal{A}_{\theta}(C)$. However, the policy gradient suffers from notoriously large variance~\cite{mnih2015human} and makes RL hard to converge. Therefore, methods such as actor critic~\cite{konda1999actor} or subtracting some baselines $l(X;C)-b$~\cite{mnih2016asynchronous} have been proposed.

\begin{wraptable}{r}{0.51\textwidth}
\scriptsize
\centering
\vspace{-0.4cm}
\begin{tabular}{@{}cccc@{}}
\toprule
              & RL       & Gumbel-softmax       & Ours  \\ \midrule
Objective       & No Limit & No Limit       & Entry-wise Concave      \\
Optimizer      & Log Trick       & Gumbel Trick  & No Limit                \\
Inference      & Sampling & Sampling       & Deter. Rounding  \\
Train. Time  & Slow     & Fast           & Fast                    \\
Convergence & Hard  & Medium & Easy \\
Infer. Time & Slow     & Slow           & Fast                    \\ \bottomrule
\end{tabular}
\vspace{-0.2cm}
\small{\caption{The comparison among RL (policy gradient), Gumbel-softmax methods and our principled objective relaxation. Our methods are in need of much less training time and inference time. 
}
\label{tab:gumbel_linear}}
\vspace{-0.1cm}
\end{wraptable}

Another way to solve Eq.\eqref{eq:EPM} is based on reparameterization tricks to reduce the variance of gradients~\cite{paulus2020gradient,struminsky2021leveraging}. Specifically, we set the entries of output $\bar{X}=\mathcal{A}_{\theta}(C)\in[0,1]^n$ as the parameters of Bernoulli distributions to generate $X$, i.e., $X_i\sim \text{Bern}(\bar{X}_i)$, for $1\leq i\leq n$. To make $d X_i/d\bar{X}_i$ computable, we may use the Gumble-softmax trick~\cite{bengio2013estimating,jang2016categorical,maddison2016concrete}. However, this approach suffers from two issues. First, 
the estimation of the gradient is biased. Second, as $\mathcal{A}_{\theta}(C)$ is essentially a randomized algorithm, sampling sufficiently many $X\sim \mathcal{A}_{\theta}(C)$ is needed to guarantee a valid and low-cost solution. However, such evaluation is costly as discussed in Sec.~\ref{sec:mot}. So, empirically, we can also compare $\mathcal{A}_{\theta}(C)$ with a threshold to determine $X$, which does not have performance guarantee. 
We compare different aspects of RL, Gumbel-softmax tricks and our relaxation approach in Table~\ref{tab:gumbel_linear}.

\section{Applying Our Relaxation Principle to Learning for CO}\label{sec:eva1}
\begin{wraptable}{r}{0.5 \textwidth}
\vspace{-0.3cm}
\resizebox{0.5\textwidth}{!}{\begin{tabular}{@{}cll@{}}
\toprule
Method    & \multicolumn{1}{c}{Twitter}                    & \multicolumn{1}{c}{RBtest}                     \\ \midrule
Badloss+R & { 0.768 ± 0.203 (0.17s/g)} & { 0.702 ± 0.102 (0.33s/g)} \\
EPM~\cite{karalias2020erdos}       & { 0.924 ± 0.133 (0.17s/g)} & { 0.788 ± 0.065 (0.23s/g)} \\
AFF (ours)       & { 0.926 ± 0.113 (0.17s/g)} & { 0.787 ± 0.065 (0.33s/g)} \\ \bottomrule
\end{tabular}}
\caption{Approximation Rate in the max clique. `s/g' denotes the average time cost per graph.}
\label{tab:erdos_lco}
\vspace{-0.2cm}
\end{wraptable}

First, we test our relaxation principle in a learning for CO (max clique) task, where we can explicitly write both the cost functions and the constraints into an entry-wise affine form. In this case, our framework and EPM~\cite{karalias2020erdos} share the same pipeline, though the relaxation principle and the deterministic performance guarantee are firstly proposed in this work. The entry-wise affine objective relaxation of the max clique is $ - (\beta+1)\sum_{(i,j)\in E} \bar{X}_i \bar{X}_j + \frac{\beta}{2} \sum_{i \neq j} \bar{X}_i \bar{X}_j$.
Here, we use a real-world dataset (Twitter~\cite{leskovec2014snap}) and a synthetic dataset (RBtest~\cite{karalias2020erdos}), and show the experiment results in Table.~\ref{tab:erdos_lco}. We follow the settings in ~\cite{karalias2020erdos} and use a 6:2:2 dataset split for training, validating and testing, each test instance runs within $8$ seeds. Our entry-wise affine pipeline achieves almost the same performance as EPM. To show the importance of the relaxation principle, we also propose `Badloss+R'. This baseline imposes trigonometric functions to the original entry-wise affine functions $f_r'(\bar{X};C) = f_r(\sin (9 \pi \bar{X} / 2);C), g_r'(\bar{X};C) = g_r(\sin (9\pi \bar{X}/2);C)$, where  $\sin(\cdot)$ operates on each entry of the input vector. Note that the relaxed functions also match the  original objectives at discrete points, i.e., $f_r'(X;C)=f_r(X;C)=f(X;C)$ when $X\in\{0,1\}$, while with different relaxations.
The poor performance of ‘Badloss+R’ reveals the importance of our principle for relaxation.


\vspace{-2mm}
\section{Applying Our Relaxation Principle to Learning for PCO} \label{sec:proxy}
\vspace{-1mm}
In this section, we apply our relaxation principle to three PCO applications: (I) feature-based edge covering \& node matching, (II) resource allocation in circuit design, and (III) imprecise functional unit assignment in approximate computing. All the applications have graph-based configurations $C$. So later, we first introduce how to use graph neural networks (GNNs) to build proxies that satisfy our relaxation principle. Such GNN-based proxies will be used as the cost function relaxation $f_r$ in all the applications. Our principle can also guide the relaxation of explicit CO objectives. The constraints in applications (I)(III) are explicit and their relaxation can be written into the entry-wise affine form. The constraint in (II) needs another GNN-based entry-wise concave proxy to learn. 







\vspace{-2mm}
\subsection{GNN-based Entry-wise Concave Proxies}
\vspace{-1mm}
We consider the data configuration $C$ as an attributed graph $(V,E,Z)$ where $V$ is the node set, $E\subseteq V\times V$ is the edge set and $Z$ is the node attributes. We associate each node with a binary variable and group them together $X:\in\{0,1\}^{|V|}$. where for each $v\in V$, $X_v=1$ indicates the choice of the node $v$. Note that our approach can be similarly applied to edge-level variables (see Appendix~\ref{apd:edge-proxy}), which is used in application (I). Let $\bar{X}$ still denote the relaxation of $X$. 

To learn a discrete function $h:\{0,1\}^{|V|}\times \mathcal{C}\rightarrow \mathbb{R}$, we adopt a GNN as the relaxed proxy of $h$. We first define a latent graph representation in $\mathbb{R}^F$ whose entries are all entry-wise affine mappings of $X$.  
\begin{align}\label{eq:latent}
   \textbf{Latent representation:}\quad\quad\quad \phi(\bar{X};C) = W + \sum_{v\in V} U_{v} \bar{X}_1 + \sum_{v,u\in V, (v,u)\in E} Q_{v,u} \bar{X}_{v}\bar{X}_{u} \quad\quad
\end{align}
where $W$ is the graph representation, $U_{v}$'s are node representations and $Q_{v,u}$ are edge representations. These representations do not contain $X$ and are given by GNN encoding $C$. Here, we consider at most 2nd-order moments based on adjacent nodes as they can be easily implemented via current GNN platforms~\cite{FeyPyG,wang2019dgl}. Then, we use $\phi$ to generate entry-wise affine \& concave proxies as follows.
\begin{align}
   &\textbf{Entry-wise Affine Proxy (AFF):}\quad\quad\quad\quad &h_r^{\text{a}}(\bar{X};C) = \langle w^a, \phi(\bar{X};C)\rangle. \quad\quad\quad\quad\quad \label{eq:aff-proxy}\\
  &\textbf{Entry-wise Concave Proxy (CON):}\quad\quad\quad\quad &h_r^{\text{c}}(\bar{X};C) = \langle w^c, -\text{ReLU}(\phi(\bar{X};C))\rangle + b.  \label{eq:con-proxy} 
\end{align}
where $w^a,w^c\in\mathbb{R}^F, b\in\mathbb{R}$ are learnt parameters and $w^c\geq0$ guarantees entry-wise concavity. Other ways to implement GNN-based entry-wise concave proxies are also introduced in Appendix~\ref{apd:other-proxy}. 
\vspace{-0.1cm}

\subsection{The Setting up of the Experiments}
\begin{wraptable}{r}{0.35\textwidth}
\vspace{-0.2cm}
\centering
\scriptsize{
\begin{tabular}{@{}cccc@{}}
\toprule
Baseline  & $f_r$,\,$g_r$    & $\mathcal{A}_{\theta}$  & Inference \\ \midrule
Na\"{i}ve + R & no limit & no limit & rounding  \\
RL        & no limit & RL       & sampling  \\
GS-Tr+S   & no limit & GS       & sampling  \\
GS-Tr+R   & no limit & GS       & rounding  \\ \bottomrule
\end{tabular}}
\vspace{-0.2cm}
\small{\caption{The baselines in the paper.}
\label{tab:baseline}}
\vspace{-0.2cm}
\end{wraptable}

\textbf{Training \& Evaluation Pipeline.} In all the applications, we adopt the following training \& evaluation pipeline. First, we have a set of observed configurations $\mathcal{D}_1\subset\mathcal{C}$. Each $C\in\mathcal{D}_1$ is paired with one $X\in\{0,1\}^n$. We use the costs $f(X,C)$ (and constraints $g(X,C)$) to train the relaxed proxies $f_r(X,C)$ (and $g_r(X,C)$, if cannot be derived explicitly), where the relaxed proxies follow either Eq.\eqref{eq:aff-proxy} (named AFF) or Eq.\eqref{eq:con-proxy} (named CON). Then, we parameterize the LCO algorithm $\mathcal{A}_{\theta}(C)\in[0,1]^n$ via another GNN. Based on the learned (or derived) $f_r$ and $g_r$, we optimize $\theta$ by minimizing $\sum_{C\in\mathcal{D}_1}l_r(\theta;C)$, where $l_r$ is defined according to Eq.\eqref{eq:relax}. We will split $\mathcal{D}_1$ into a training set and a validation set for hyperparameter-tuning of both proxies and $\mathcal{A}_{\theta}$. We have another set of configurations $\mathcal{D}_2\subset\mathcal{C}$ used for testing. For each $C\in\mathcal{D}_2$, we use the relaxation  $\bar{X}=\mathcal{A}_{\theta}(C)$ plus our rounding to evaluate the learned algorithm $\mathcal{A}_{\theta}(\cdot)$. We follow~\cite{karalias2020erdos} and do not consider fine-tuning $\mathcal{A}_{\theta}$ over the testing dataset $\mathcal{D}_2$ to match the potential requirement of the fast inference.





\textbf{Baselines.} We consider $4$ common baselines that is made up of different learnable relaxed proxies $f_r,\,g_r$, algorithms $\mathcal{A}_{\theta}$ and inference approaches as shown in Table~\ref{tab:baseline}. For the proxies $f_r,\,g_r$ for baselines, we apply GNNs without the entry-wise concave constraint and use $X$ as one  node attribute while keeping all other hyper-parameters exactly the same as CON other than the way to deal with the discrete variables to make fair comparison (See details in Appendix.~\ref{apd:implementation_details}); For the algorithm $\mathcal{A}_{\theta}$, we provide the Gumbel-softmax trick based methods (GS-Tr)~\cite{maddison2016concrete,jang2016categorical}, the actor-critic-based RL method~\cite{konda1999actor} (RL) and the na\"{i}ve relaxation method (Na\"{i}ve); For the inference approaches, we consider Monte Carlo sampling (S) and our proposed rounding (R) procedure.
Although the baselines adopt proxies that are different from ours, we guarantee that their proxies approximate the ground-truth $f,g$ over the validation dataset at least no worse than ours. 
In application II, we also consider two non-learnable algorithms to optimize the proxies without relaxation constraints,  simulated annealing (SA)~\cite{bertsimas1993simulated} and genetic algorithms (GA)~\cite{mirjalili2019genetic,whitley1994genetic}. 
In application III, we put all of the required AxC units either close to the input (C-In) or close to the output (C-Out) of the approximating computing circuit as additional baselines. More details of the experiments setups and hyperparameter tuning can be found in Appendix~\ref{apd:implementation_details}. We also obtain the optimal solutions (OPT) for applications I and III via brute-force search for comparison.

\begin{figure*}[t]
\centering
\begin{subfigure}{0.235\textwidth}
   \includegraphics[width=\linewidth]{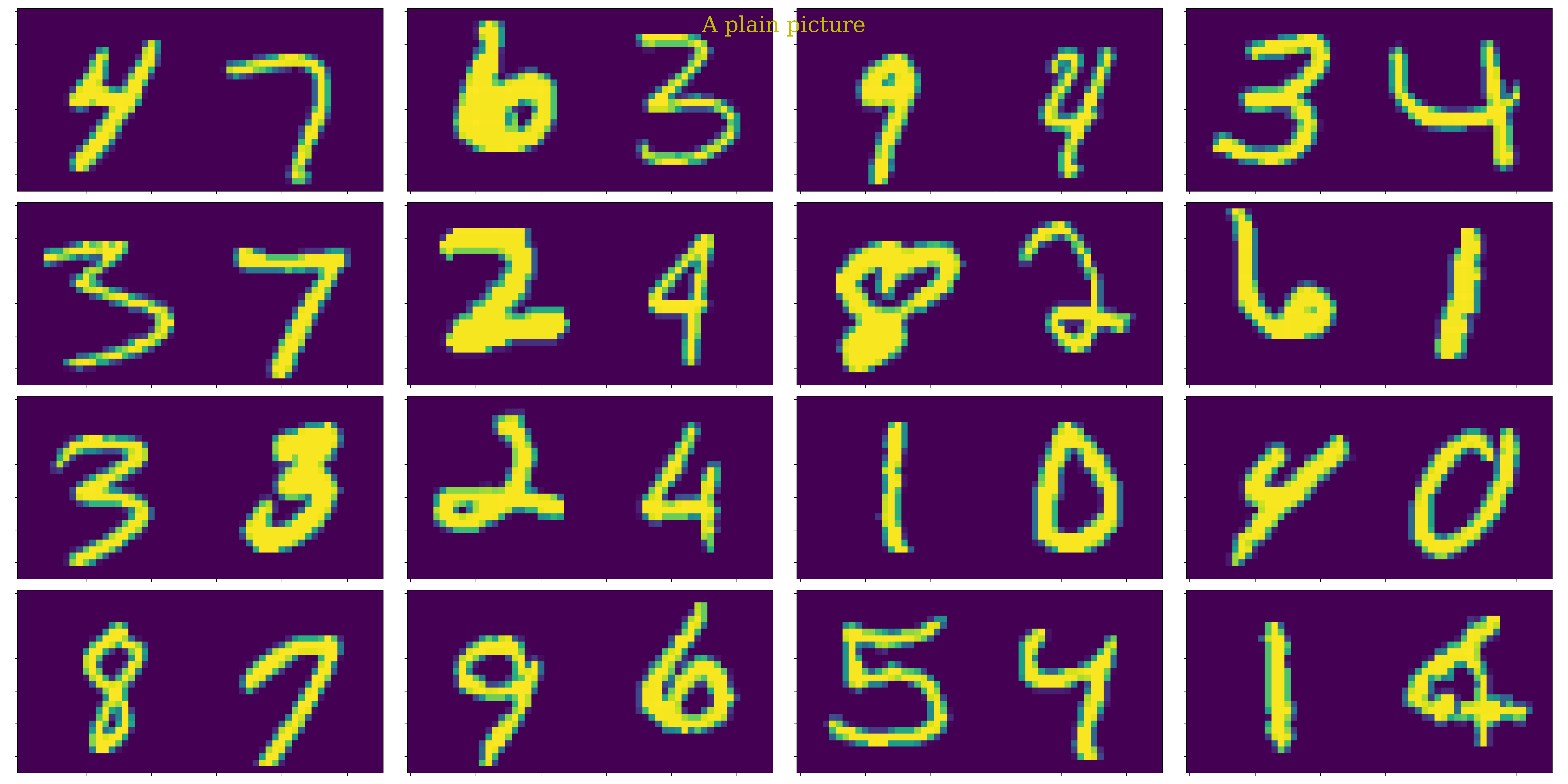}
   \vspace{-6mm}
   \caption{ One Config.} \label{fig:application_1_a}
\end{subfigure}
\hspace*{\fill}
\begin{subfigure}{0.235\textwidth}
   \includegraphics[width=\linewidth]{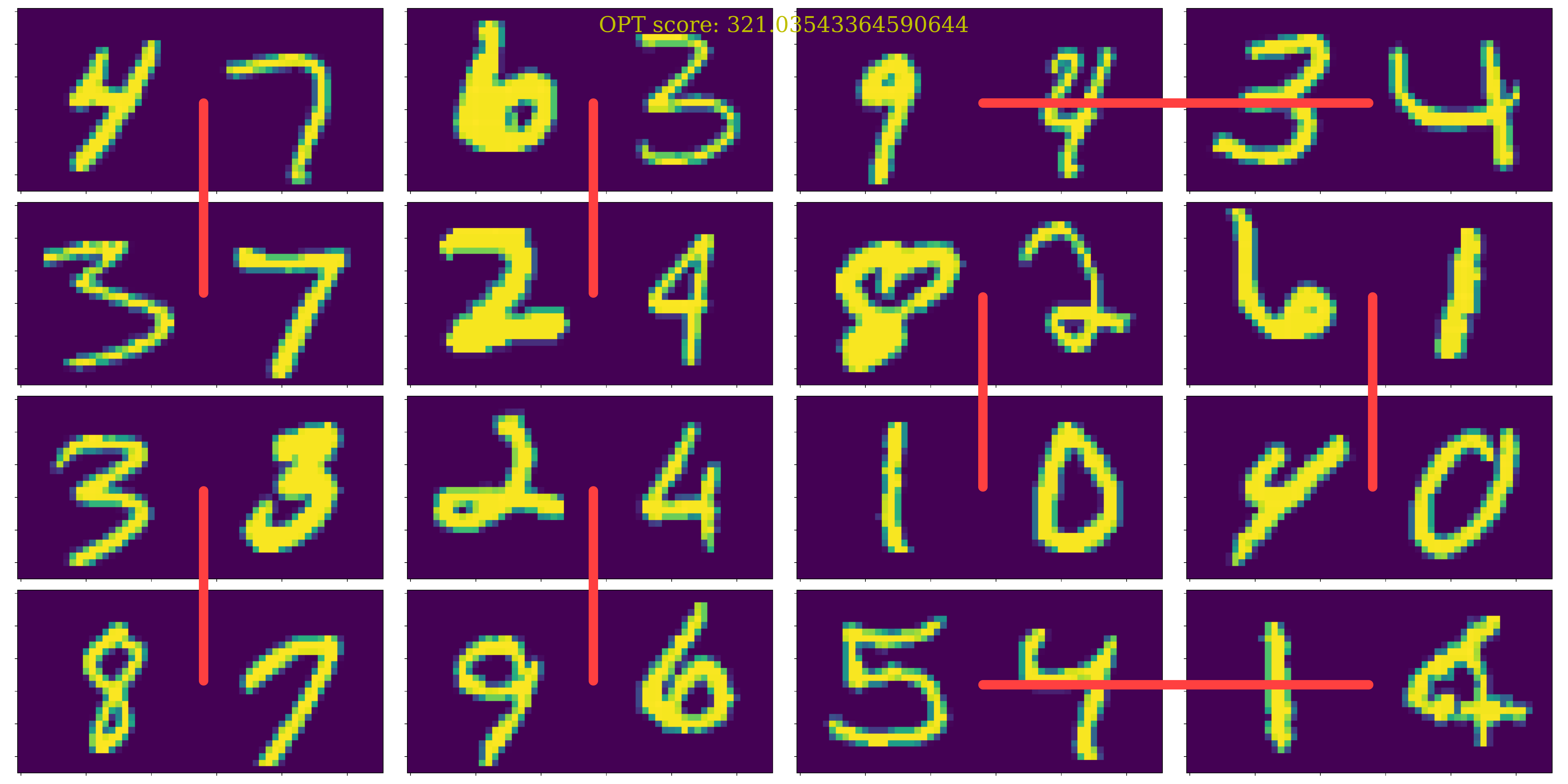}
   \vspace{-6mm}
   \caption{ Optimal $X^*$} \label{fig:application_1_b}
\end{subfigure}
\hspace*{\fill}
\begin{subfigure}{0.235\textwidth}
   \includegraphics[width=\linewidth]{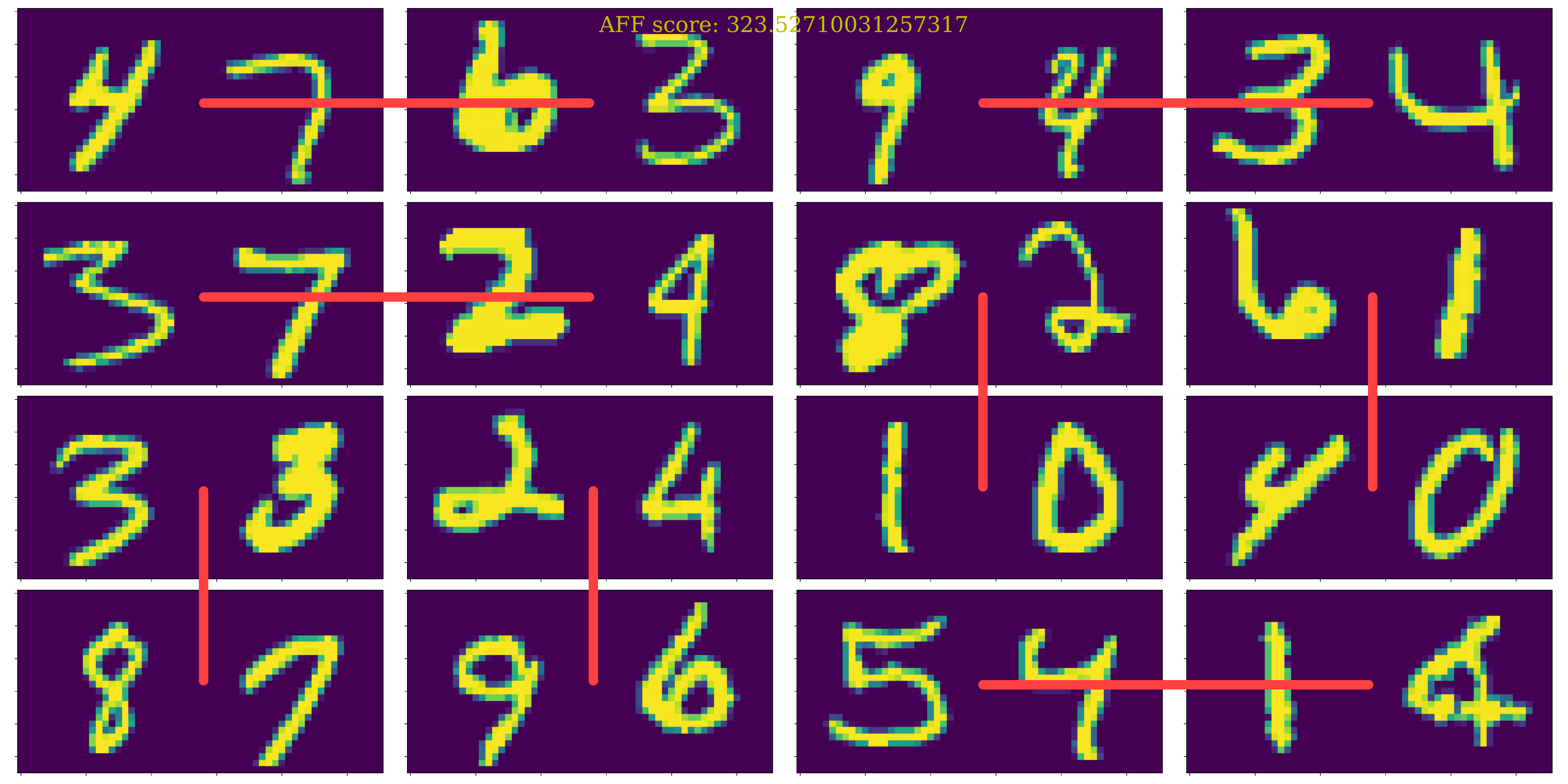}
   \vspace{-6mm}
   \caption{ AFF (Ours)} \label{fig:application_1_c}
\end{subfigure}
\hspace*{\fill}
\begin{subfigure}{0.235\textwidth}
   \includegraphics[width=\linewidth]{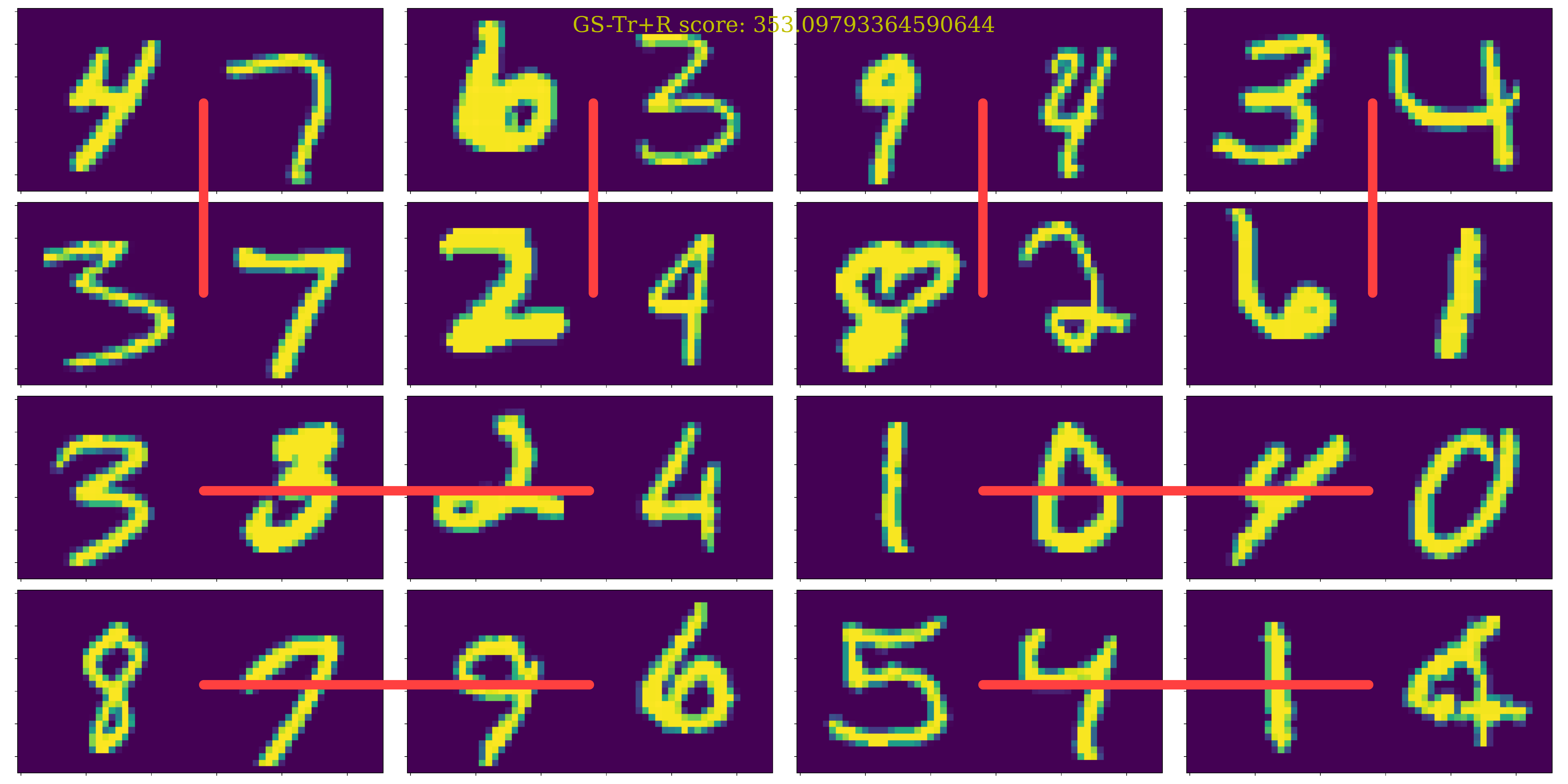}
   \vspace{-6mm}
   \caption{ GS Trick + R} \label{fig:application_1_d}
\end{subfigure}
\vspace{-2mm}
\caption{The visualization for node matching in Application I. Our method avoids large multiplications $87*96$ and $94*82$ where GS-Trick cannot, and generate a solution different but close to OPT.}
\vspace{-0.4cm}
\label{fig:application_1}
\end{figure*}

\subsection{Application I: Feature-based Edge Covering \& Node Matching in Graphs}
This application is inspired by~\cite{poganvcic2019differentiation}. Here, each configuration $C$ is a $4 \times 4$ grid graph whose node attributes are two-digit images generated by random combinations of the pictures in MNIST~\cite{deng2012mnist}. We associated each edge with variables $X\in\{0,1\}^{|E|}$. The objective is the sum of edge weights $f(X;C)=\sum_{e\in E} w_e X_e$ where $w_e$ is unknown in prior and needed to be learned. The ground truth of $w_e$ is a multiplication of the numbers indicated by the images on the two adjacent nodes. We adopt ResNet-50~\cite{he2016deep} (to refine node features) plus GraphSAGE~\cite{hamilton2017inductive} to encode $C$. We consider using both  Eq.\eqref{eq:aff-proxy} and Eq.\eqref{eq:con-proxy} to formulate the relaxed cost $f_r(\bar{X};C)$. Training and validating $f_r$ are based on 100k randomly sampled $C$ paired with randomly sampled $X$. Note that 100k is much smaller than the entire space $\{0,1\}^{|E|}\times \mathcal{C}$ is of size $2^{24}\times 100^{16}$. 

Next, as the constraint here is explicit, we can derive the relaxation of the constraints for this application. First, the constraint relaxation of the edge covering problem can be written as  
\begin{align} \label{eq:edge-cover}
    \textbf{Edge Covering Constraint:} \quad g_r(\bar{X};C) = \sum_{v\in V} \prod_{e:v\in e}(1-\bar{X}_e). 
\end{align}
Each production term in Eq.\eqref{eq:edge-cover} indicates that for each node, at least one edge is selected. We can easily justify that $g_r$ is entry-wise affine and $\Omega=\{X\in\{0,1\}^{|E|}:g_r(X;C)<1\}$ exactly gives the feasible solutions to the edge covering problem. 

Similarly, we can derive the constraint for node matching by adding a further term to Eq.\eqref{eq:edge-cover}.
\begin{align} \label{eq:node-matching}
\textbf{Node Matching Constraint:} \, g_r(\bar{X};C) = \sum_{v\in V} [\prod_{e:v\in e}(1-\bar{X}_e) + \prod_{e_1,e_2:v\in e_1,e_2, e_1\neq e_2}\bar{X}_{e_1}\bar{X}_{e_2}]. 
\end{align}
Here, the second term indicates that no two edges adjacent to the same node can be  selected. This is a case with two constraints while we combine them together. We can easily justify that $g_r$ is entry-wise affine and  $\Omega=\{X\in\{0,1\}^{|E|}:g_r(X;C)<1\}$ groups exactly the feasible solutions to the node matching problem.  

Note that our above derivation also generalizes the node-selection framework in~\cite{karalias2020erdos} to edge selection. With the learned $f_r$ and the derived $g_r$, we further train and validate $\mathcal{A}_{\theta}$ over the 100k sampled $(X,C)$'s and test on another $500$ randomly sampled $C$'s.


\begin{wraptable}{r}{0.43\textwidth}
\vspace{-1.0cm}
\footnotesize{\begin{tabular}{@{}ccc@{}}
\toprule
Method    & Edge covering & Node matching \\ \midrule
Naive+R   & 68.52         & 429.12        \\
RL        & 51.29         & 426.97        \\
GS-Tr+S   & 63.36         & -             \\
GS-Tr+R   & 46.91         & 429.39        \\
CON(ours) & 49.59         & 422.47        \\
AFF(ours) & \textbf{44.55}         & \textbf{418.96}        \\
OPT(gt)   & 42.69         & 416.01        \\ \bottomrule
\end{tabular}}
\vspace{-0.2cm}
\small{
\caption{Performance on application I (graph optimization).}
\label{tab:application_1}}
\vspace{-0.4cm}
\end{wraptable} 
\textbf{Evaluation.} Table~\ref{tab:application_1} shows the evaluation results. In the GS-Tr+S method, the number of sampling is set to $120$ (about $2.5$ times the inference time of our deterministic rounding). Note that for node matching, GS-Tr+S could hardly sample a feasible node matching solution within 120 samples. 
The experiment results show that our principled proxy relaxation exceeds the other baselines on both tasks. Also, we observe that AFF outperforms CON, which results from the fact that $f(X;C)$ in these two problems are naturally in entry-wise affine forms with low-order (1st-order) moments. One instance of node matching randomly selected from the test set is shown in Fig.~\ref{fig:application_1}. More visualization results can be found in Fig.~\ref{fig:application_1_apd} in the appendix.

\subsection{Application II: Resource Allocation in Circuit Design}

\begin{figure}[t]
\vspace{-0.1cm}
\centering
   \includegraphics[width=0.32\linewidth]{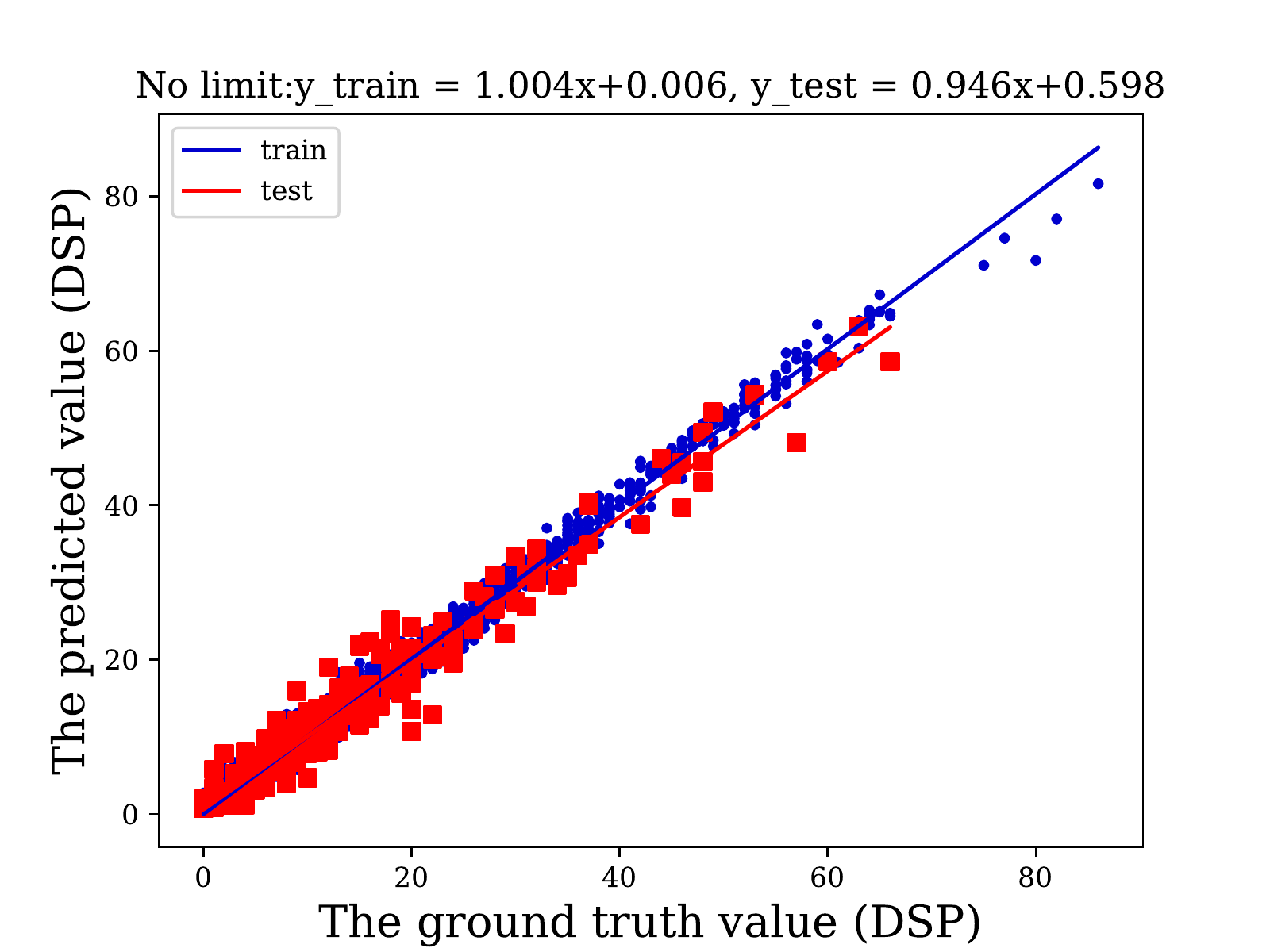}
   \includegraphics[width=0.32\linewidth]{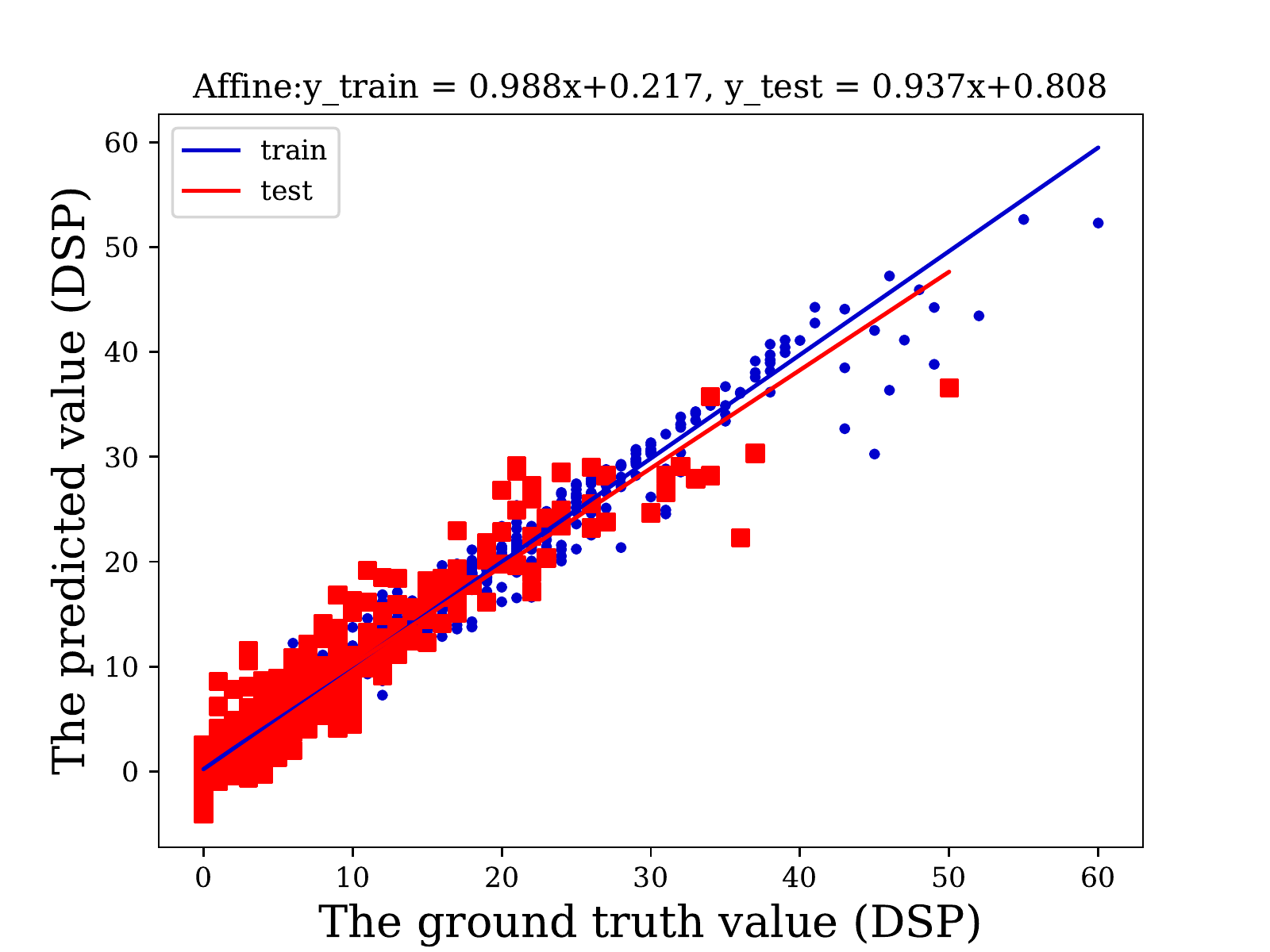}
   \includegraphics[width=0.32\linewidth]{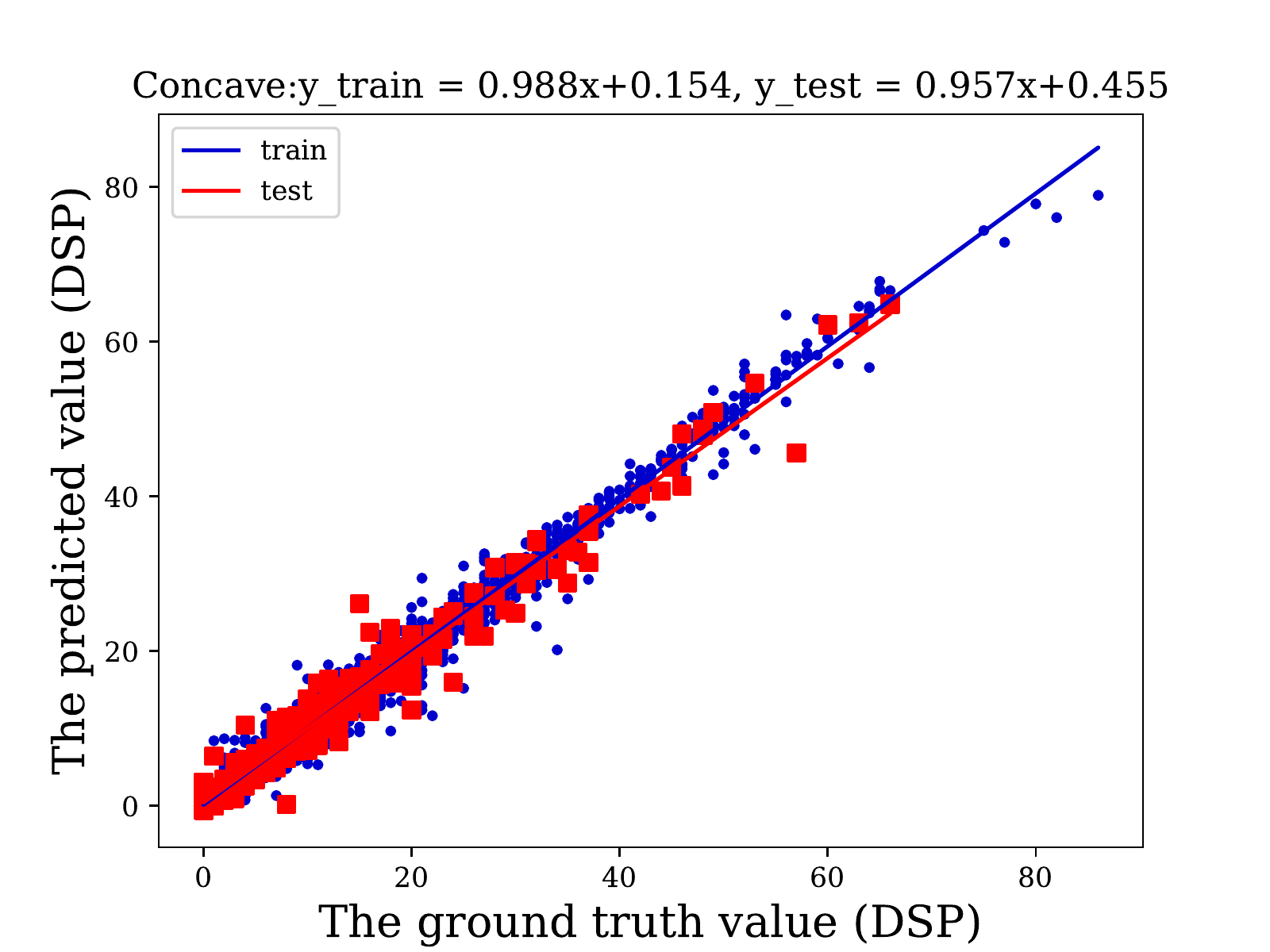}
\vspace{-0.2cm}
\small{\caption{Comparing different proxies for learning DSP usage. Left, no constraint; Middle, entry-wise affine constraint (Eq.~\eqref{eq:aff-proxy}); Right, entry-wise concave constraint (Eq.\eqref{eq:con-proxy})}
\label{fig:application_2_proxy}}
\vspace{-0.7cm}
\end{figure}

Resource allocation in field-programmable gate array (FPGA) design is a fundamental problem which can lead to largely varied circuit quality after synthesis, such as area, timing, and latency. In this application, we follow the problem formulation in~\cite{wu2021ironman, wu2022highlevel}, where the circuit is represented as a data flow graph (DFG), and each node represents an arithmetic operation such as multiplication or addition.
The goal is to find a resource allocation for each node to be either digital signal processor (DSP) or look-up table (LUT), such that the final circuit area (i.e., actual DSP and LUT usage) after synthesis is minimized. Notably, different allocation solutions result in greatly varied DSP/LUT usage due to complicated synthesis process, which cannot be simply summed up over each node. To obtain precise DSP/LUT usage, one must run high-level synthesis (HLS)~\cite{VitisHLS} and place-and-route~\cite{Vivado} tools, which can take up to hours~\cite{wu2021ironman, wu2022highlevel}.


In this application, each configuration $C$ is a DFG with $>100$ nodes, where each node is allocated to either DSP or LUT.
Node attributes include operation type (i.e., multiplication or addition) and data bitwidth. More details about the dataset can be found in Appendix~\ref{apd:application_2-proxy}. Let $X\in \{0,1\}^{|V|}$ denote the mapping to DSP or LUT. Let $f_r$ and $g_r$ denote the proxies of actual LUT and actual DSP usage, respectively. Note that given different constraints on the DSP usage, we will normalize $g_r$ as introduced in Sec.~\ref{def_CO}. We train and validate $f_r, g_r, \mathcal{A}_{\theta}$ on $8,000$ instances that consist of $40$ DFGs ($C$), each DFG with $200$ different mappings ($X$), and test $\mathcal{A}_{\theta}$ over $20$ DFGs. Note that the actual LUT and DSP usages of each training instance has been collected by running HLS in prior. We also run HLS to evaluate the actual LUT and DSP usages for the testing cases given the learned mappings.

\begin{figure}[t]
\begin{minipage}{0.6\textwidth}\small
\flushright
\setlength\tabcolsep{3.5pt}
\resizebox{0.9\textwidth}{!}{\begin{tabular}{@{}ccccccccc@{}}
\toprule
DSP usage  & 40\% & 45\% & 50\% & 55\% & 60\% & 65\% & 70\% & rank-avg \\ \midrule
\specialrule{0em}{3pt}{1pt}
SA      & 3.41 & 3.08 & 3.50 & 3.33 & 3.66 & 4.16 & 4.08 & 3.60     \\
\specialrule{0em}{3pt}{1pt}
GA      & 2.75 & 3.41 & 2.83 & 2.91 & 3.00 & 3.00 & 2.75 & 2.95     \\
\specialrule{0em}{3pt}{1pt}
RL      & 3.33 & 3.58 & 3.83 & 3.25 & 2.91 & 2.83 & \textbf{2.41} & 3.16     \\
\specialrule{0em}{3pt}{1pt}
GS-Tr+R & 3.58 & 2.91 & 2.58 & 3.16 & \textbf{2.33} & \textbf{2.58} & 3.00 & 2.87     \\
\specialrule{0em}{3pt}{1pt}
CON     & \textbf{1.83} & \textbf{2.00} & \textbf{2.25} & \textbf{2.25} & 3.08 & \textbf{2.41} & 2.66 & \textbf{2.35}     \\ \bottomrule
\end{tabular}}
\end{minipage}
\hfill
\begin{minipage}{0.34\textwidth}
\flushleft\includegraphics[width=1.5in]{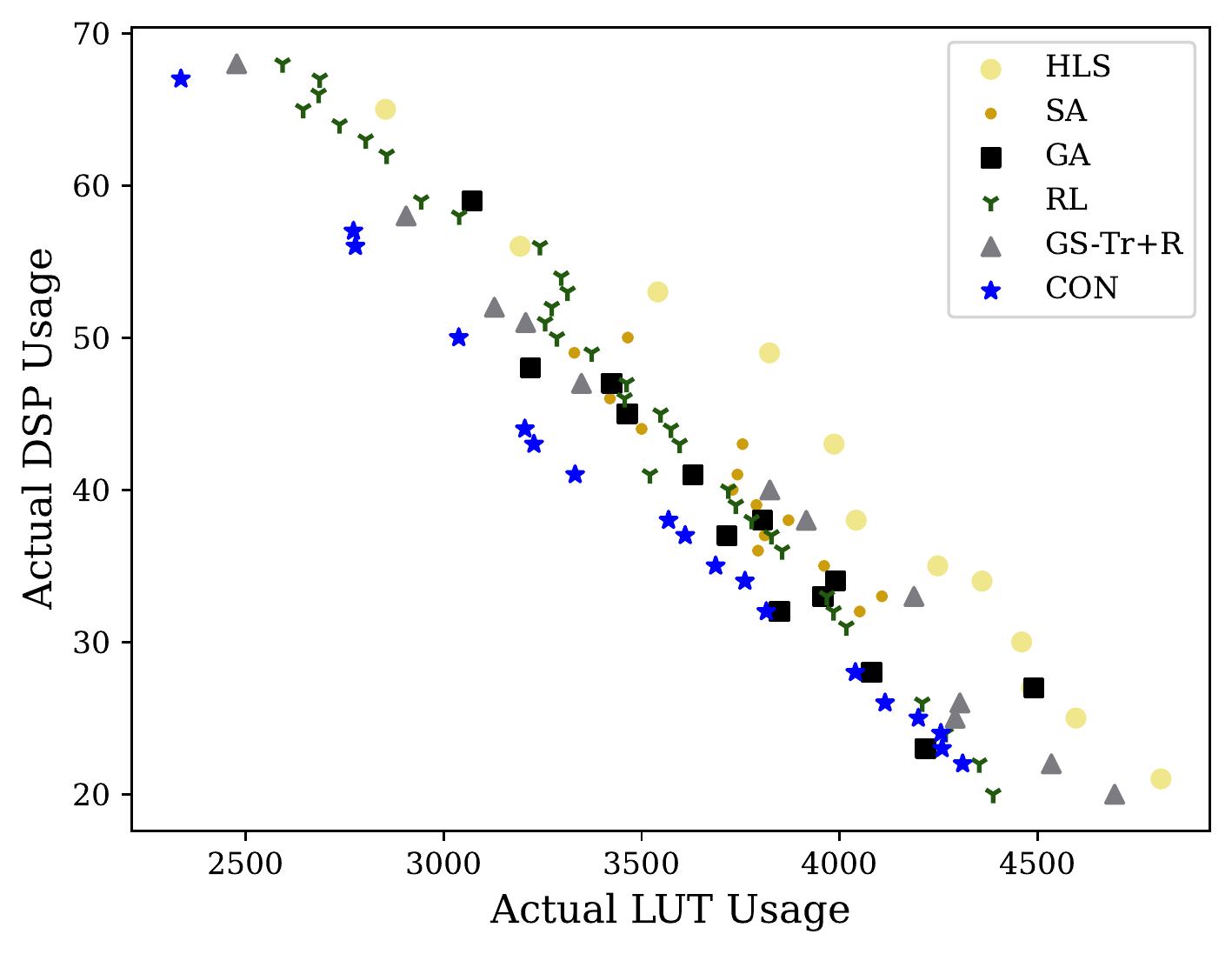}
\end{minipage}
\vspace{-0.2cm}
\small{\caption{The left table shows averaged ranks of the LUT usage given by different methods with the constraint of different percentage of DSP usage in Application II (resource allocation). The right figure shows the DSP-LUT usage amount relationship on one test configuration. The HLS baseline denotes the optimal synthesis results among 200 random mappings.}
\label{tab:application_2_ranking}}
\vspace{-0.4cm}
\end{figure}

\textbf{Evaluation.}
We rank each method's best actual LUT usage under the constraint of different percentages ($40 \%$ - $70 \%$) of the maximum DSP usage in each testing instance, then calculate the averaged ranks. Fig.~\ref{tab:application_2_ranking} shows the results. Our entry-wise concave proxy achieves the best performance. GS-Tr+R is slightly better than RL, and both of them exceed SA and GA. We do not include our entry-wise affine proxy in the ranking list, because the affine proxy could be much less accurate than the proxy without constraints and the entry-wise concave proxy. The comparison between these proxies on learning DSP usage (\& LUT usage) is shown in Fig.~\ref{fig:application_2_proxy} (\& Fig.~\ref{fig:application_2_proxy_apd} in the appendix, respectively). The gap between different proxies indicates the FPGA circuit contains high-order moments of the input optimization variables and 2-order entry-wise affine proxy cannot model well. 
We do not include the result of GS-Tr+S and Naive+R, because these methods perform poor and could hardly generate feasible solutions given a constraint of DSP usage. We leave their results in Table.~\ref{tab:application_2_value_apd} in the appendix. 
Moreover, we compare the training time between different methods. To be fair, all methods run on the same server with a Quadro RTX 6000 GPU. The RL based optimizer takes $22$ GB GRAM, while other optimizers only take $7$ GB on average.  Fig.~\ref{fig:application_2_time_apd} in the appendix further demonstrates that our methods and GS-T methods require much less training time than RL. 

Also, to give a fair comparison between learning-based approaches and traditional approaches, we implement GA with parallel (on GPU) cost-value inference for all the populations in each generation. We set the population size as $256$, which is the same as the batch size that we used to train/infer $\mathcal{A}_{\theta}$. The performance of GA in Fig.~\ref{tab:application_2_ranking} is obtained under the condition that the inference time of the implemented parallel GA is about the same as that of our CON method.  Fig.~\ref{fig:application_2_apd_ga} in the appendix provides more detailed comparison on the performance and the inference time between GA with different numbers of generations and our CON method.


\vspace{-0.1cm}
\subsection{Application III: Imprecise Functional Unit Assignment in Approximate Computing}
\vspace{-0.1cm}
One fundamental problem in approximate computing (AxC) is to assign imprecise functional units (a.k.a., AxC units) to execute operations such as multiplication or addition~\cite{han2013approximate, mittal2016survey, venkatesan2011macaco, ma2021workload, li2015joint}, aiming to significantly reduce circuit energy with tolerable error. We follow the problem formulation in~\cite{ma2021workload}, where given a computation graph, each node represents either multiplication or addition.
The incoming edges of a node represent its two operands.
The goal is to assign AxC units to a certain number of nodes while minimizing the expected relative error of the output of the computation graph.

In this application, each configuration $C$ is a computation graph with 15 nodes (either multiplication or addition) that maps a vector in $\mathbb{R}^{16}$ to $\mathbb{R}$. A fixed number $\theta$ of nodes are assigned to AxC units with produce $10\%$ relative error. 
Let $X\in \{0,1\}^{|V|}$ denote whether a node is assigned to an AxC unit or not;
the proxy of the objective $f_r$ is the expected relative error at the output.
We use 100k $(X,C)$ as the training dataset and the entire solution space is $2^{15}\times 2^{15}$. For each $(X,C)$, the ground-truth, i.e., expected relative error, is computed by averaging 1k inputs sampled uniformly at random from $[0,10]^{16}$. The constraint $g_r$ is $\sum_{v\in V}X_v\geq \theta$ with normalization, where $\theta\in\{3,5,8\}$. We test the learned $\mathcal{A}_{\theta}$ on $500$ unseen configurations.

\begin{table}[t]
\centering
\resizebox{0.8\textwidth}{!}{\begin{tabular}{@{}cccccccccc@{}}
\toprule
Threshold $\theta$ & C-In & C-Out & Na\"{i}ve & RL    & GS-Tr+S & GS-Tr+R & CON & AFF & OPT \\ \midrule
3 AxC units     & 12.42           & 12.44          & 3.62 & 7.68 & 4.87    & 3.24    & 3.18      & \textbf{3.10}     & 2.77    \\
5 AxC units    &  14.68          & 14.65          & 6.20 & 10.15 & 8.03    & 5.86    & \textbf{5.13}      & 5.38      & 4.74    \\
8 AxC units    & 17.07           & 17.04          & 11.12 & 12.83 & 12.65   & 10.62   & 10.17     & \textbf{10.04}     & 8.56    \\ \bottomrule
\end{tabular}}
\vspace{1mm}
\small{\caption{Relative errors of different methods  with the AxC unit constraint as 3,5,8 in Application III.}
\label{tab:application_3}}
\vspace{-0.7cm}
\end{table}

\textbf{Evaluation.} Table.~\ref{tab:application_3} shows the averaged relative errors of the assignments by different methods. 
The problem is far from trivial. Intuitively, assigning AxC units closed to the output, we may expect small error. However, C-Out performs bad.  
Our proxies AFF and CON obtain comparable best results. The MAE loss values of the two proxies are also similar, as shown in Table~\ref{tab:proxy_loss_apd} in the appendix. The reason is that the circuit is made up of $4$ layers in total which leads to at most $4$-order moments in the objective function, which is in a medium-level complexity. 
Training time is also studied for this application, resulting in the same conclusion as application II (See Table~\ref{tab:application_3_apd} in the appendix). 

\vspace{-0.1cm}
\section{Conclusion}
\vspace{-0.1cm}
\label{sec:conclusion}
This work introduces an unsupervised end-to-end framework to resolve LCO problems based on the relaxation-plus-rounding technique. With our entry-wise concave architecture, our framework guarantees that a low objective value could lead to qualified discrete solutions. 
Our framework is particularly good at solving PCO problems where the objectives need to be modeled and learned. Real-world applications demonstrate the superiority of our method over RL and gradient-relaxation approaches in both optimization performance and training efficiency.
In the future, we aim to further broaden our framework to the applications where the optimization variables allow using more complex embeddings than  binary embeddings.  


\section{Acknowledgement}
We greatly thank all the reviewers for their valuable feedback and the insightful suggestions. H. Wang and P. Li are partially supported by 2021
JPMorgan Faculty Award and the NSF award OAC-2117997.

\bibliographystyle{ACM}
\bibliography{reference}
\newpage
\newpage
\appendix





\section{Deferred Theoretical Arguments}
\subsection{Proof of Theorem~\ref{thm:main}} 
\label{sec:proof-main}
We analyze the rounding process of the relaxed solution $\bar{X} = \mathcal{A}_{\theta}(C), \bar{X} \in [0,1]^n$ into the integral solution $\hat{X} \in \{0,1\}^n$. Let $\bar{X}_i, \hat{X}_i, i = \{1,2,...,n\}$ denote their entries respectively. The rounding procedure has no requirement on the order of the rounding sequence, w.l.o.g, suppose we round from index $i=1$ to $i=n$. In practice, it might be better to sort $\bar{X}$ and round the entries according to their ranks. We have the following inequations:
\begin{equation}
\begin{aligned}
& \quad l_r(\theta;C) \\
& = f_r([\bar{X}_1,\bar{X}_2,...,\bar{X}_n];C) + \beta g_r([\bar{X}_1, \bar{X}_2,..., \bar{X}_n];C)\\
& \stackrel{(a)}{\geq} \bar{X}_1
(f_r([1,\bar{X}_2,...,\bar{X}_n];C) + \beta g_r([1,\bar{X}_2,...,\bar{X}_n];C)) \\
& + (1 - \bar{X}_1)
(f_r([0,\bar{X}_2,...,\bar{X}_n];C) + \beta g_r([0,\bar{X}_2,...,\bar{X}_n];C)) \\
& \geq \bar{X}_1 (\min_{j_1=\{0,1\}} f_r([j_1,\bar{X}_2,...,\bar{X}_n];C) + \beta g_r([j_1,\bar{X}_2,...,\bar{X}_n];C)) \\
& + (1-\bar{X}_1) (\min_{j_1=\{0,1\}} f_r([j_1,\bar{X}_2,...,\bar{X}_n];C) + \beta g_r([j_1,\bar{X}_2,...,\bar{X}_n];C))\\
& \stackrel{(b)}{=}  (f_r([\hat{X}_1,\bar{X}_2,...,\bar{X}_n];C) + \beta g_r([\hat{X}_1,\bar{X}_2,...,\bar{X}_n];C))\\
& \geq \min_{j_2 = \{0,1\}}
(f_r([\hat{X}_1,j_2,...,\bar{X}_n];C) + \beta
g_r([\hat{X}_1,j_2,...,\bar{X}_n];C))\\
& =  (f_r([\hat{X}_1,\hat{X}_2,...,\bar{X}_n];C) + \beta
g_r([\hat{X}_1,\hat{X}_2,...,\bar{X}_n];C)) \\
& \geq ...\\
& \geq \min_{j_n \in \{0,1\}}
(f_r([\hat{X}_1,\hat{X}_2,...,j_n];C) + \beta
g_r([\hat{X}_1,\hat{X}_2,...,j_n];C)\\
& = f_r(\hat{X};C) + \beta g_r(\hat{X};C)\\
& \stackrel{(c)}{=} f(\hat{X};C) + \beta g(\hat{X};C),
\end{aligned}
\end{equation}where (a) is due to $l_r(\theta;C)$'s entry-wise concavity w.r.t. $\bar{X}$ and Jensen's inequality, (b) is due to the definition $\hat{X}_1=\arg\min_{j_1=\{0,1\}} f_r([j_1,\bar{X}_2,...,\bar{X}_n];C) + \beta g_r([j_1,\bar{X}_2,...,\bar{X}_n];C)$, and (c) is due to the assumption that the neural network based proxies could learn the objective and the constraints perfectly for $\hat{X} \in \{0,1\}^n$.
The inequalities above demonstrate the fact that the loss value is non-increasing via the whole rounding process.
By this, once the learnt parameter $\theta$ achieves $l_r(\theta;C) < \beta$, we could get $f(\hat{X};C)+\beta g(\hat{X};C) \leq l_r(\theta;C) < \beta$. Because of the settings that $f(\cdot), g(\cdot) \geq 0$, we have $f(\hat{X};C) < l_r(\bar{X};C), \ \ \text{s.t.} \ \ g(\hat{X};C) < 1$.

\subsection{Proof of Theorem~\ref{thm:app}}
\label{sec:proof_thm:app}
Set $h_r(\bar{X}) = \sum_{X\in \{0,1\}^n} h(X) \prod_{j=1}^n \bar{X}_j^{X_j}(1-\bar{X}_j)^{(1-X_j)}$.

We first prove that $h_r(\bar{X})$ with the form above satisfies (a) $h_r(X)=h(X)$ for $X\in \{0,1\}^n$. 

Given one $X'\in\{0,1\}^n$, by setting $\bar{X} = X'$, we have 
\begin{align*}
    \prod_{j=1}^n X_j'^{X_j}(1-X_j')^{(1-X_j)} = 1, \quad \text{if}\; X=X', \;\text{and otherwise 0} . 
\end{align*}
Therefore, in $h_r(X')$, there is only one term $h(X')\prod_{j=1}^n X_j'^{X_j'}(1-X_j')^{(1-X_j')}=h(X')$ left. So, $h_r(X')=h(X')$, which satisfies (a).

Then we prove that $h_r(\bar{X})$ satisfies (b) $h_r$ is entry-wise affine. From the definition, we have:
\begin{align*}
\bar{X}_j^{X_j}(1-\bar{X}_j)^{1-X_j} = \begin{cases}
\bar{X}_j & X_j = 1 \\
1 - \bar{X}_j & X_j = 0.
\end{cases}
\end{align*}
Consider two sequences $\bar{X}, \bar{X}' \in \{0,1\}^n$ with the entries $\bar{X}_i, \bar{X}_i', i=\{1,2,...,n\}$. We have
\begin{align*}
& \quad \gamma[\bar{X}_i^{X_i}(1-\bar{X}_i)^{1-X_i}]+(1-\gamma)[\bar{X}_i'^{X_i}(1-\bar{X}_i')^{1-X_i}] \\
& = \begin{cases}
1 - \gamma \bar{X}_i - (1-\gamma)\bar{X}_i' & X_i = 0 \\
\gamma \bar{X}_i + (1-\gamma)\bar{X}_i' & X_i = 1.
\end{cases}\\
& = \left[\gamma \bar{X}_i+(1-\gamma)\bar{X}_i\right]^{X_i} \left[1-\gamma \bar{X}_i  - (1-\gamma)\bar{X}_i'\right]^{(1-X_i)}. 
\end{align*}
W.l.o.g, we assume that only the entries $\bar{X}_i$ and $\bar{X}_i'$ in the two sequences are different. For any $\gamma\in[0,1]$, we may use the above equality and have
\begin{align*}
& \gamma h_r(\bar{X}) + (1-\gamma) h_r(\bar{X}') \\ 
& = \gamma \sum_{X\in \{0,1\}^n} h(X) \prod_{j=1}^n \bar{X}_j^{X_j}(1-\bar{X}_j)^{(1-X_j)} + (1-\gamma) \sum_{X\in \{0,1\}^n} h(X) \prod_{j=1}^n \bar{X}_j'^{X_j}(1-\bar{X}_j')^{(1-X_j)} \\
& = \gamma \sum_{X \in \{0,1\}^n} h(X) \bar{X}_i^{X_i}(1-\bar{X}_i)^{1-X_i} \prod_{j=1, j\neq i}^n \bar{X}_j^{X_j}(1-\bar{X}_j)^{(1-X_j)} \\
&+ (1-\gamma) \sum_{X \in \{0,1\}^n} h(X) \bar{X}_i'^{X_i}(1-\bar{X}_i')^{1-X_i} \prod_{j=1, j\neq i}^n \bar{X}_j^{X_j}(1-\bar{X}_j)^{(1-X_j)} \\
& = \! \sum_{X \in \{0,1\}^n} \! h(X) \left[\gamma\bar{X}_i^{X_i}(1 \! - \! \bar{X}_i)^{1 \! - \! X_i}+(1 \! - \! \gamma)\bar{X}_i'^{X_i}(1 \! - \! \bar{X}_i')^{1 \! - \! X_i}\right]  \prod_{j = 1,j \neq i}^n \bar{X}_j^{X_j}(1 \! - \! \bar{X}_j)^{(1 \! - \! X_j)} \\
& =  \! \sum_{X \in \{0,1\}^n} \! h(X) \left[\gamma \bar{X}_i+(1 \! - \! \gamma)\bar{X}_i\right]^{X_i} \left[1 \! - \! \gamma \bar{X}_i \! - \!  (1 \! - \! \gamma)\bar{X}_i'\right]^{(1 \! - \! X_i)}
\prod_{j = 1, j\neq i}^n \bar{X}_j^{X_j}(1 \! - \! \bar{X}_j)^{(1 \! - \! X_j)} \\
& = h_r(\gamma\bar{X} + (1-\gamma)\bar{X'})
\end{align*}

Thus, we prove that the form of $h_r(X)$ is entry-wise affine.

\subsection{Proof of Proposition~\ref{proposition_concave}}
\label{sec:proof-prop}
Suppose that the output of function $h(X_1, X_2)$ is denoted as follows.
\begin{equation}
\begin{aligned}
a_0 = h(0, 0), \ \ a_1 = h(0, 1), \ \ a_2 = h(1, 0), \ \ a_3 = h(1, 1)
\end{aligned}
\end{equation} 
Then, we pick out the largest value $a_i$ among $a_0, a_1, a_2, a_3$. W.l.o.g, we assume that $a_0$ is the largest and they hold the following inequations:
\begin{equation}
\begin{aligned}
a_0 \geq a_1, \ \ a_0 \geq a_2, \ \ a_0 \geq a_3.
\end{aligned}
\end{equation}
Then, we define our entry-wise concave function $h_r(X_1, X_2)$ as follows.
\begin{equation}
\begin{aligned}
h_r(X_1, X_2) & = a_0 - \sum_{i = 1}^3 \text{Relu}(f[i])
\\ & = a_0 - \text{Relu}(f[1]) - \text{Relu}(f[2]) - \text{Relu}(f[3]), 
\end{aligned}
\end{equation}where
\begin{equation}
\begin{aligned}
\text{Relu}(f[1]) = \text{Relu}((a_0-a_1)(X_2-X_1)) = 
\begin{cases}
a_0 - a_1 & X_1 = 0, X_2 = 1 \\
0 & \text{otherwise},
\end{cases} \\
\text{Relu}(f[2]) = \text{Relu}((a_0 - a_2)(X_1-X_2)) = 
\begin{cases}
a_0 - a_2 & X_1 = 1, X_2 = 0 \\
0 & \text{otherwise},
\end{cases} \\
\text{Relu}(f[3]) = \text{Relu}((a_0 - a_3)(X_1 + X_2 - 1)) = 
\begin{cases}
a_0 - a_3 & X_1 = 1, X_2 = 1 \\
0 & \text{otherwise}.
\end{cases}
\end{aligned}
\end{equation}

\section{Additional Results}
\subsection{Application I}
\begin{figure*}[h]
\centering
\begin{subfigure}{0.235\textwidth}
   \includegraphics[width=\linewidth]{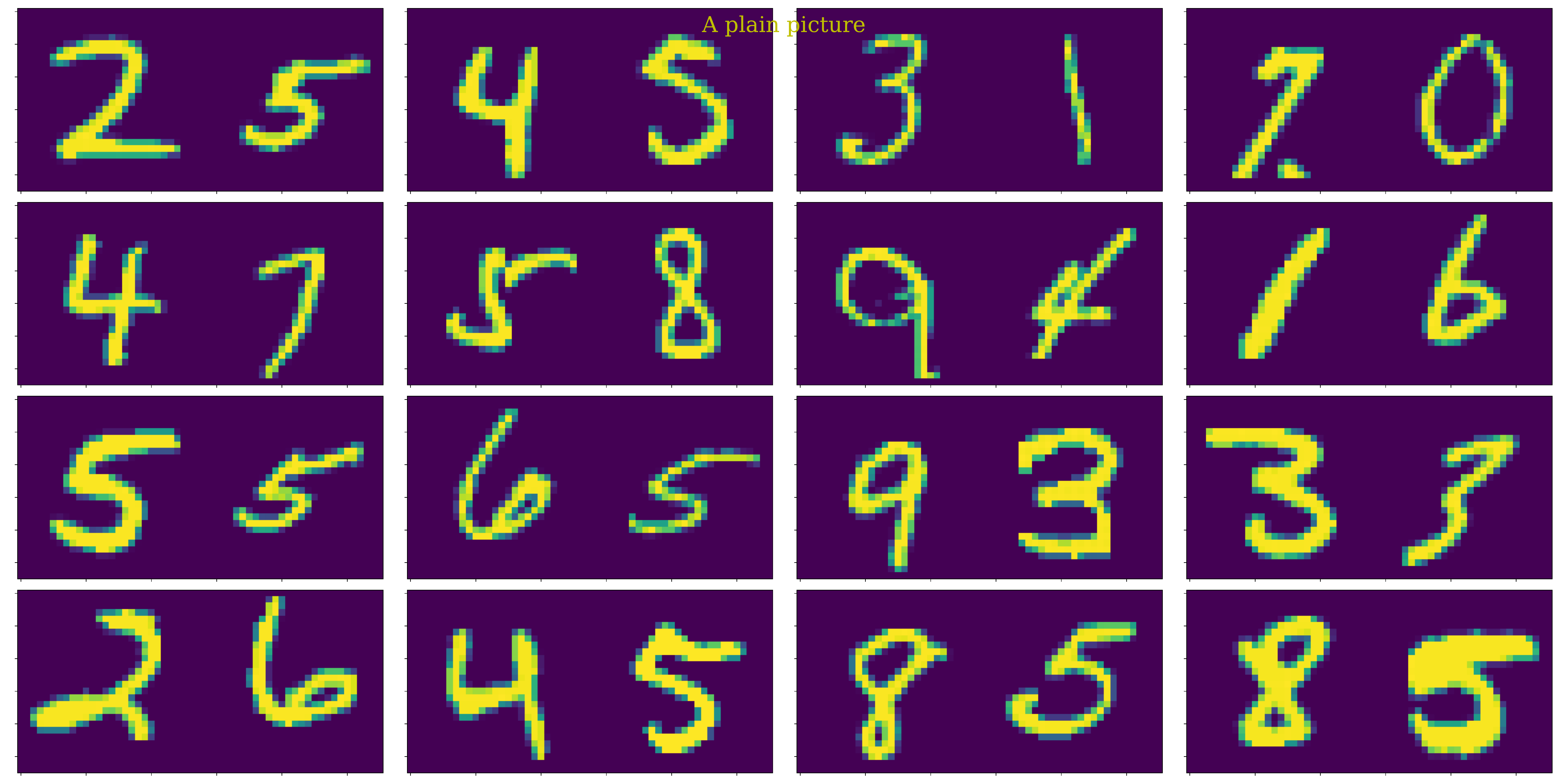}
   \vspace{-6mm}
   \caption{ Matching:Config} \label{fig:application1_apd1}
\end{subfigure}
\hspace*{\fill}
\begin{subfigure}{0.235\textwidth}
   \includegraphics[width=\linewidth]{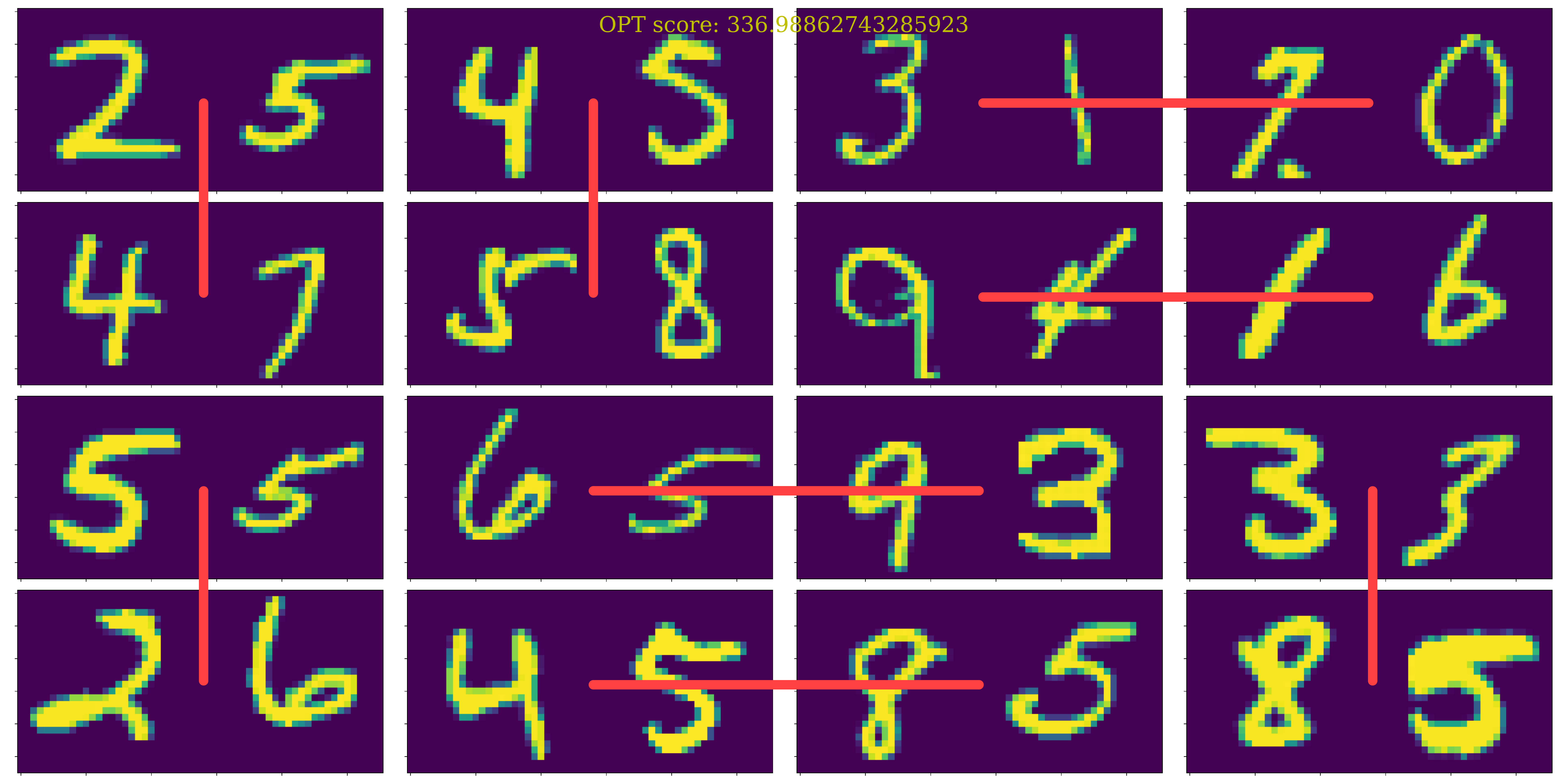}
   \vspace{-6mm}
   \caption{ Matching:Optimal $X^*$} \label{fig:application1_apd2}
\end{subfigure}
\hspace*{\fill}
\begin{subfigure}{0.235\textwidth}
   \includegraphics[width=\linewidth]{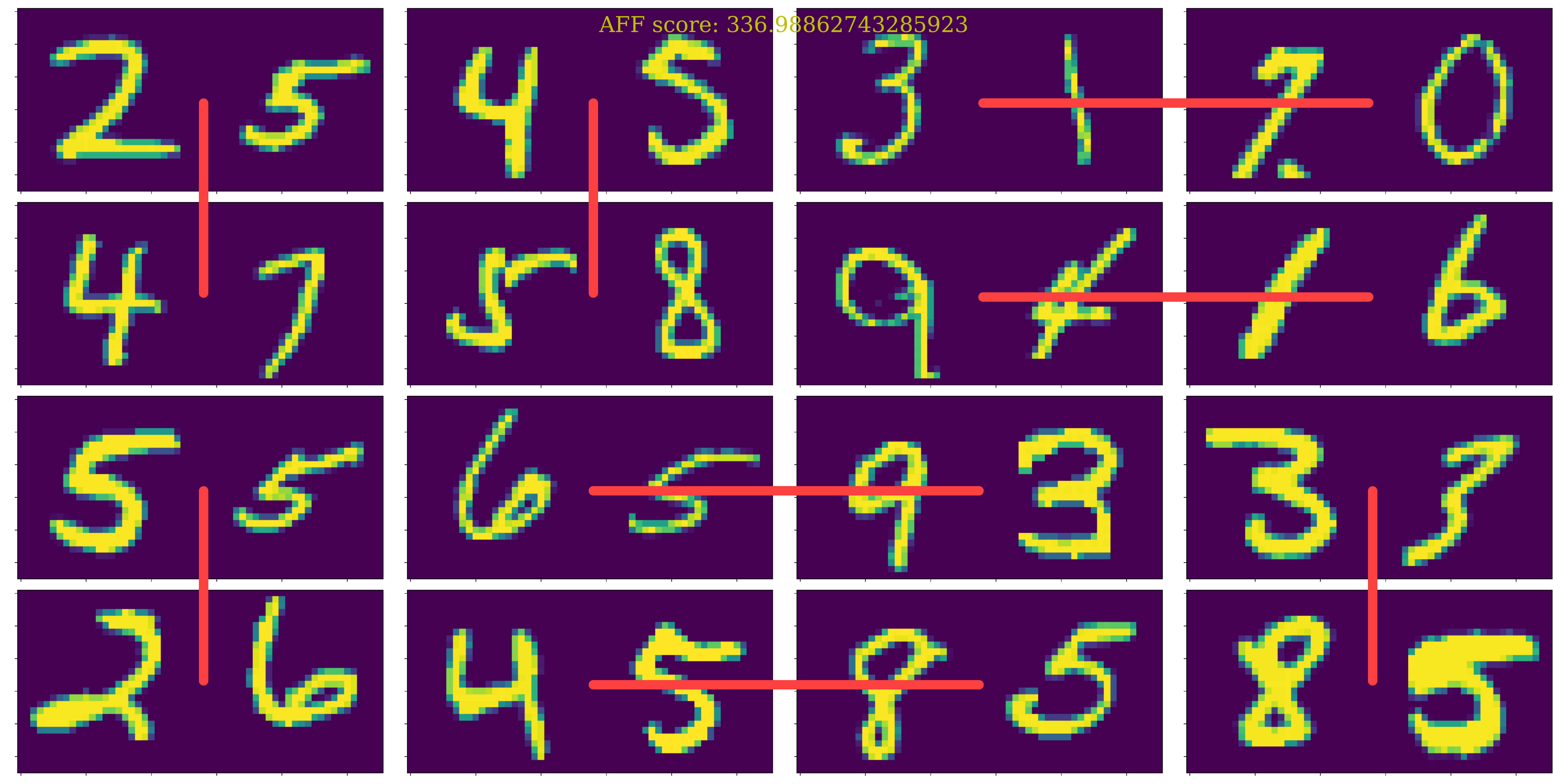}
   \vspace{-6mm}
   \caption{ Matching:AFF (Ours)} \label{fig:application1_apd3}
\end{subfigure}
\hspace*{\fill}
\begin{subfigure}{0.235\textwidth}
   \includegraphics[width=\linewidth]{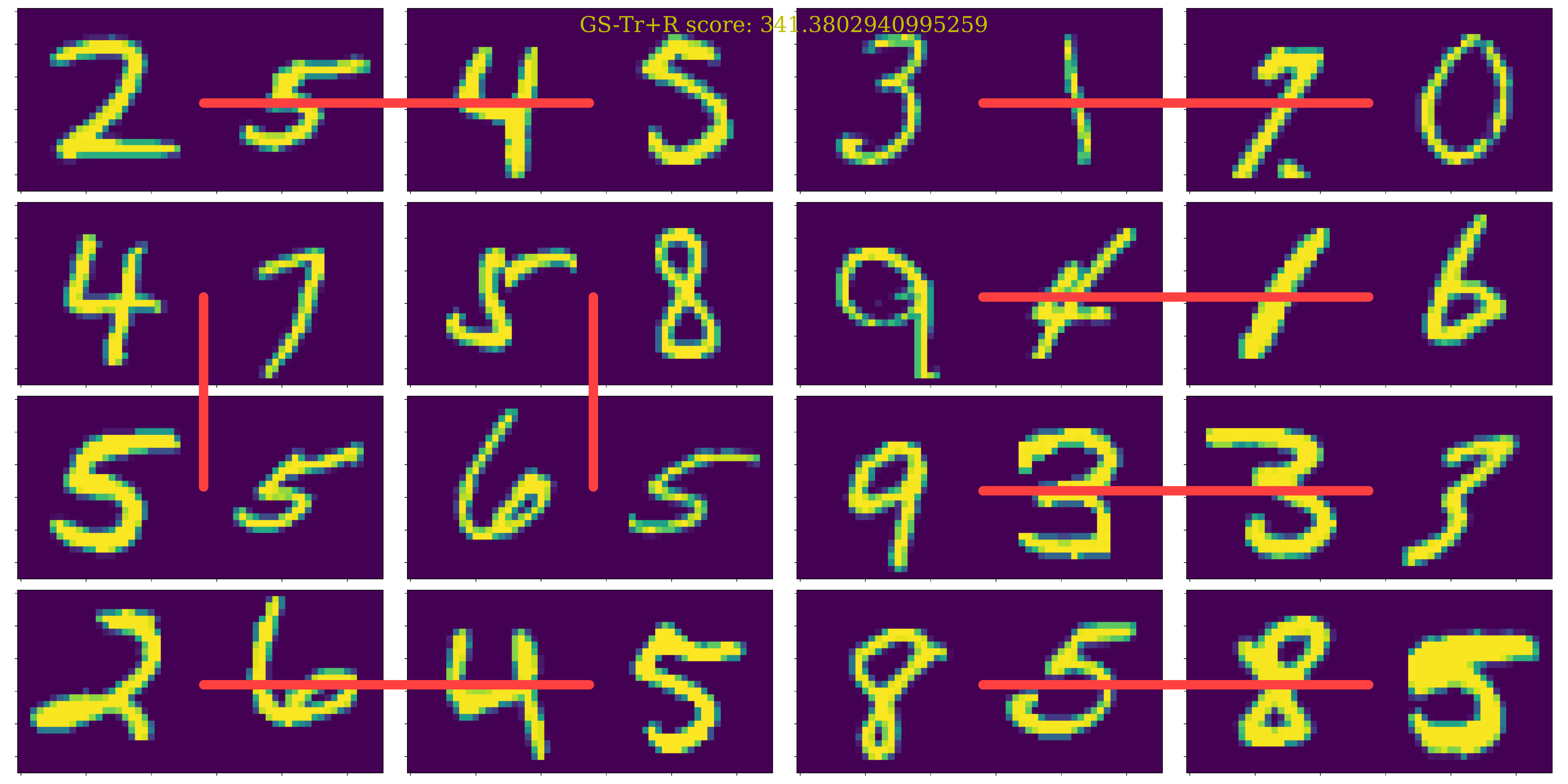}
   \vspace{-6mm}
   \caption{ Matching:GS-Tr + R} \label{fig:application1_apd4}
\end{subfigure}

\begin{subfigure}{0.235\textwidth}
   \includegraphics[width=\linewidth]{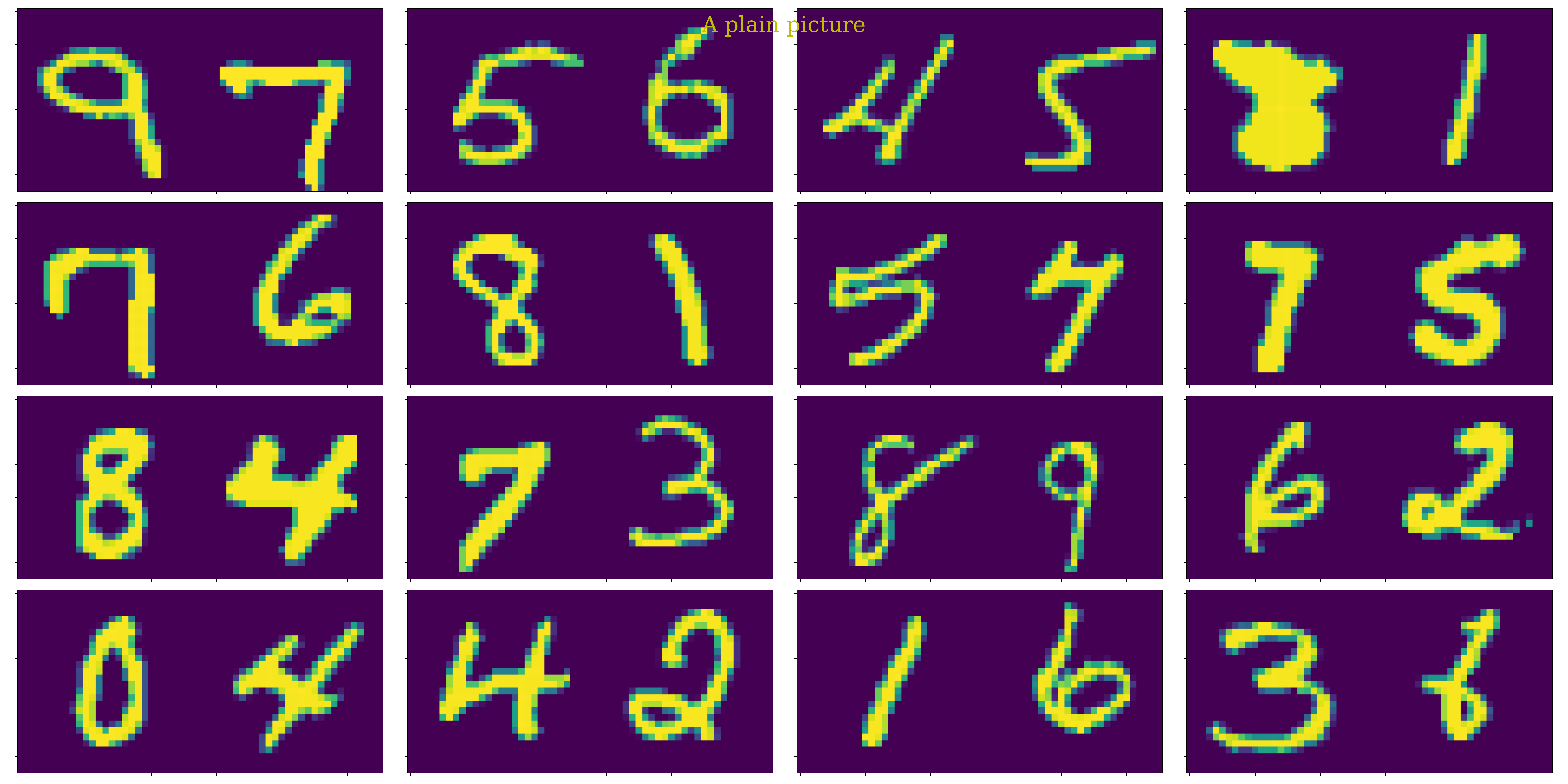}
   \vspace{-6mm}
   \caption{ Matching:Config} \label{fig:application1_apd5}
\end{subfigure}
\hspace*{\fill}
\begin{subfigure}{0.235\textwidth}
   \includegraphics[width=\linewidth]{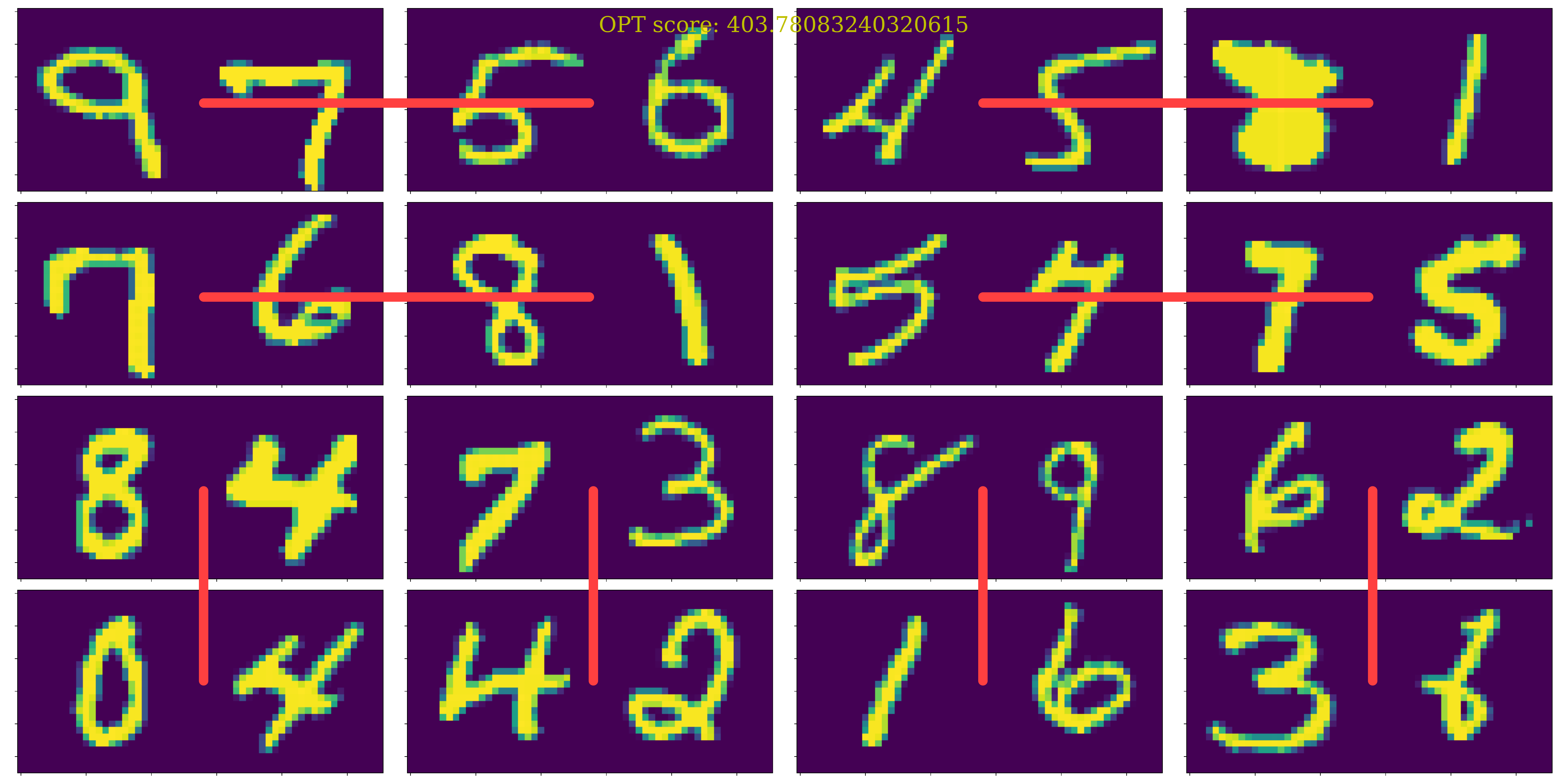}
   \vspace{-6mm}
   \caption{ Matching:Optimal $X^*$} \label{fig:application1_apd6}
\end{subfigure}
\hspace*{\fill}
\begin{subfigure}{0.235\textwidth}
   \includegraphics[width=\linewidth]{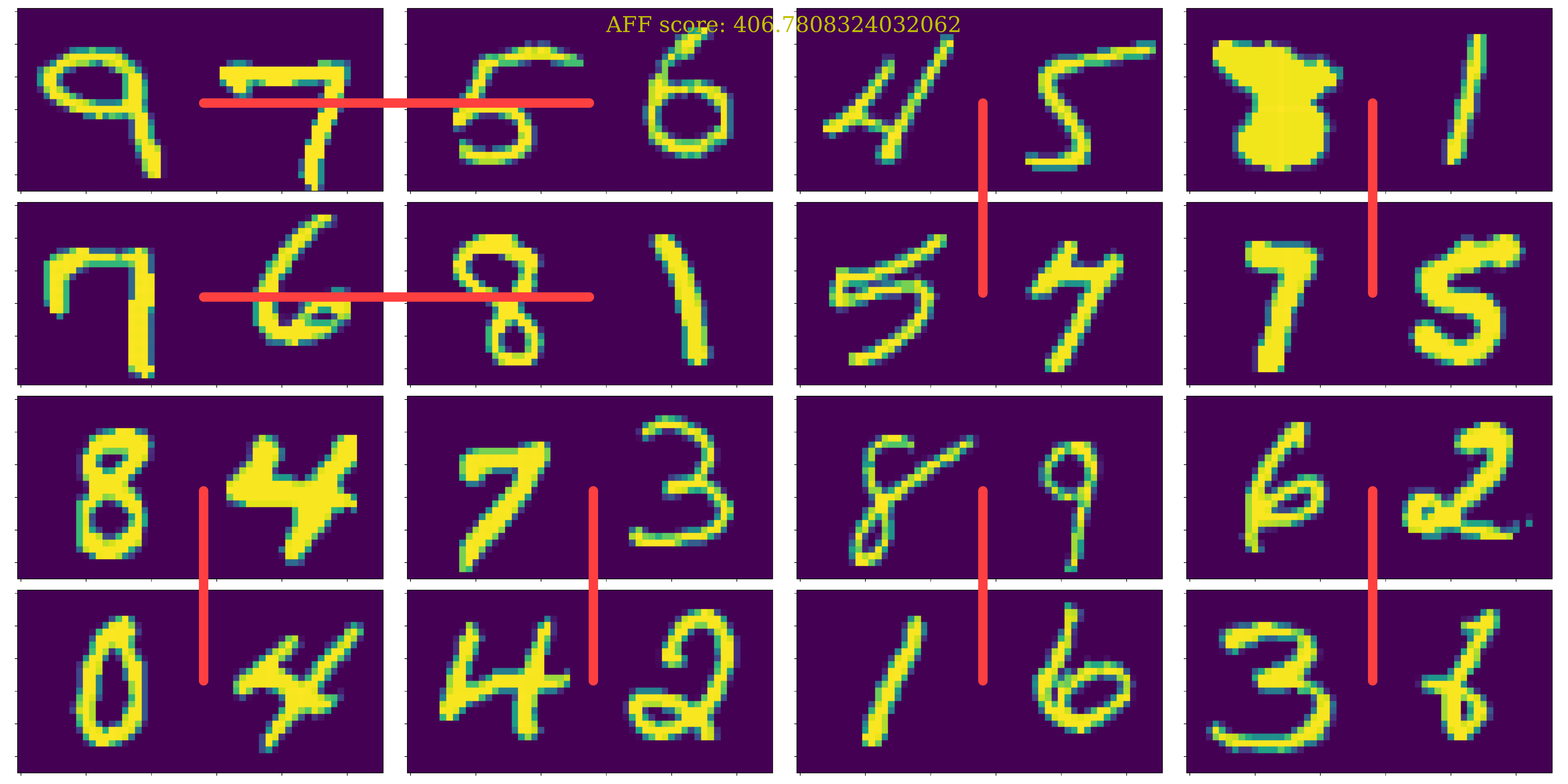}
   \vspace{-6mm}
   \caption{ Matching:AFF (Ours)} \label{fig:application1_apd7}
\end{subfigure}
\hspace*{\fill}
\begin{subfigure}{0.235\textwidth}
   \includegraphics[width=\linewidth]{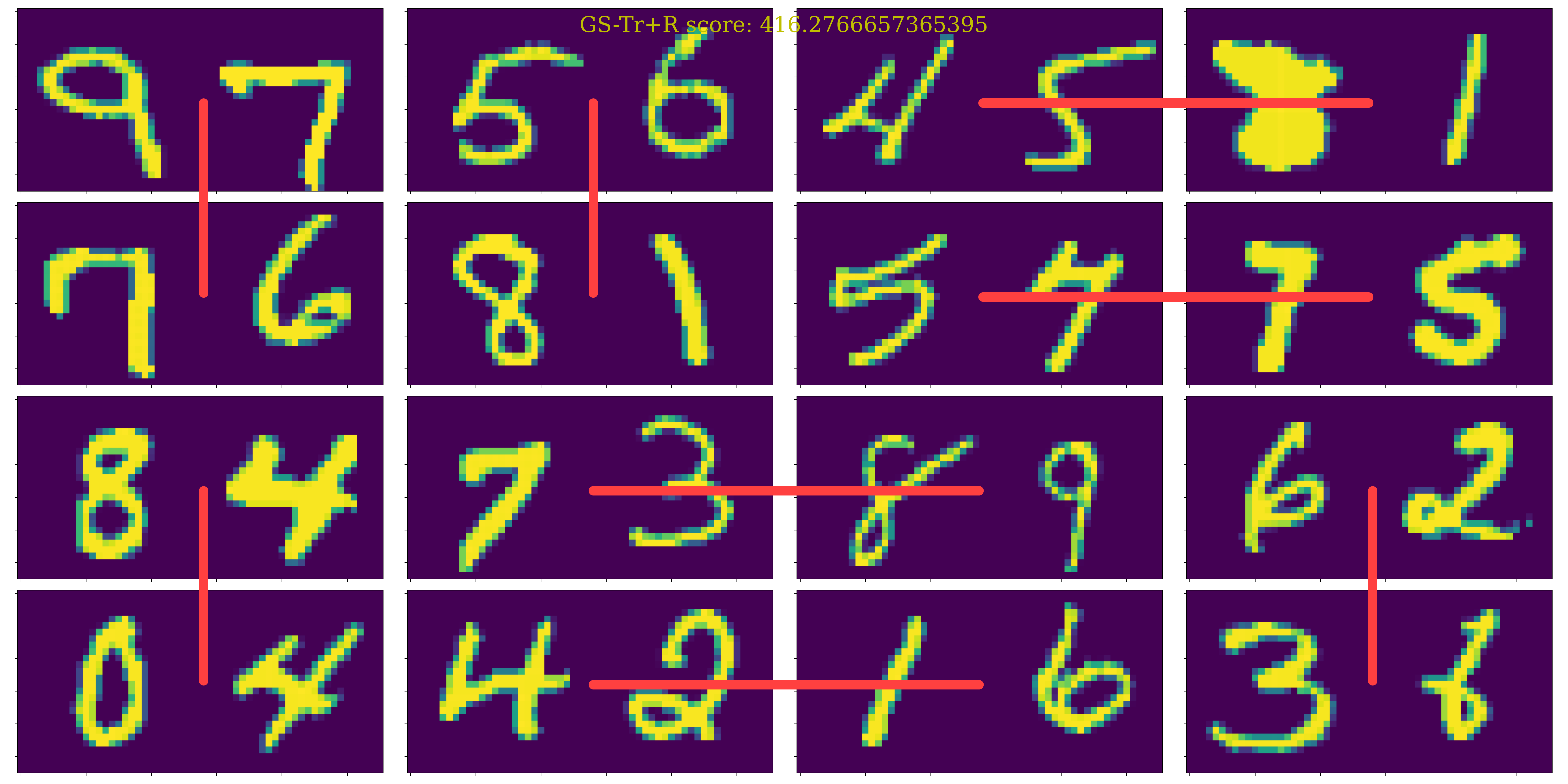}
   \vspace{-6mm}
   \caption{ Matching:GS-Tr + R} \label{fig:application1_apd8}
\end{subfigure}
\begin{subfigure}{0.235\textwidth}
   \includegraphics[width=\linewidth]{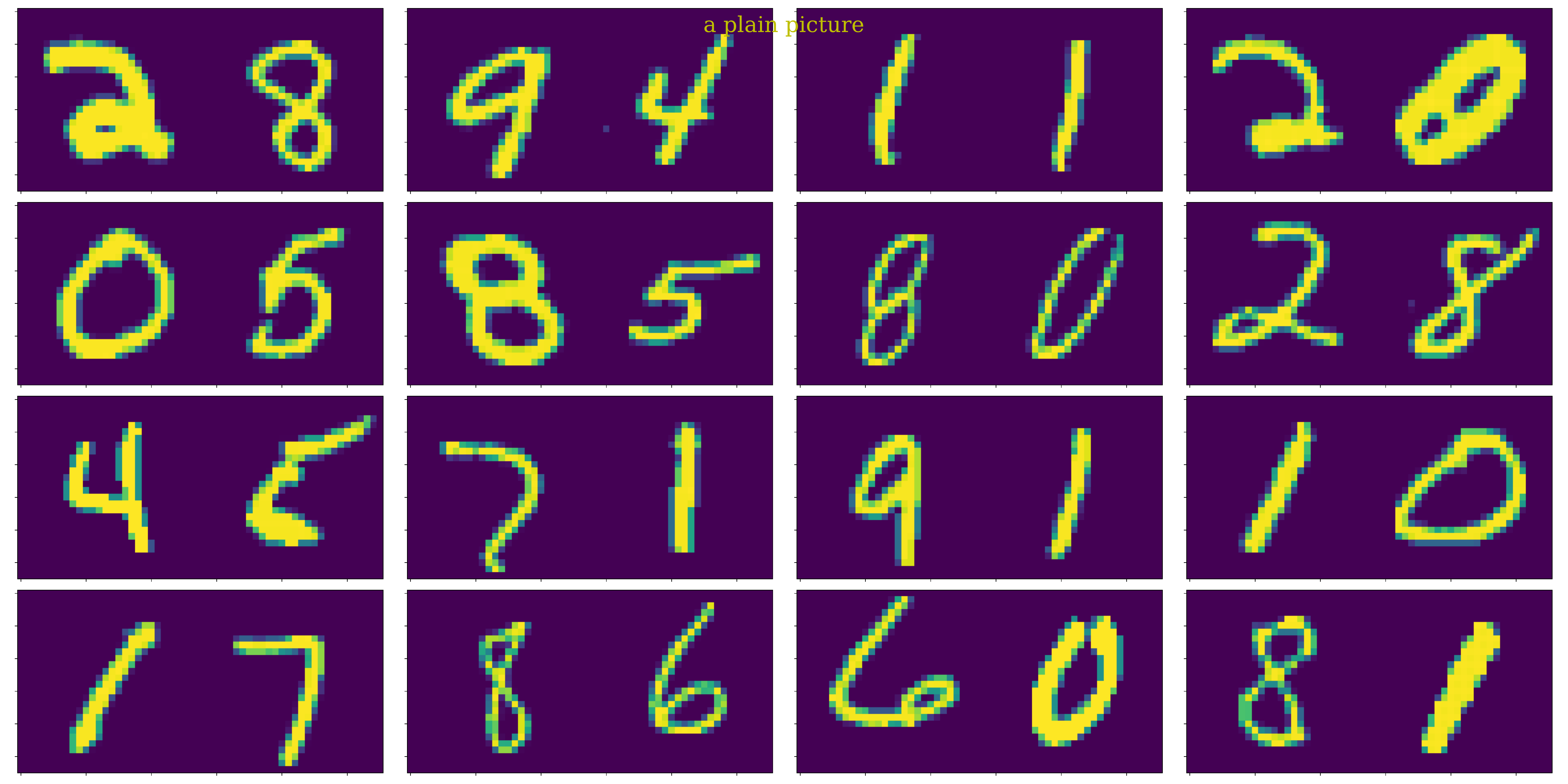}
   \vspace{-6mm}
   \caption{ Covering:Config} \label{fig:application1_apd9}
\end{subfigure}
\hspace*{\fill}
\begin{subfigure}{0.235\textwidth}
   \includegraphics[width=\linewidth]{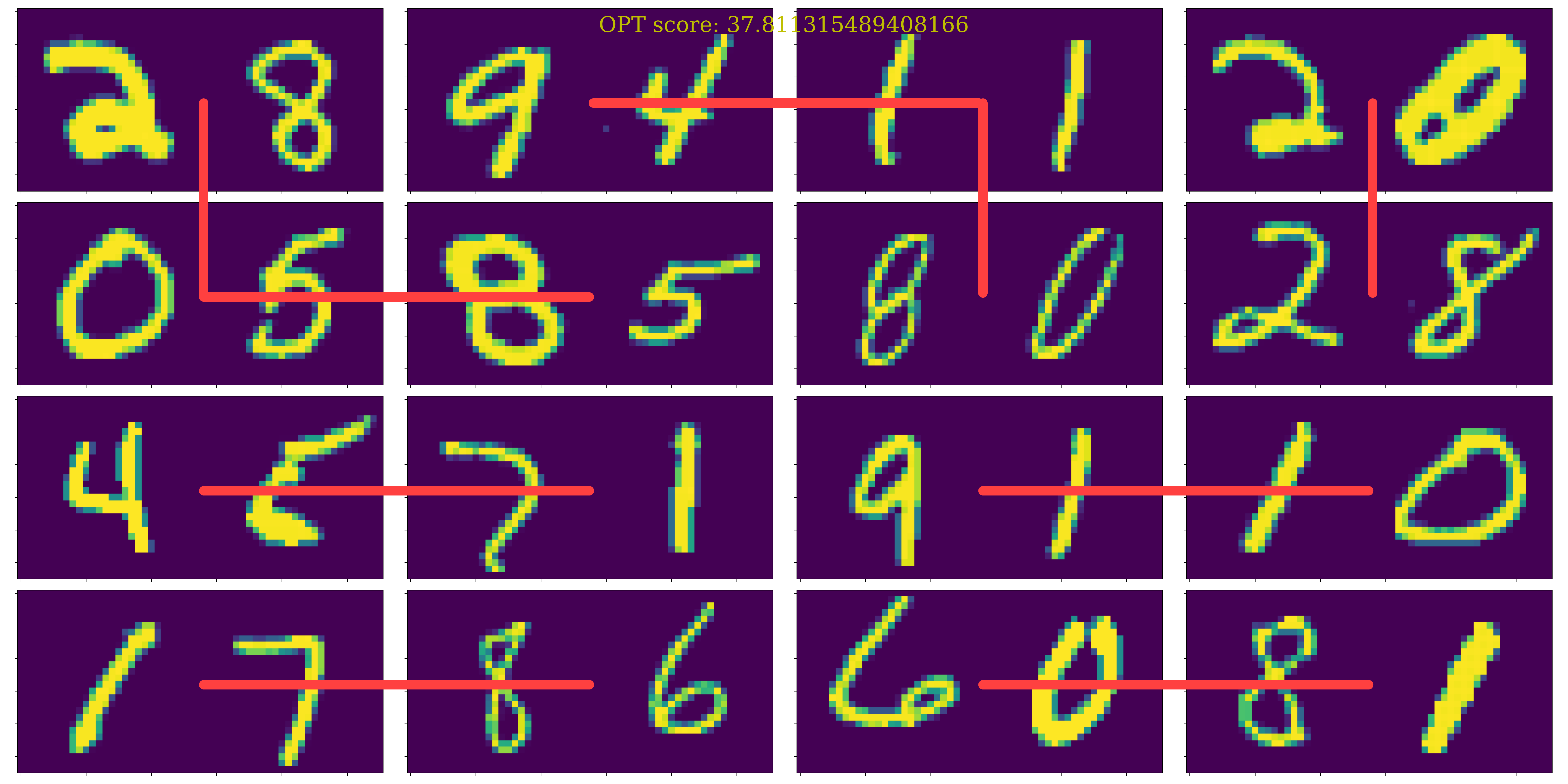}
   \vspace{-6mm}
   \caption{ Covering:Optimal $X^*$} \label{fig:application1_apd10}
\end{subfigure}
\hspace*{\fill}
\begin{subfigure}{0.235\textwidth}
   \includegraphics[width=\linewidth]{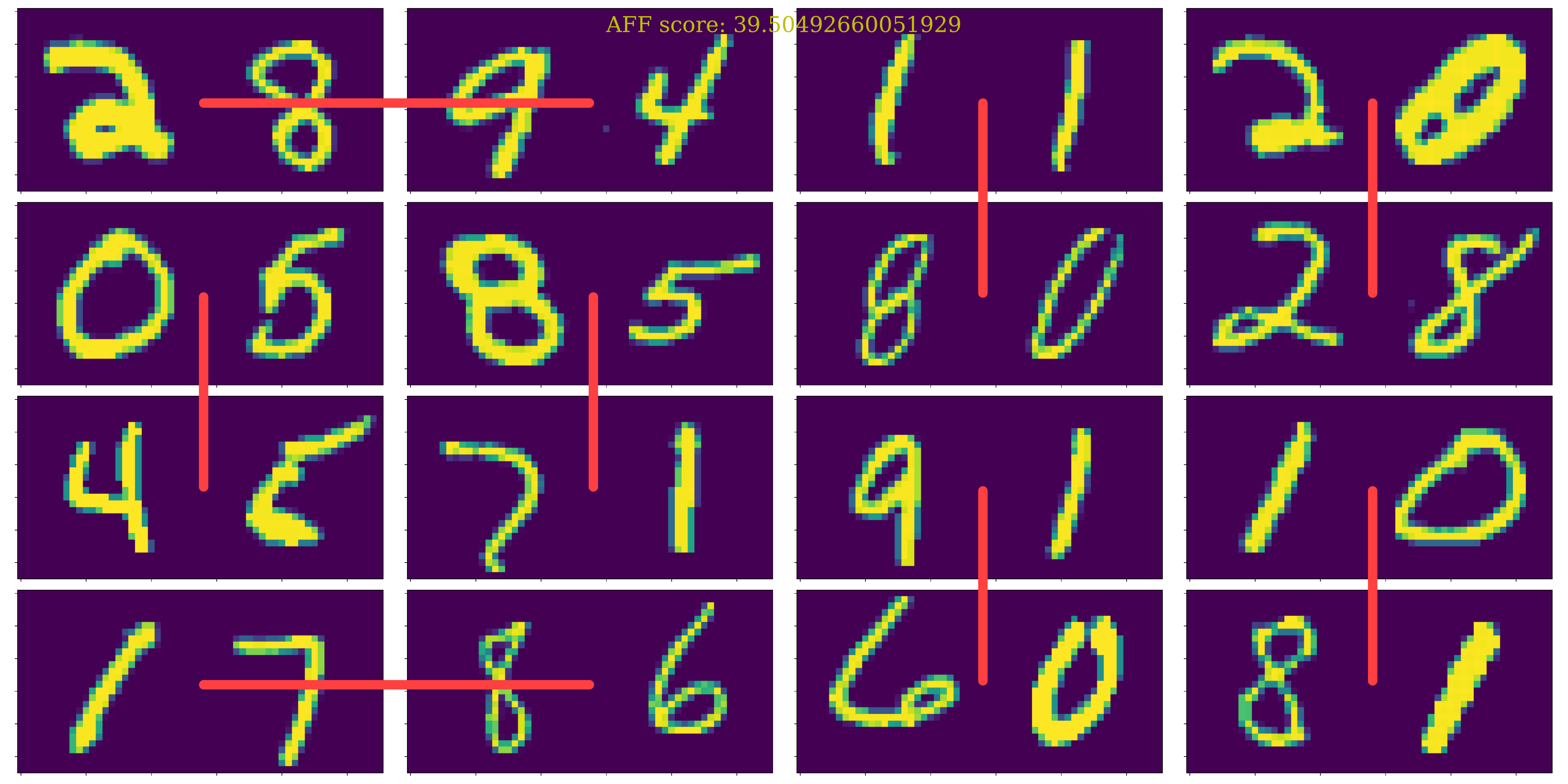}
   \vspace{-6mm}
   \caption{ Covering:AFF (Ours)} \label{fig:application1_apd11}
\end{subfigure}
\hspace*{\fill}
\begin{subfigure}{0.235\textwidth}
   \includegraphics[width=\linewidth]{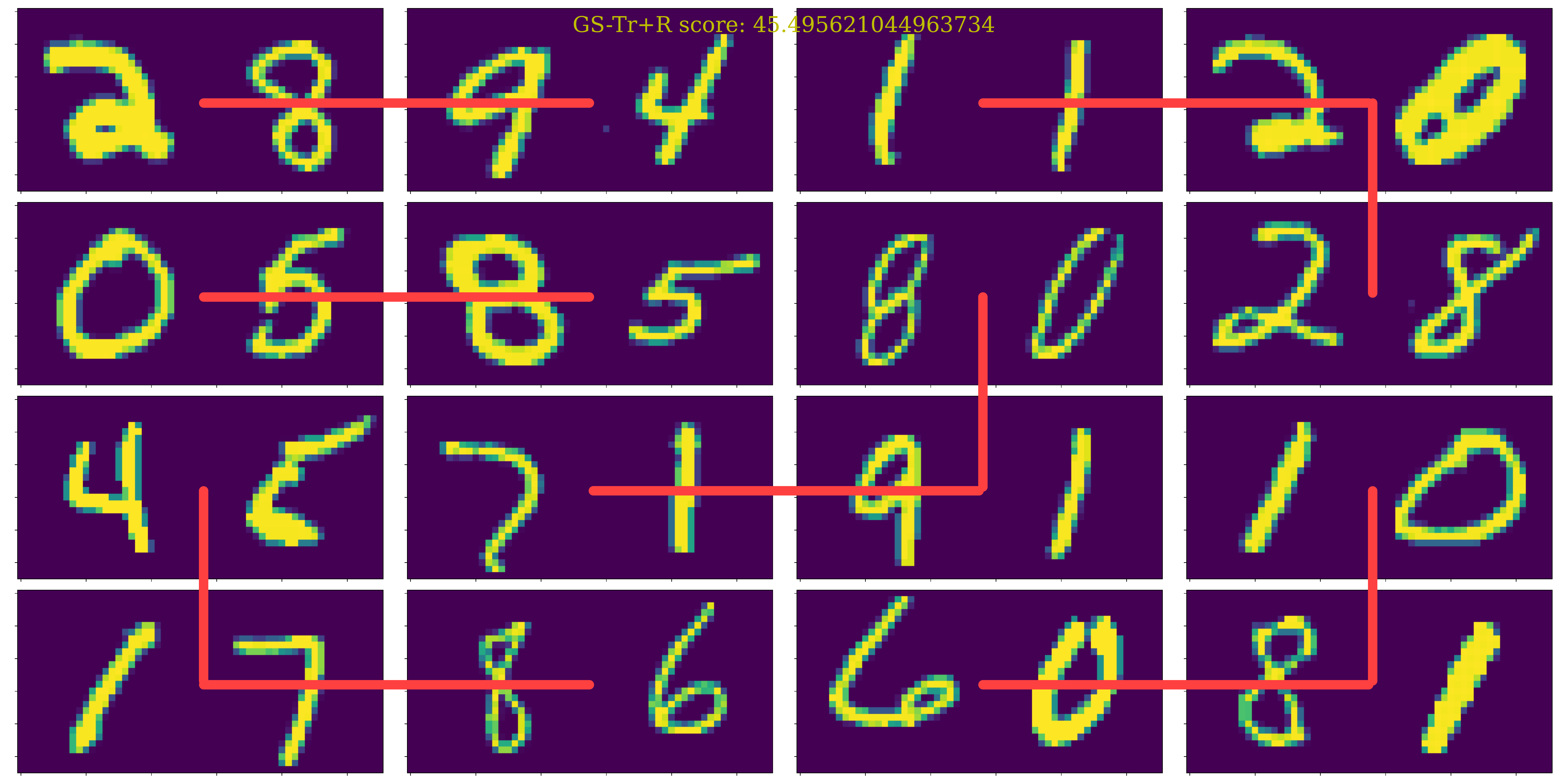}
   \vspace{-6mm}
   \caption{ Covering:GS-Tr + R} \label{fig:application1_apd12}
\end{subfigure}
\begin{subfigure}{0.235\textwidth}
   \includegraphics[width=\linewidth]{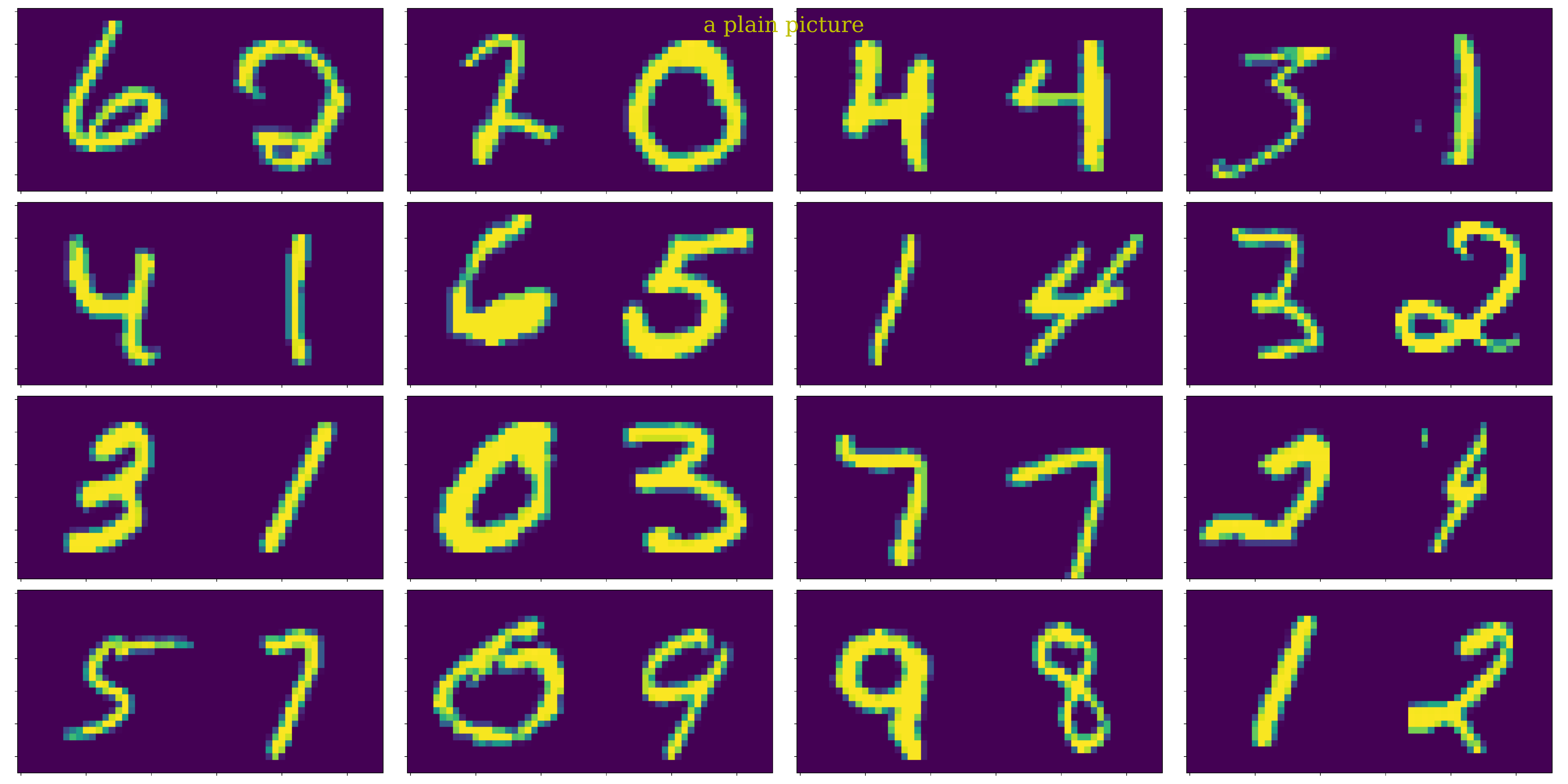}
   \vspace{-6mm}
   \caption{ Covering:Config} \label{fig:application1_apd13}
\end{subfigure}
\hspace*{\fill}
\begin{subfigure}{0.235\textwidth}
   \includegraphics[width=\linewidth]{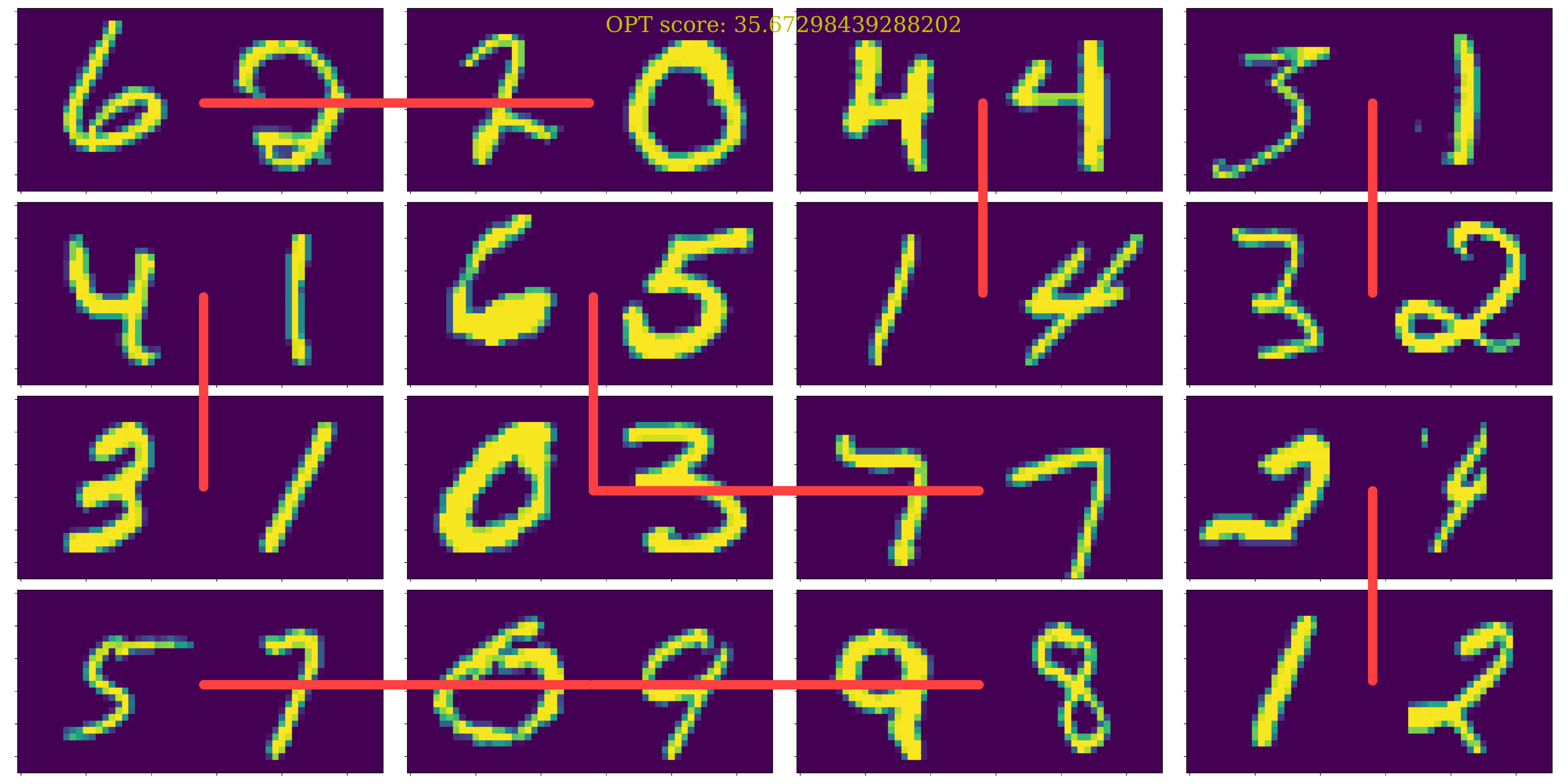}
   \vspace{-6mm}
   \caption{ Covering:Optimal $X^*$} \label{fig:application1_apd14}
\end{subfigure}
\hspace*{\fill}
\begin{subfigure}{0.235\textwidth}
   \includegraphics[width=\linewidth]{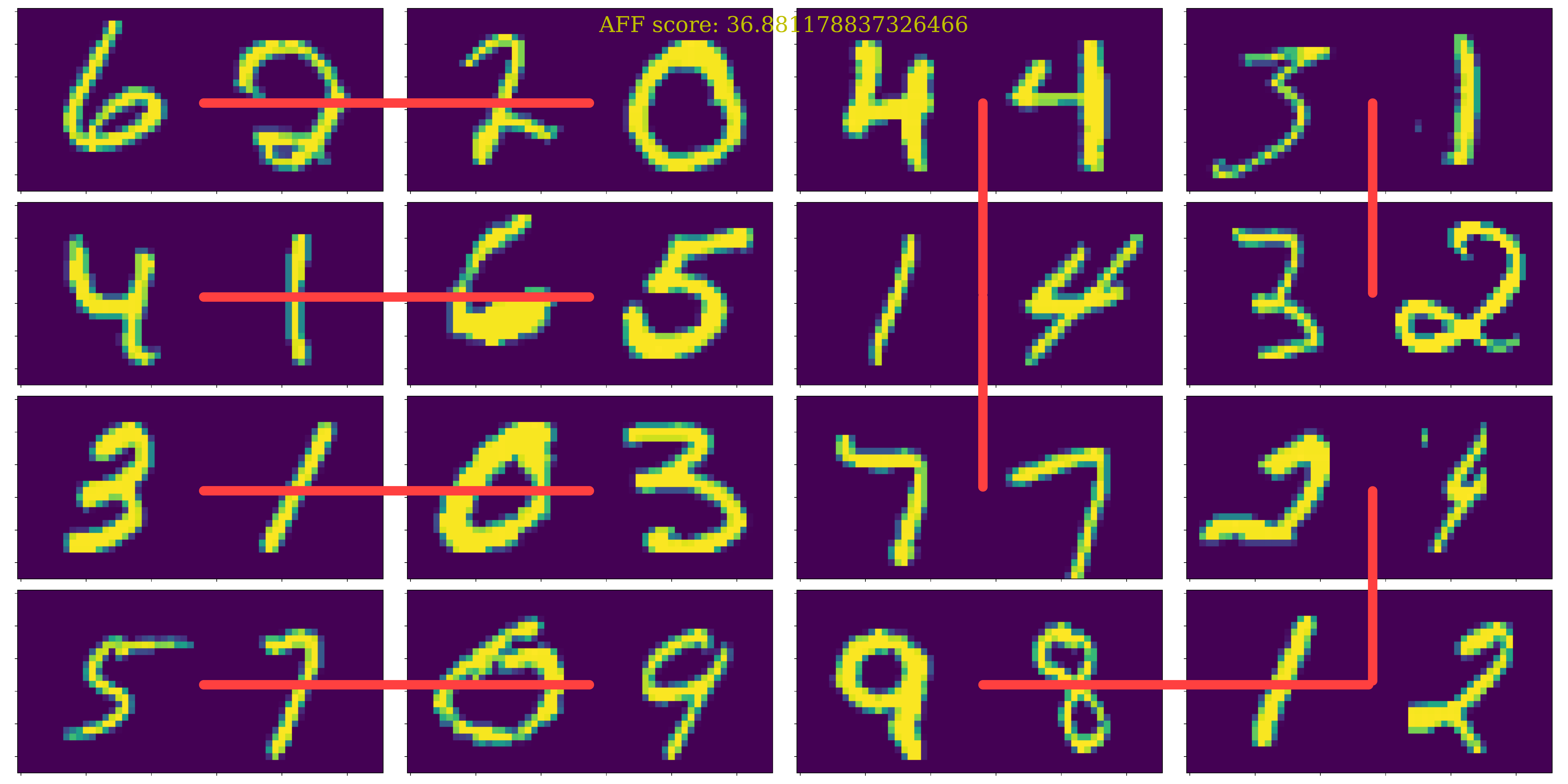}
   \vspace{-6mm}
   \caption{ Covering:AFF (Ours)} \label{fig:application1_apd15}
\end{subfigure}
\hspace*{\fill}
\begin{subfigure}{0.235\textwidth}
   \includegraphics[width=\linewidth]{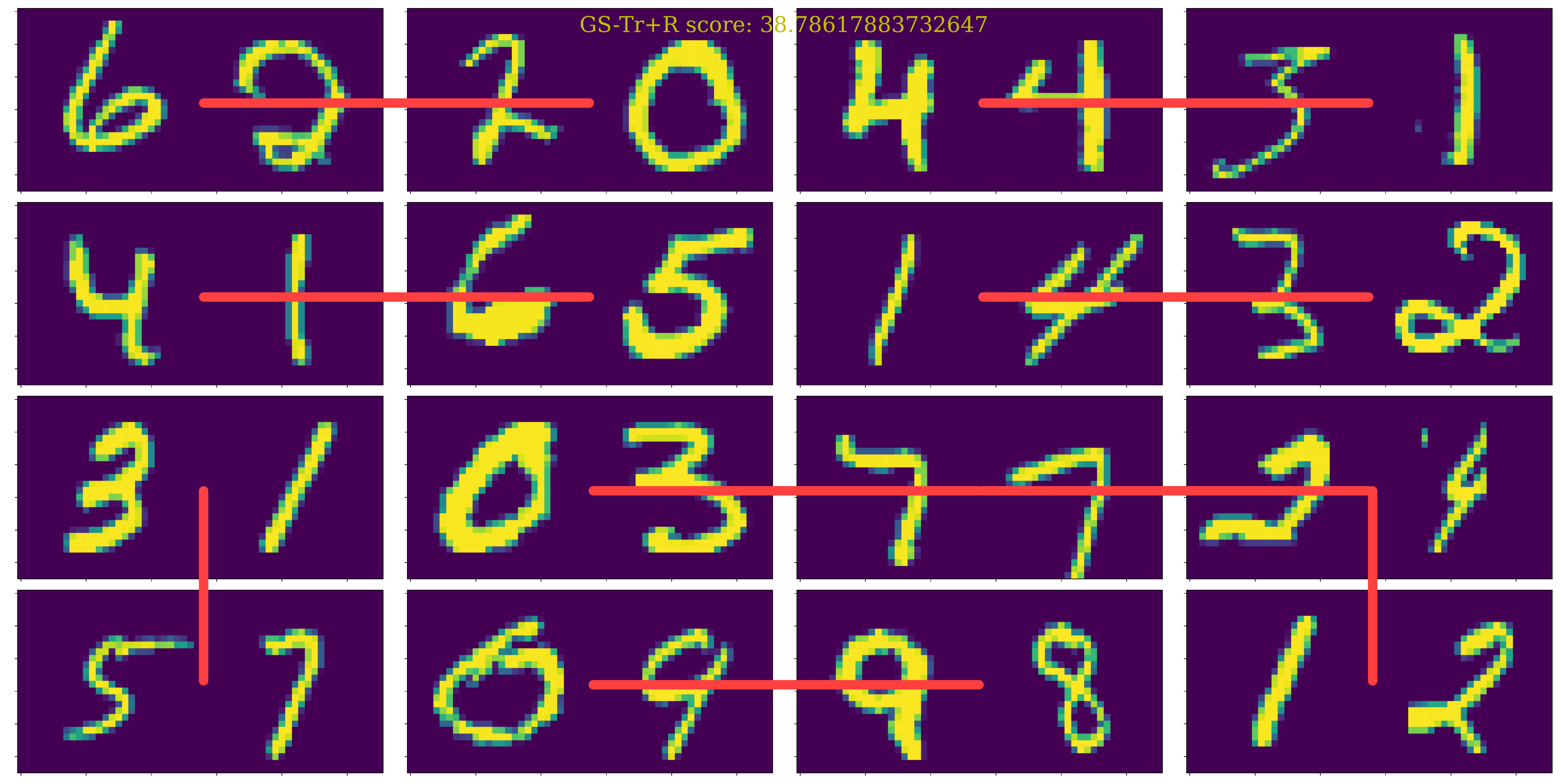}
   \vspace{-6mm}
   \caption{ Covering:GS-Tr + R} \label{fig:application1_apd16}
\end{subfigure}
\vspace{-2mm}
\caption{The additional visualization of the edge covering and node matching for Application I.}
\label{fig:application_1_apd}
\end{figure*}

Here, we display some additional visualization results of the feature-based edge covering and node matching problems in application I in Fig.~\ref{fig:application_1_apd}. In both edge covering and node matching problems, our method based on entry-wise affine proxies could avoid the multiplication between some large numbers so that the final cost could be low enough. For example, our method avoids 85*85 in the first row, 97*76 in the second row, 71*91 in the third row, and 69*98 in the forth row, which are all selected by the method with Gumbel-Softmax tricks.  
\subsection{Application II}
\begin{figure*}[t]
\centering
\begin{subfigure}{0.33\textwidth}
   \includegraphics[width=\linewidth]{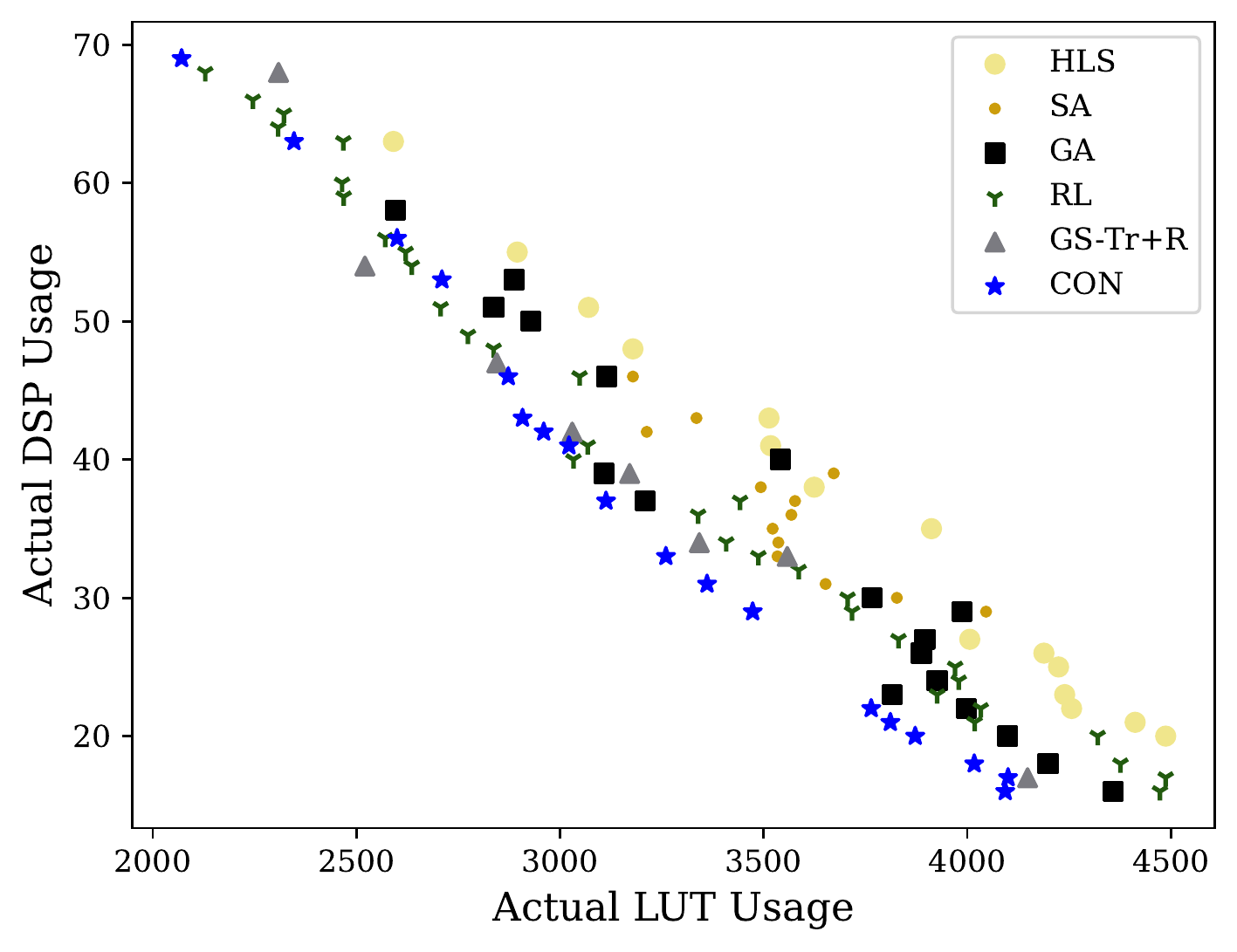}
\end{subfigure}
\hspace*{-1mm}
\begin{subfigure}{0.33\textwidth}
   \includegraphics[width=\linewidth]{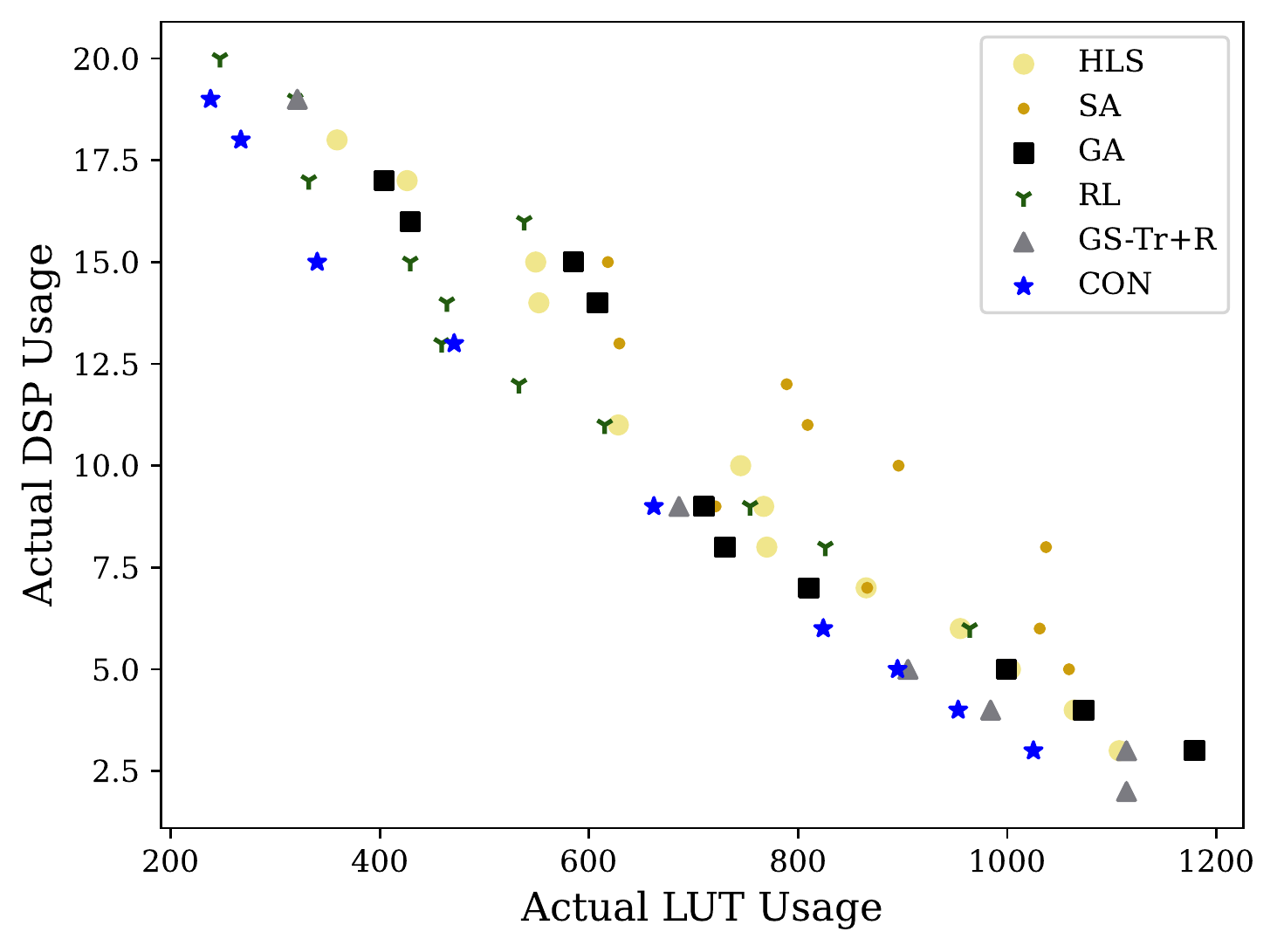}
\end{subfigure}
\hspace*{-1mm}
\begin{subfigure}{0.33\textwidth}
   \includegraphics[width=\linewidth]{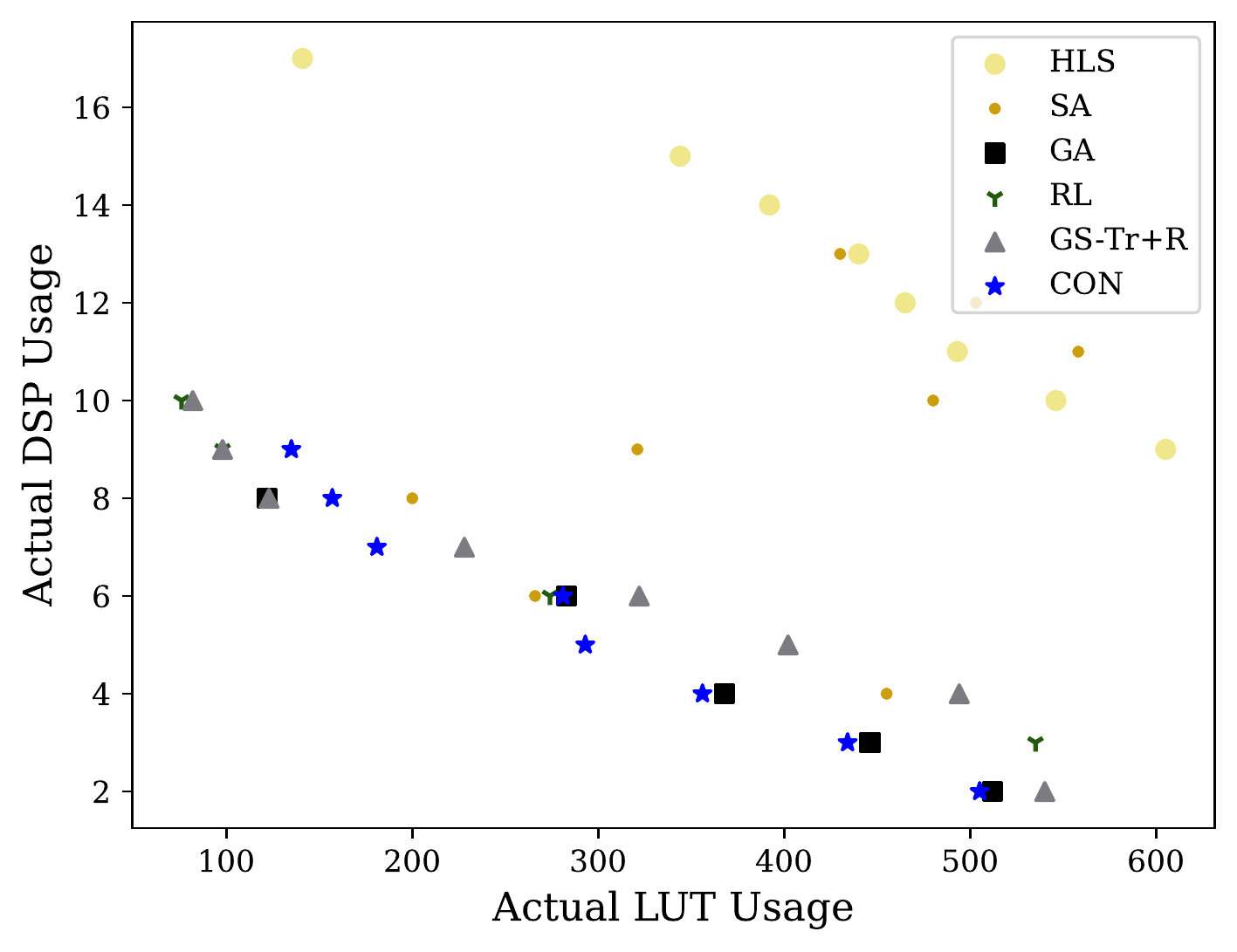}
\end{subfigure}
\vspace{-2mm}
\begin{subfigure}{0.33\textwidth}
   \includegraphics[width=\linewidth]{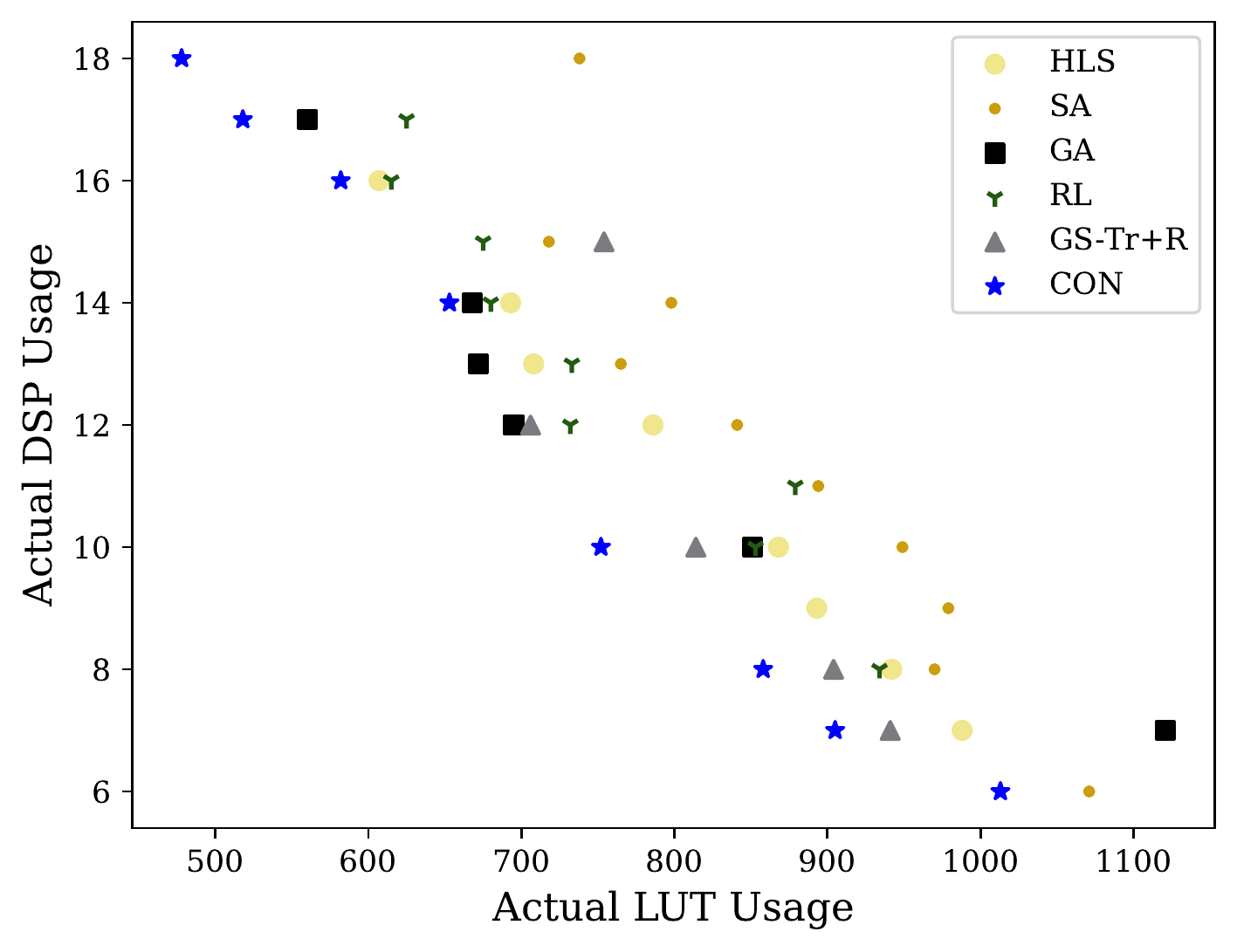}
\end{subfigure}
\hspace*{-1mm}
\begin{subfigure}{0.33\textwidth}
   \includegraphics[width=\linewidth]{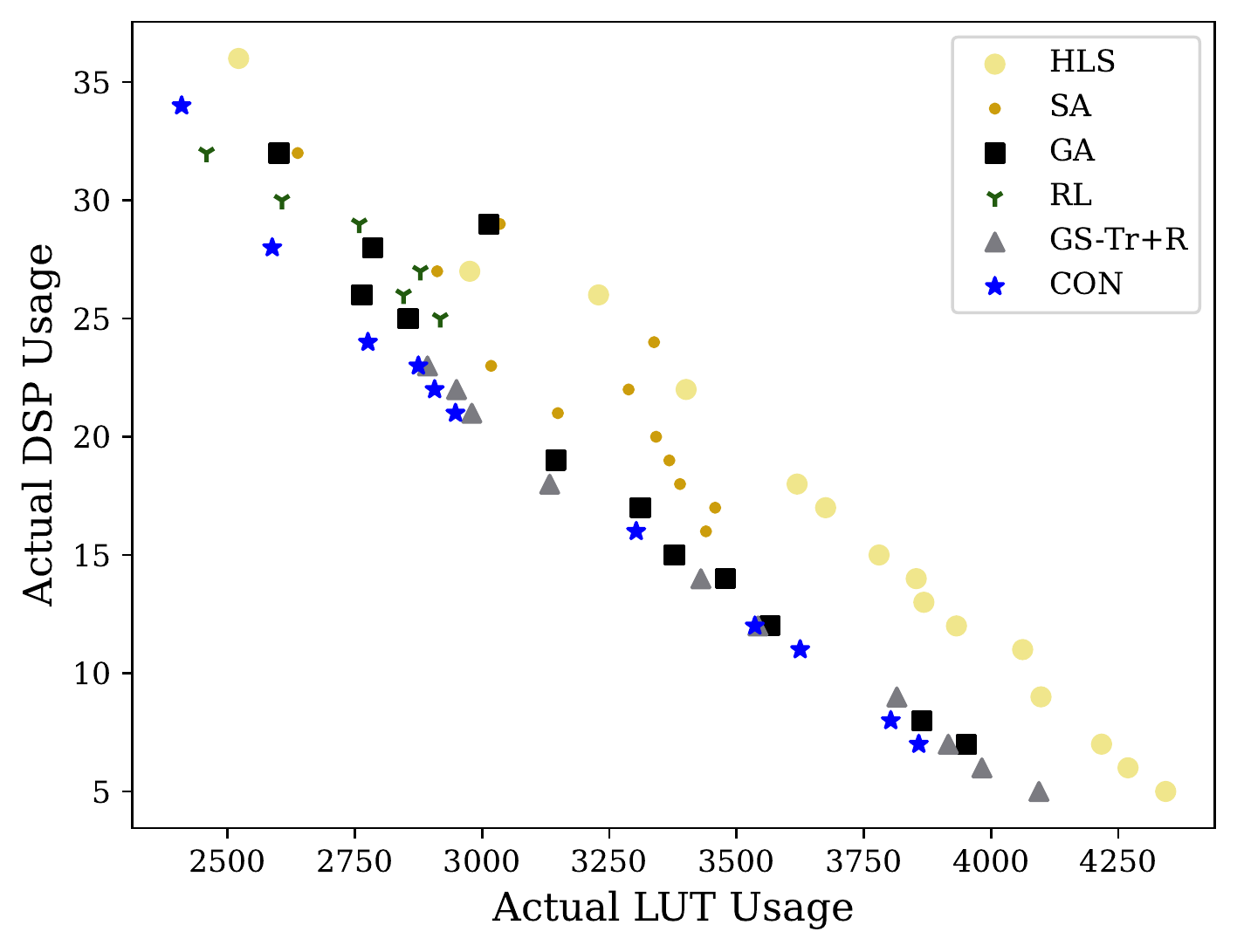}
\end{subfigure}
\hspace*{-1mm}
\begin{subfigure}{0.33\textwidth}
   \includegraphics[width=\linewidth]{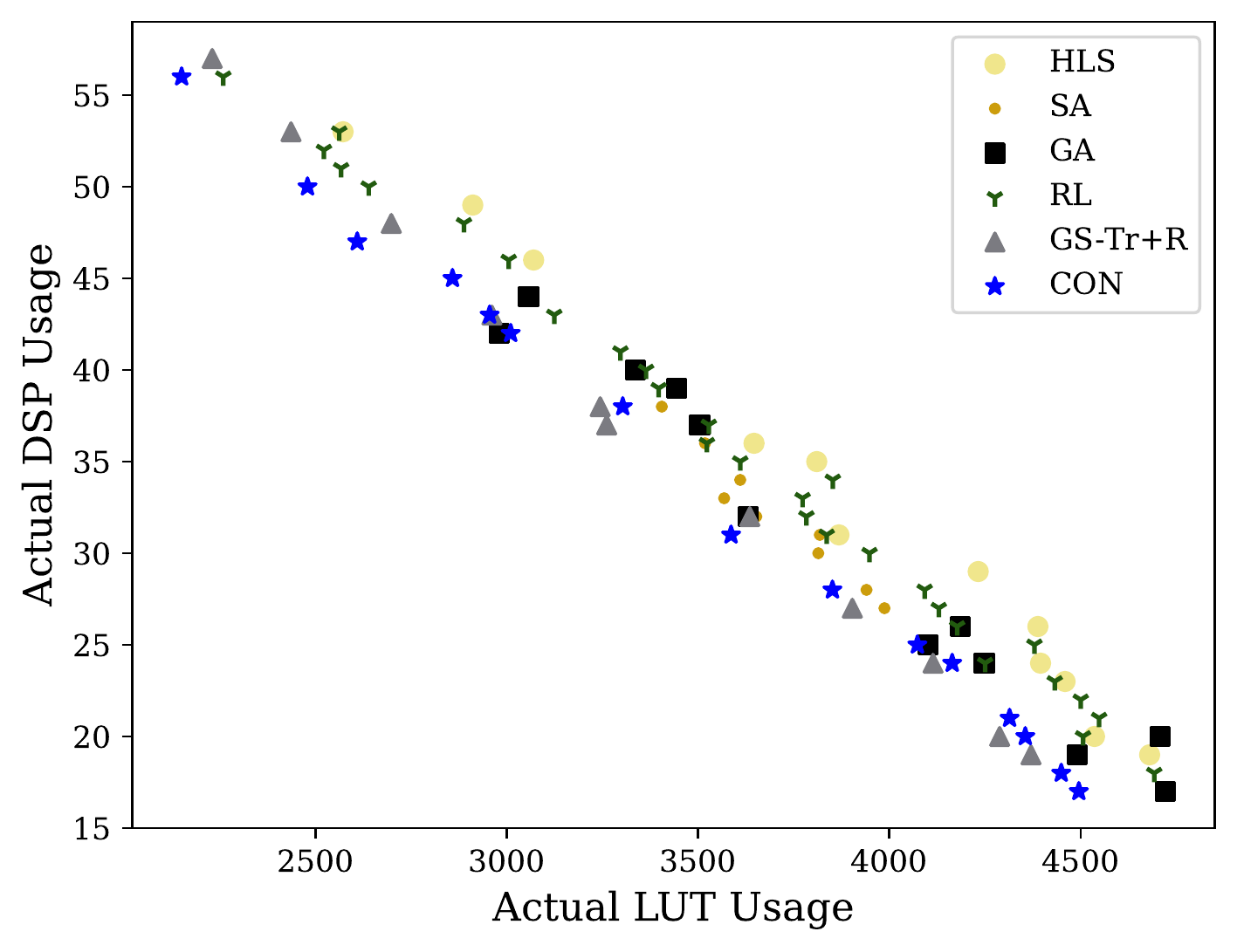}
\end{subfigure}
\caption{The additional visualization of LUT-DSP usage relationship. The HLS baseline denotes the optimal synthesis results among 200 random mappings.}
\label{fig:application_2_apd}
\vspace{-0.2cm}
\end{figure*}

\begin{wraptable}{r}{0.45 \textwidth}
\centering
\begin{tabular}{@{}cccc@{}}
\toprule
Method  & 1-st (\%) & 2-nd (\%) & 3-rd (\%) \\ \midrule
SA      & 12.3      & 10.3      & 13.4      \\
GA      & 14.8      & 15.7      & 18.2      \\
RL      & 18.5      & 21.8      & 19.5      \\
GS-Tr+R & 20.9      & \textbf{27.2}      & 22.7      \\
CON     & \textbf{39.5}      & 24.8      & \textbf{26.0}      \\ \bottomrule
\end{tabular}
\caption{The percentage that each method occupies in the first, second and third place of the ranking.}
\vspace{-0.5cm}
\label{tab:application_2_ranking_2}
\end{wraptable}
The additional visualization results of DSP-LUT usage amount relationship on the test configurations is shown in Figure.~\ref{fig:application_2_apd}. Our entry-concave proxies generates the lowest LUT-DSP combinations among all the methods. To be fair, we also pick the best results from $200$ randomly sampled HLS tool's assignment as a baseline (called HLS), to show the superiority of the other optimization methods. The GS-Tr+R and the RL method outperforms the HLS baseline, while the generic methods SA and GA only marginally outperform and are comparable with the best of HLS random solutions.

The average ranking of the LUT usage under the constraint of the maximum DSP usage amount is shown in Table.~\ref{tab:application_2_value_apd}, which adds to another two baselines, the `Naive' baseline and the GS-Tr+S method. It turns out that these two methods could not generate feasible results that satisfy the DSP usage threshold when the threshold is relatively low, thus we put them in the last place in the ranking if they could not generate feasible solutions.

\begin{table}[t]
\centering
\begin{tabular}{@{}clllllllc@{}}
\toprule
DSP usage &
  \multicolumn{1}{c}{40\%} &
  \multicolumn{1}{c}{45\%} &
  \multicolumn{1}{c}{50\%} &
  \multicolumn{1}{c}{55\%} &
  \multicolumn{1}{c}{60\%} &
  \multicolumn{1}{c}{65\%} &
  \multicolumn{1}{c}{70\%} &
  rank-avg \\ \midrule
SA      & 4.25 & 4.16 & 4.08 & 4.08 & 4.25 & 4.75 & 4.66 & 4.31 \\
GA      & 3.50 & 3.75 & 3.33 & 3.50 & 3.50 & 3.50 & 3.25 & 3.47 \\
Na\"{i}ve   & 5.75 & 6.16 & 6.41 & 6.00 & 6.25 & 6.33 & 6.33 & 6.17 \\
RL      & 4.25 & 4.33 & 4.50 & 4.00 & 3.58 & 3.41 & \textbf{3.00} & 3.86 \\
GS-Tr+S & 4.33 & 3.41 & 3.58 & 3.66 & 3.75 & 3.83 & 3.83 & 3.77 \\
GS-Tr+R & 3.33 & 3.58 & 3.16 & 3.91 & \textbf{2.91} & 3.16 & 3.58 & 3.37 \\
CON     & \textbf{2.41} & \textbf{2.50} & \textbf{2.83} & \textbf{2.75} & 3.66 & \textbf{3.00} & 3.25 & \textbf{2.91} \\ \bottomrule
\end{tabular}
\caption{The result of LUT DSP balancing problem in application II. The DSP usage thresholds are from $40\%$ to $70\%$, For the Na\"{i}ve and GS-Tr+S method, if there are no feasible results under the DSP usage constraint, we put them in the last place.}
\label{tab:application_2_value_apd}
\end{table}

\begin{figure*}[t]
\centering
\begin{subfigure}{0.33\textwidth}
   \includegraphics[width=\linewidth]{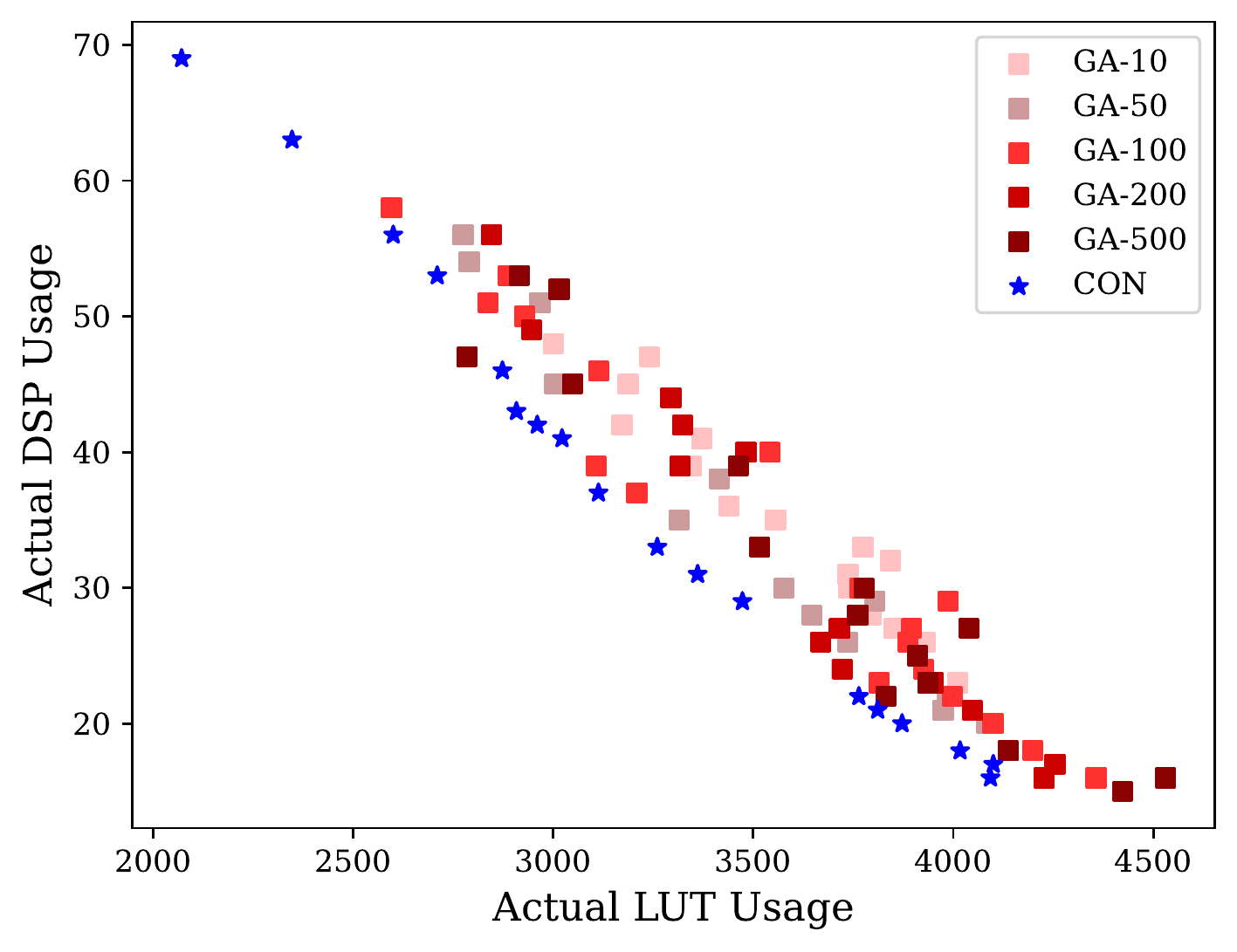}
\end{subfigure}
\hspace*{-1mm}
\begin{subfigure}{0.33\textwidth}
   \includegraphics[width=\linewidth]{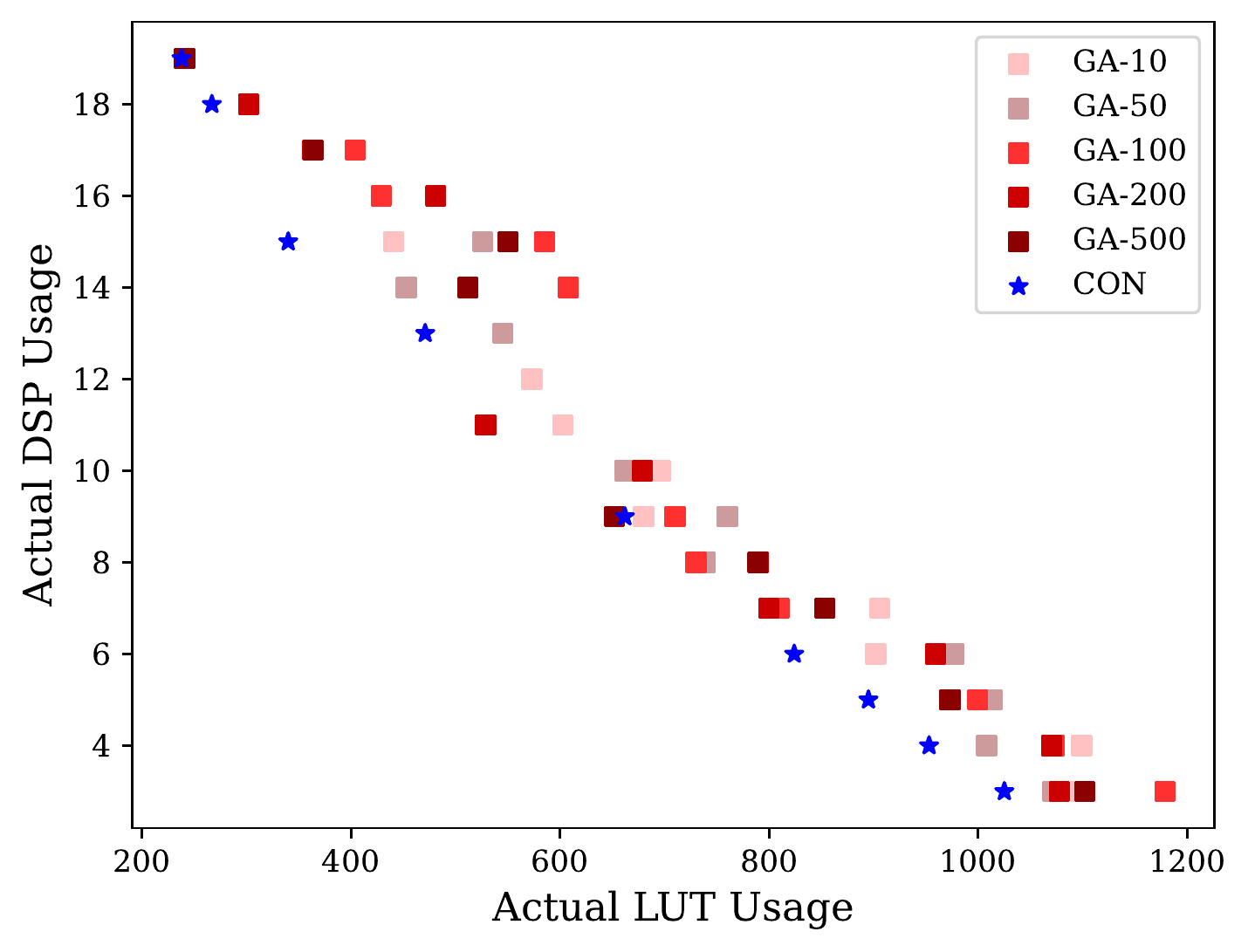}
\end{subfigure}
\hspace*{-1mm}
\begin{subfigure}{0.33\textwidth}
   \includegraphics[width=\linewidth]{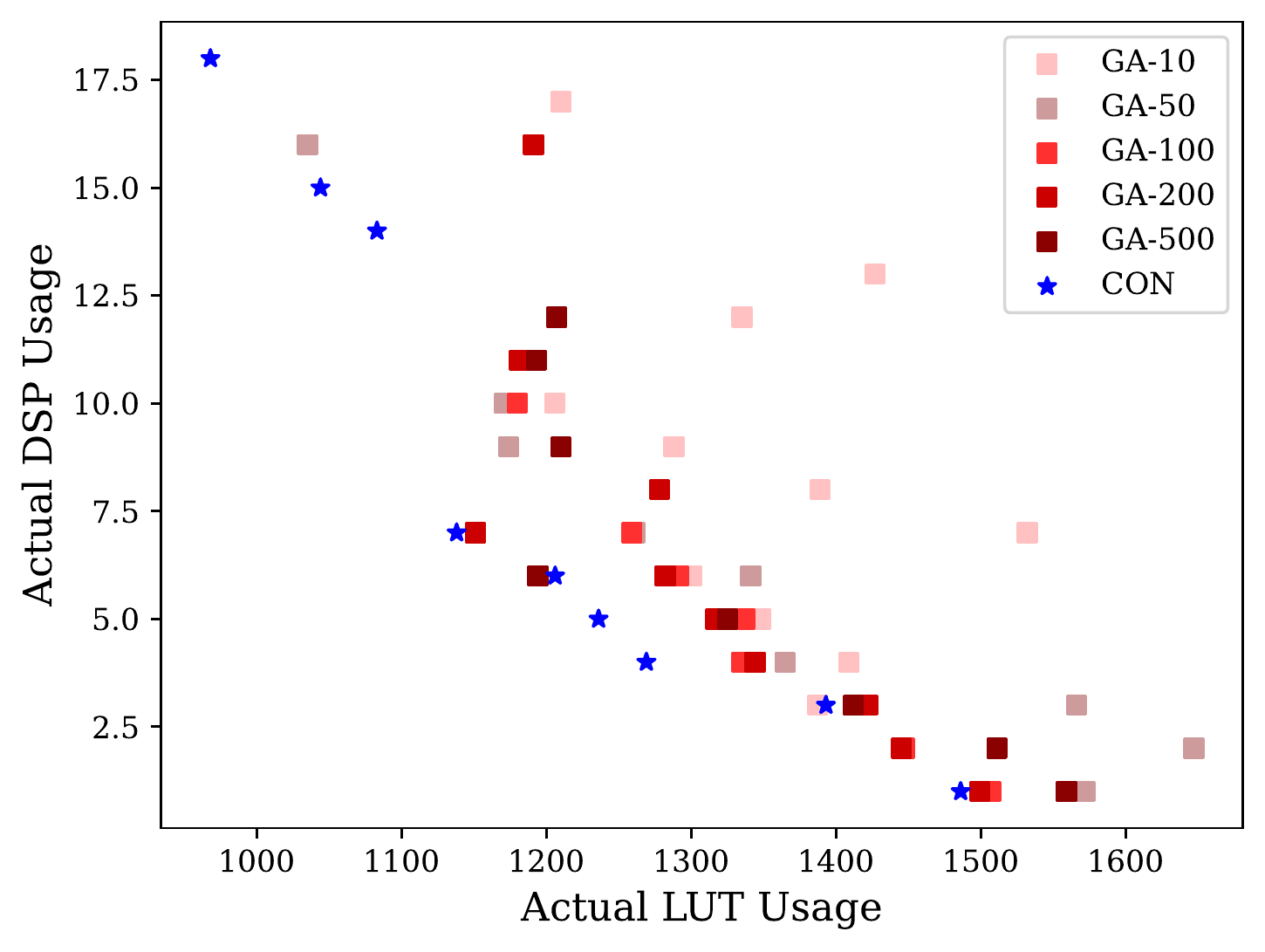}
\end{subfigure}
\vspace{-2mm}
\begin{subfigure}{0.33\textwidth}
   \includegraphics[width=\linewidth]{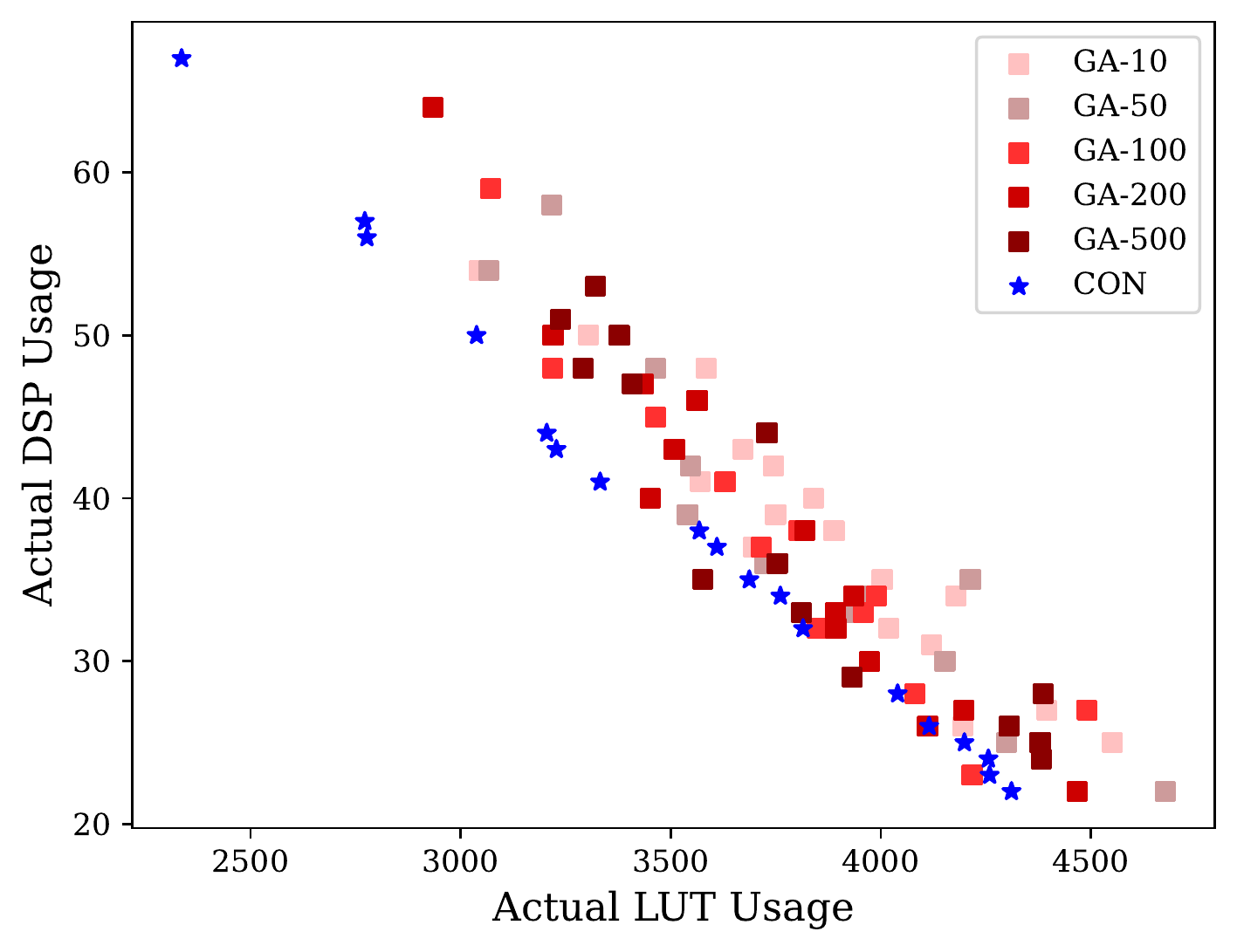}
\end{subfigure}
\hspace*{-1mm}
\begin{subfigure}{0.33\textwidth}
   \includegraphics[width=\linewidth]{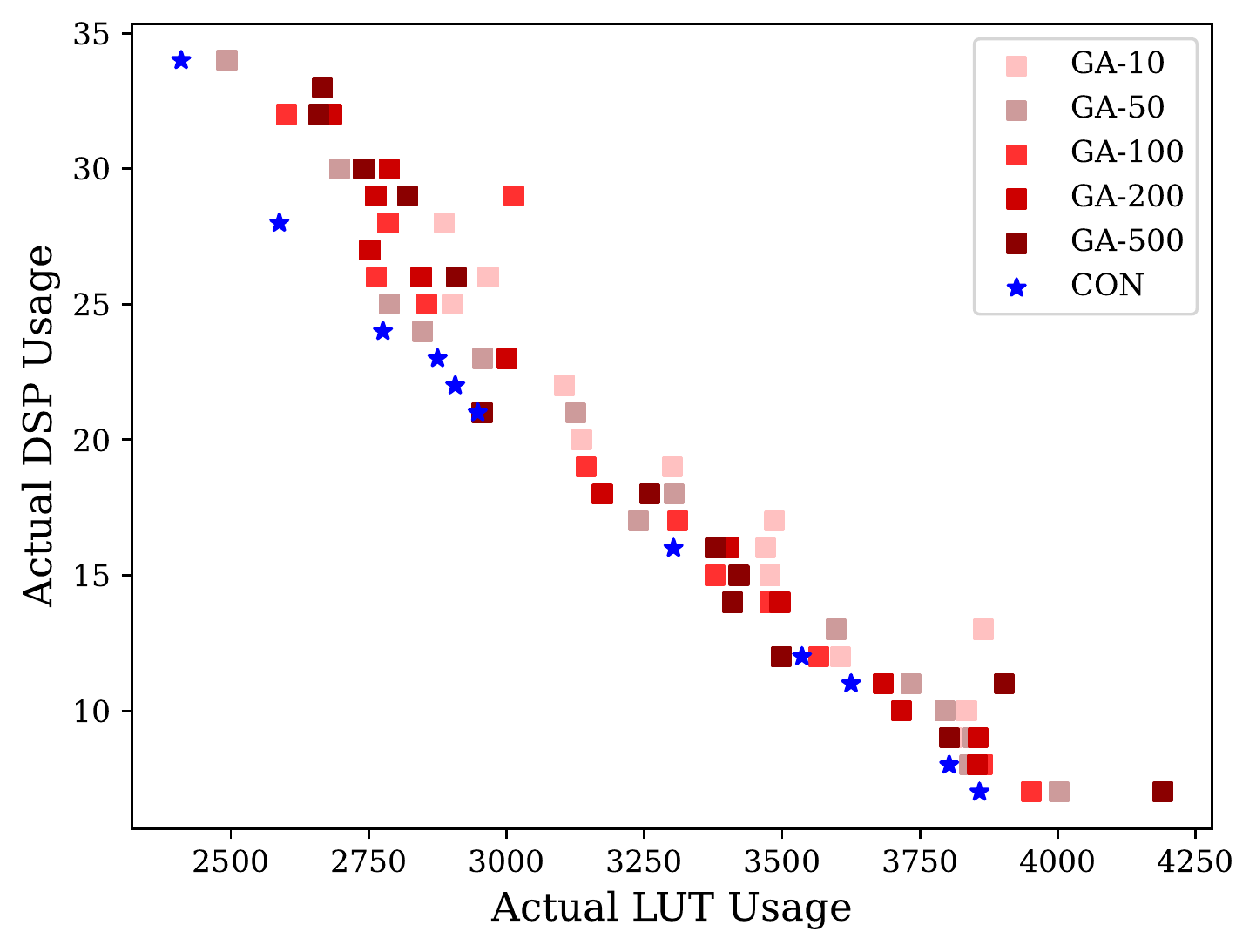}
\end{subfigure}
\hspace*{-1mm}
\begin{subfigure}{0.33\textwidth}
   \includegraphics[width=\linewidth]{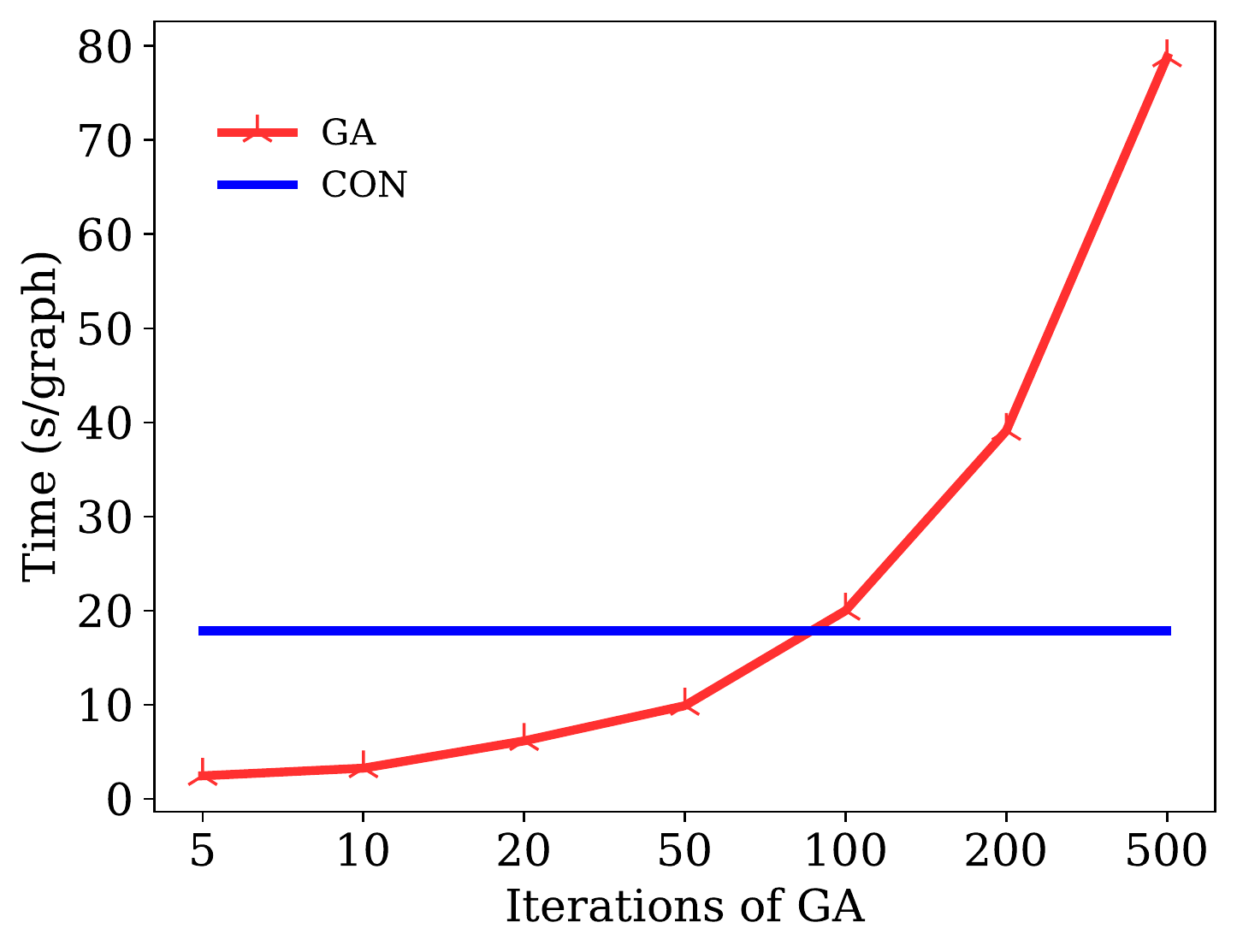}
\end{subfigure}
\caption{The LUT-DSP usage relationship of CON and GA with different iterations. Here GA runs in parallel on GPU with the population of $256$ which is the same as the batch size of our $\mathcal{A}_{\theta}$ training. We run GA in $5,10,20,50,100,200,500$ generations and show the inference time-iteration relation in the bottom right figure.}
\label{fig:application_2_apd_ga}
\vspace{-0.2cm}
\end{figure*}

\begin{figure*}[h]
\centering
\vspace{-2mm}
\begin{subfigure}{0.33\textwidth}
   \includegraphics[width=\linewidth]{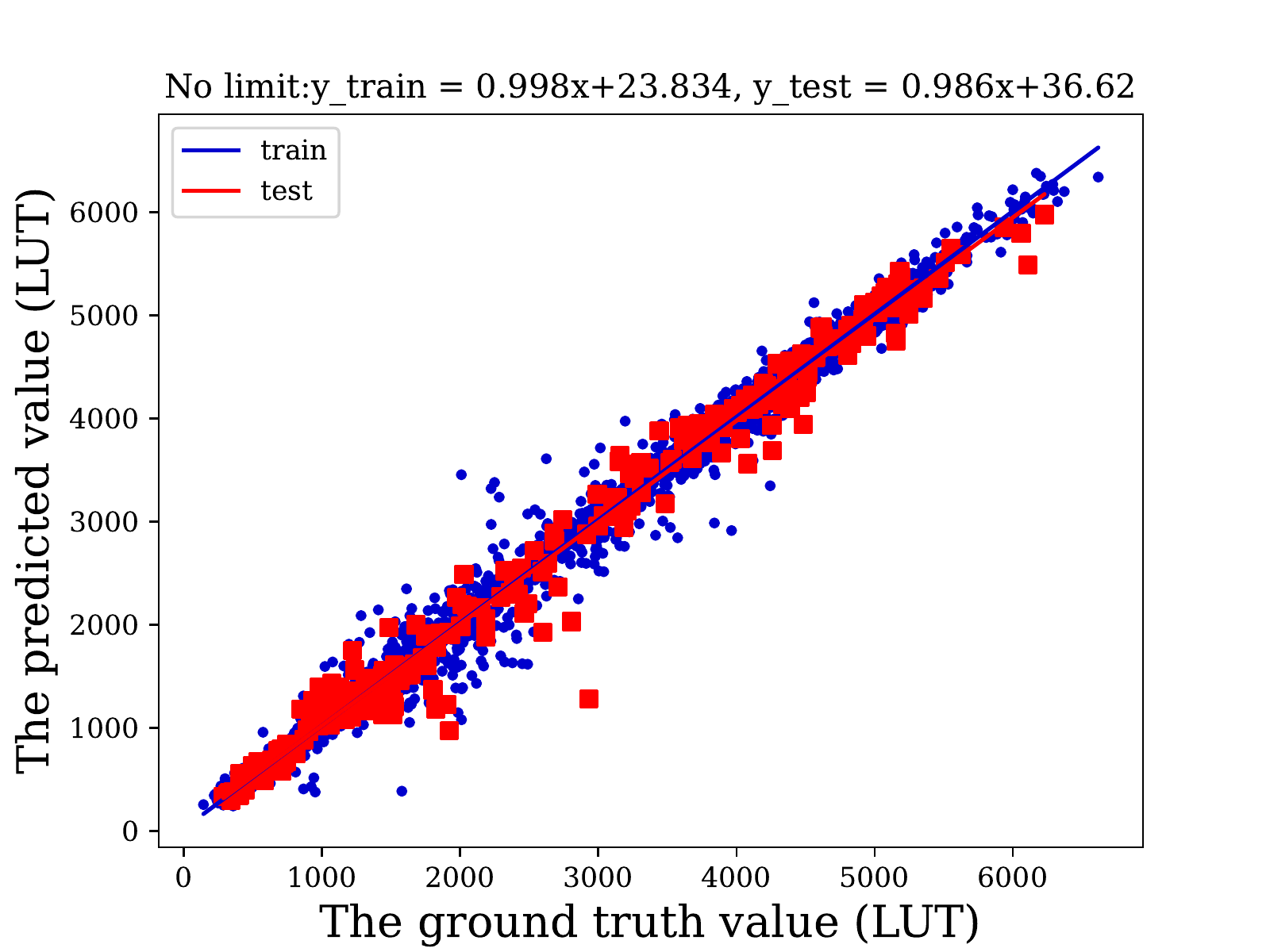}
   \caption{LUT proxy with no limits}
\end{subfigure}
\hspace*{-1mm}
\begin{subfigure}{0.33\textwidth}
   \includegraphics[width=\linewidth]{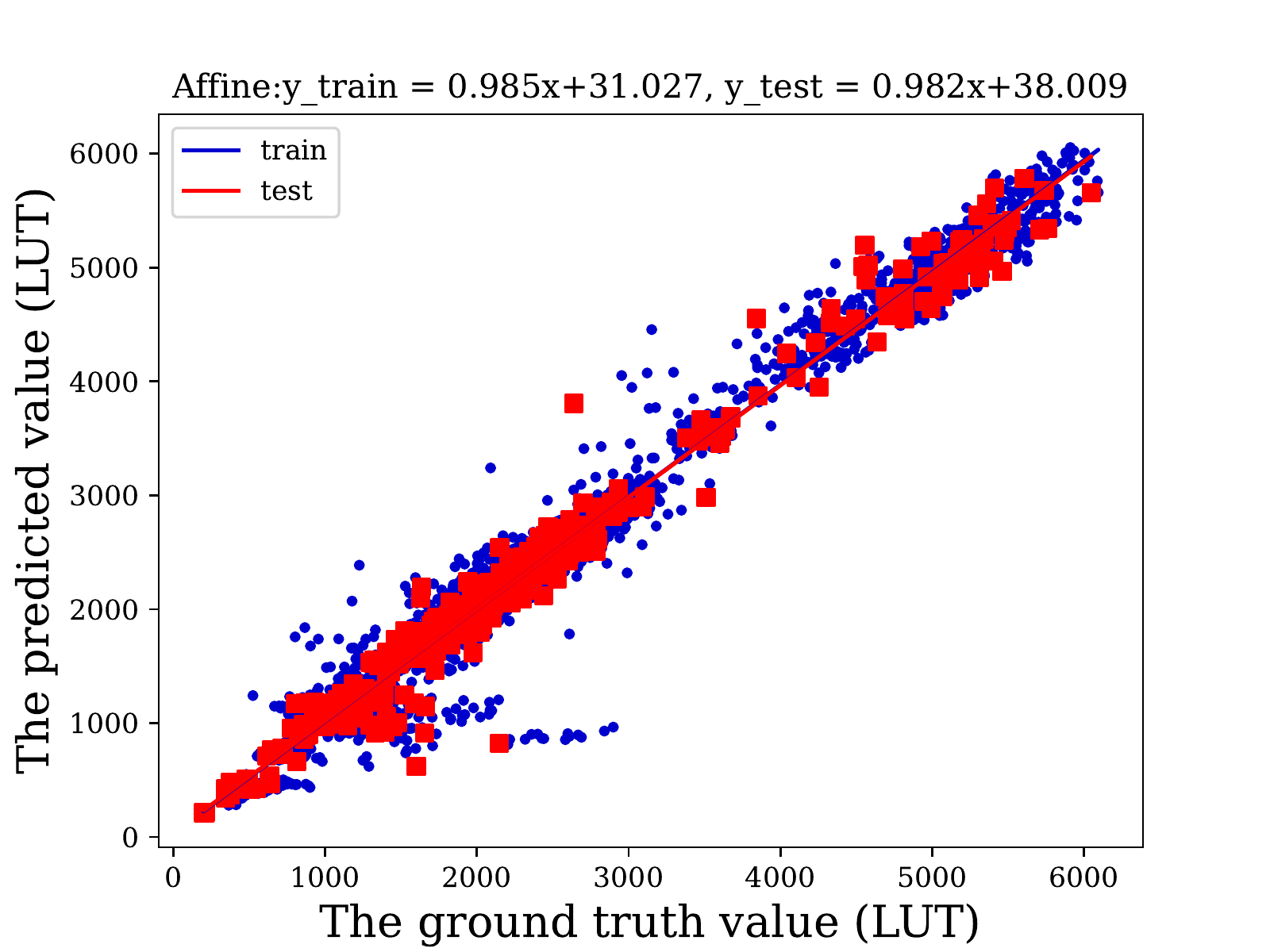}
   \caption{LUT entry-wise affine proxy}
\end{subfigure}
\hspace*{-1mm}
\begin{subfigure}{0.33\textwidth}
   \includegraphics[width=\linewidth]{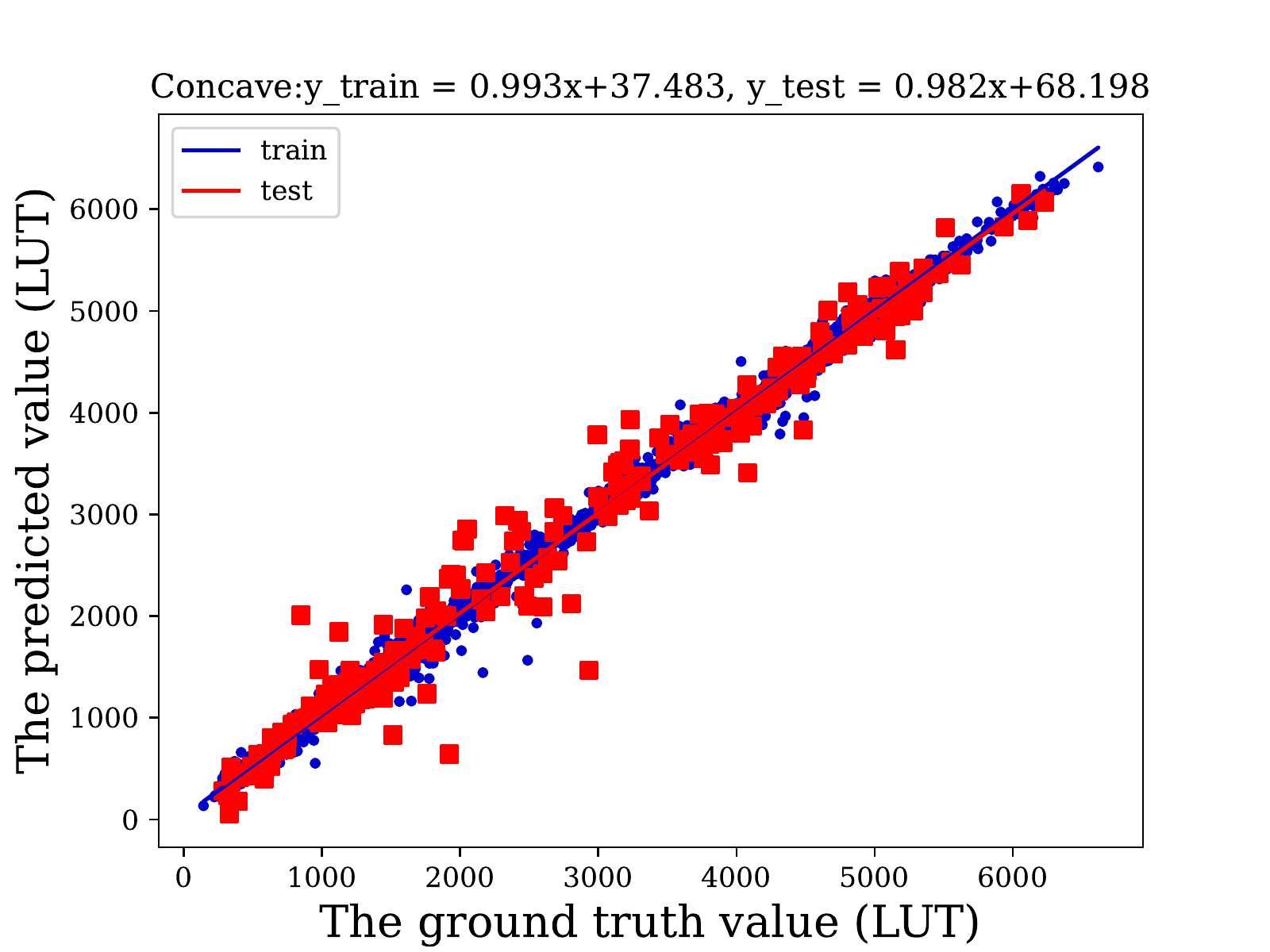}
   \caption{LUT entry-wise concave proxy}
\end{subfigure}
\caption{The visualization of different proxies in LUT learning in application II.}
\label{fig:application_2_proxy_apd}
\end{figure*}
\begin{figure*}[h]
\centering
\begin{subfigure}{0.23\textwidth}
   \includegraphics[width=\linewidth]{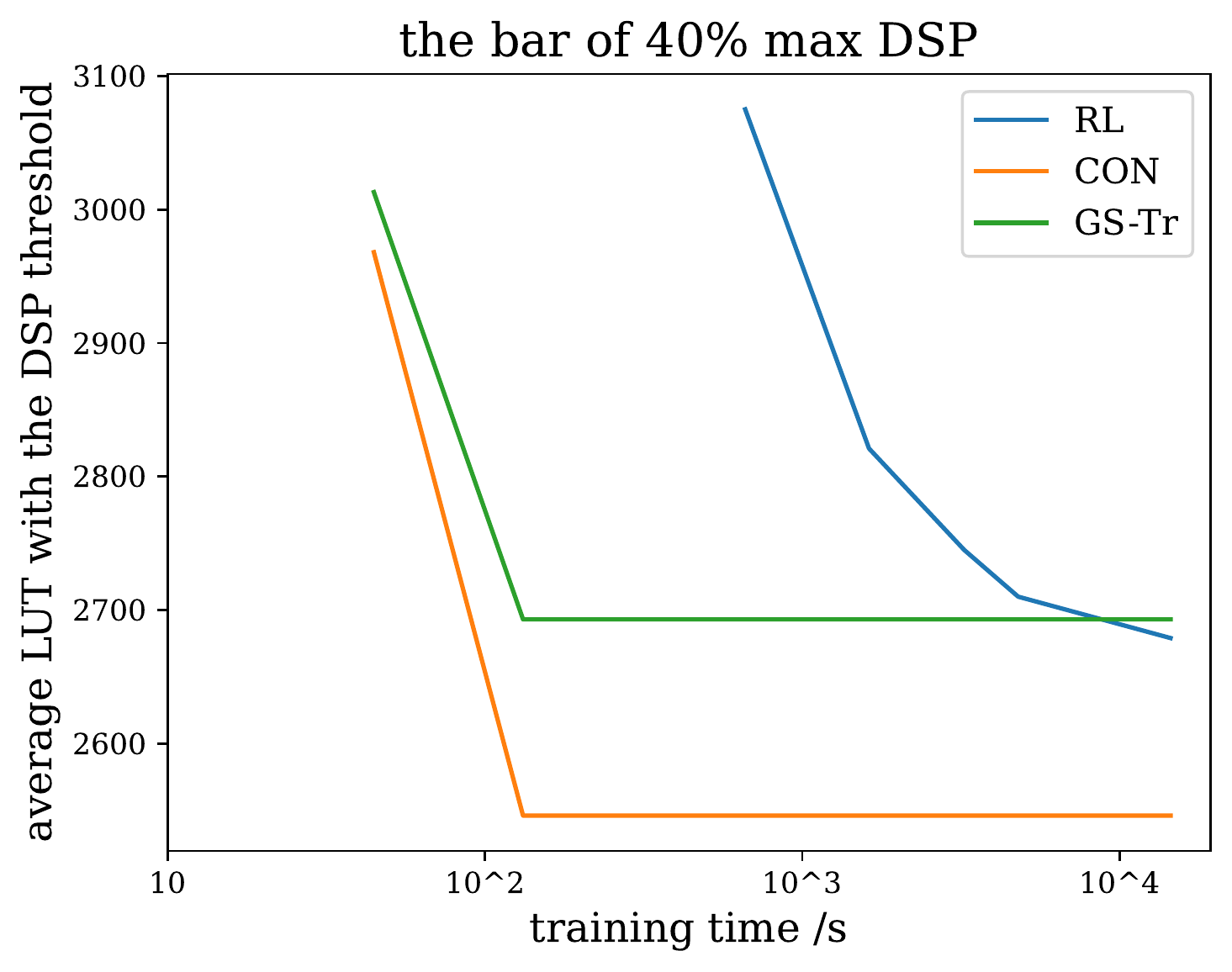}
   \caption{DSP threshold: 40 \%} \label{fig:application_1_time_1}
\end{subfigure}
\hspace*{\fill}
\begin{subfigure}{0.23\textwidth}
   \includegraphics[width=\linewidth]{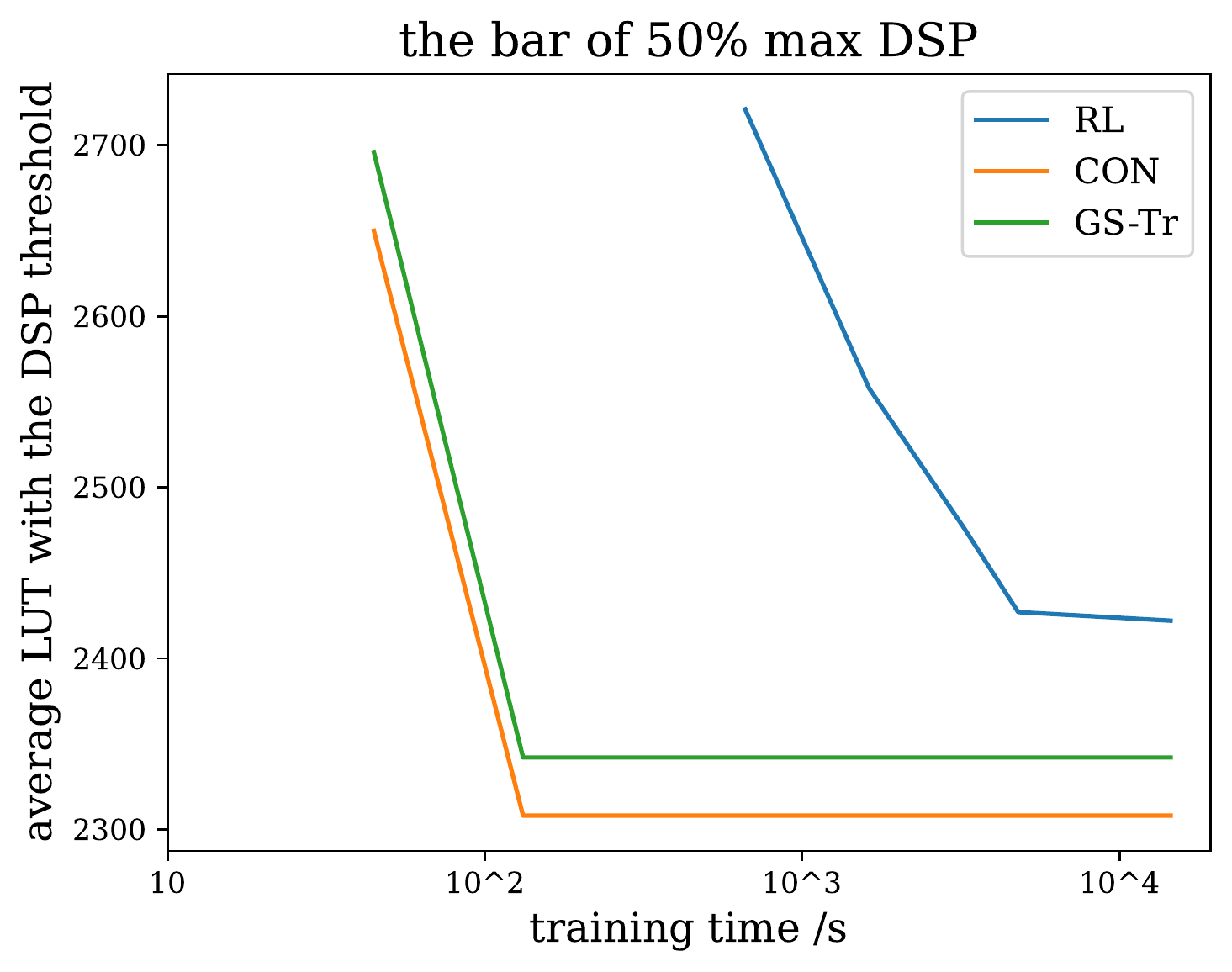}
   \caption{DSP threshold: 50 \%} \label{fig:application_1_time_2}
\end{subfigure}
\hspace*{\fill}
\begin{subfigure}{0.23\textwidth}
   \includegraphics[width=\linewidth]{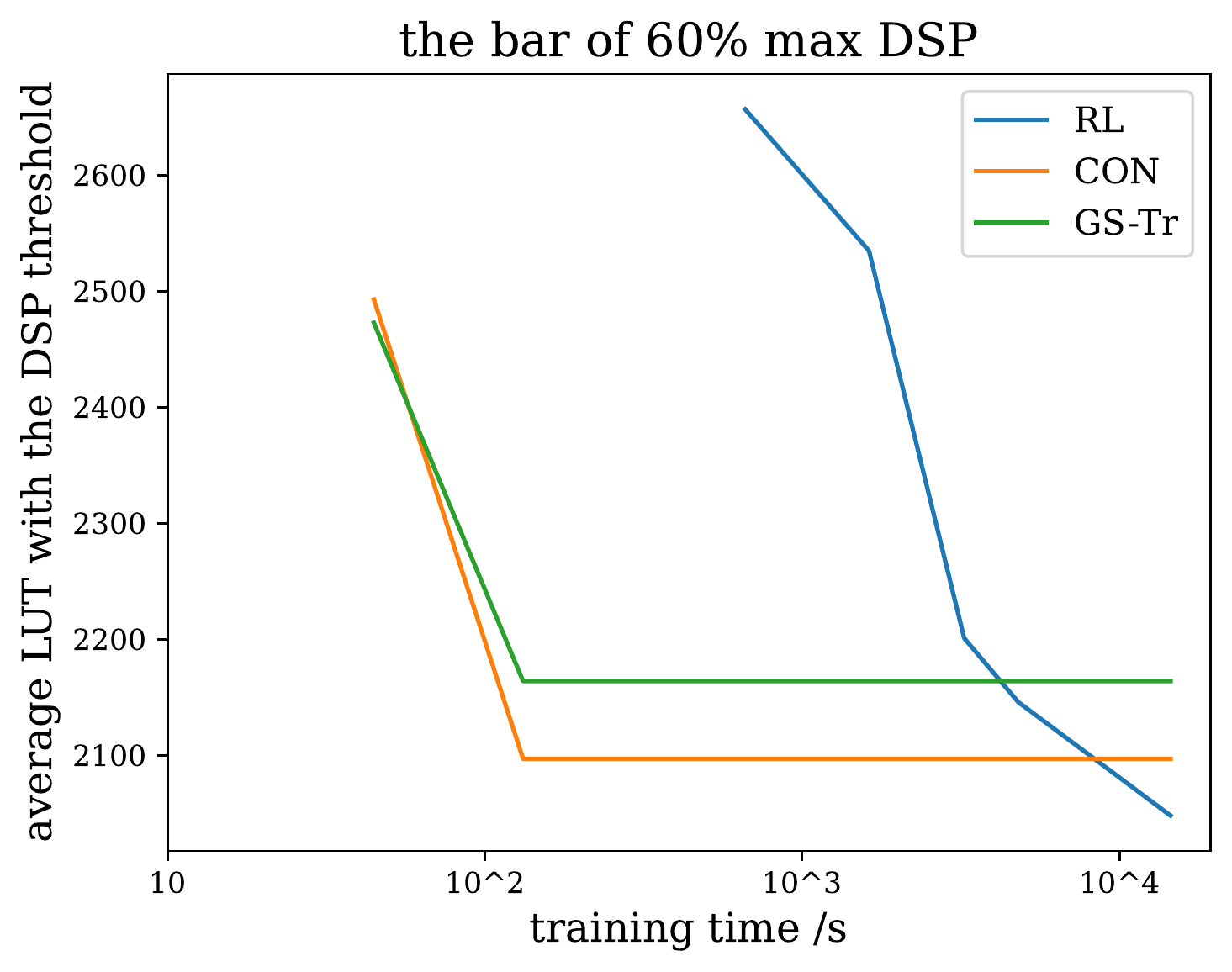}
   \caption{DSP threshold: 60 \%} \label{fig:application_1_time_3}
\end{subfigure}
\hspace*{\fill}
\begin{subfigure}{0.23\textwidth}
   \includegraphics[width=\linewidth]{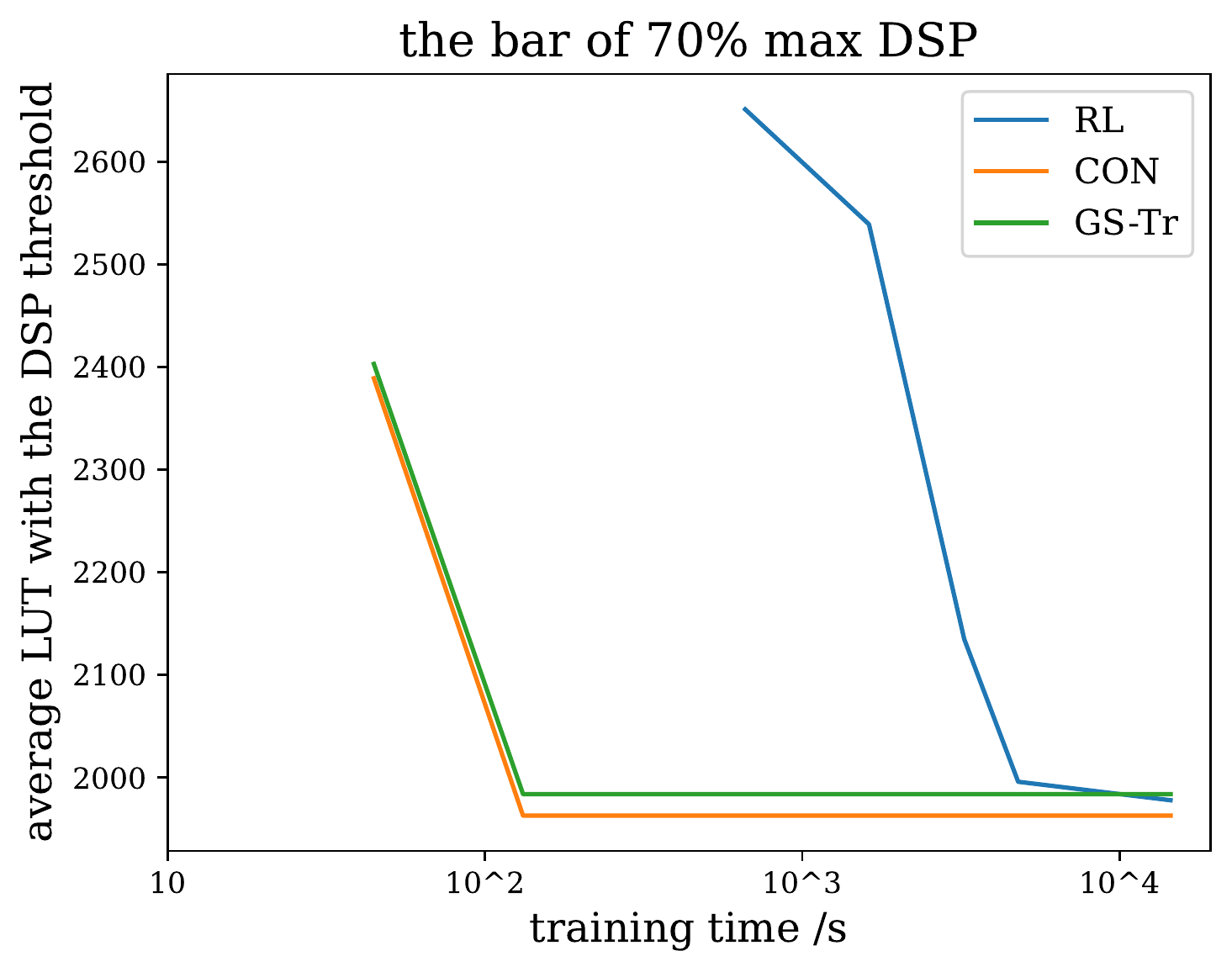}
   \caption{DSP threshold: 70 \%} \label{fig:application_1_time_4}
\end{subfigure}
\caption{Training time of RL, vs GS-Tr, vs CON with different constraints on DSP usage.}
\label{fig:application_2_time_apd}
\end{figure*}

We also include the percentages of cases where each method takes the first, second and third places according to the rank, which is shown in Table~\ref{tab:application_2_ranking_2}. The comparison about how different proxies approximate the ground-truth LUT usage amount is shown in Fig~\ref{fig:application_2_proxy_apd}. Again, the entry-wise affine proxy may introduce large error while the entry-wise concave proxy could approximate in a better sense. 

We investigate the training time of RL method, GS-Tr method and the entry-wise concave method. The comparison among these methods is shown in Fig~\ref{fig:application_2_time_apd}. We run all the three methods on the same server with $2$ Intel(R) Xeon(R) Gold 6248R CPUs, $1000$GB RAM in total. In each experiment we take $26$ processes of the CPU and run on one Quadro RTX 6000 GPU card. We count the time cost during training and select the models at different epochs for testing at intervals. Note that the training objectives in all methods use proxies while the testing results are the outputs given by the HLS tool. Due to the fact that inferring via HLS consumes a lot of time, we only test limited numbers of cases to draw the figure and thus the curves are not smooth.

We further investigate the inference time of GA method and the entry-wise concave method. The comparison is in Fig.~\ref{fig:application_2_apd_ga}. In each generation, the inference of all the population runs in parallel on one Quadro RTX $6000$ GPU card, which is the same as our training $\mathcal{A}_{\theta}$ process. Our rounding process takes almost the same time as GA in $100$ generations, while our CON method achieves competitive results with GA in $500$ generations which takes $4$ times longer inference time than our method.

\subsection{Application III}
For each method, we count the ratio that the relative error based on the assignment given by this method exceeds the optimal assignment (ratio = relative error / OPT relative error - 1). The smaller the ratio is, the closer the method's relative error is to the optimal solution.
The results are shown in Table~\ref{tab:application_3_apd}. We also include the training time that the methods require to achieve the corresponding performance in the table. All the methods run on the same server with $2$ Intel(R) Xeon(R) Gold 6248R CPUs, $1000$GB RAM in total. In each experiment we take $26$ processes of the CPU and run on one Quadro RTX 6000 GPU card. The time is obtained by counting the least epoch that a model achieves its reported performance, the time of OPT is the time that the brute-force search takes in the testset.

\begin{table}[h]
\centering
\setlength\tabcolsep{3.5pt}
\begin{tabular}{@{}cccccccccc@{}}
\toprule
Threshold$\theta$ & C-In   & C-Out  & Naive & RL     & GS-Tr+S & GS-Tr+R & CON   & AFF   & OPT    \\ \midrule
3         & 348.47 & 349.09 & 30.68 & 282.31 & 75.68   & 16.78   & 14.80 & 11.91 & 0      \\
5         & 209.70 & 209.07 & 30.80 & 159.07 & 69.17   & 23.53   & 8.2 & 13.50 & 0      \\
8         & 99.41  & 99.06  & 29.90 & 77.21  & 47.68   & 24.03   & 18.80 & 17.28 & 0      \\
Time / s  & 0      & 0      & 90+   & 9851+  & 90+     & 90+     & 92+   & 92+   & 32776+ \\ \bottomrule
\end{tabular}
\caption{The averaged ratios (\%) that the relative errors of different methods exceed the OPT on application III. The required training time to achieve the performance is listed at the bottom row.}
\label{tab:application_3_apd}
\end{table}

\subsection{Study I: Further Evaluation on The Learning Capability of Different Proxies.}

\begin{wraptable}{r}{0.4 \textwidth}
\vspace{-0.45cm}
\centering
\begin{tabular}{@{}cccc@{}}
\toprule
Proxy & AFF  & CON  & W/O limit \\ \midrule
Train & 0.19 & 0.17 & 0.10      \\
Test  & 0.19 & 0.18 & 0.19      \\ \bottomrule
\end{tabular}
\caption{Huber loss on application III. }
\label{tab:proxy_loss_apd}
\vspace{-1.2cm}
\end{wraptable}
We have studied the learning capability of different proxies for Application II, as shown in Fig.~\ref{fig:application_2_proxy_apd} and Fig.~\ref{fig:application_2_proxy}. We further study how proxies under different constraints fit with historical data on application III. We show the Huber loss~\cite{huber1992robust} of the proxies in Table~\ref{tab:proxy_loss_apd}. The objective of application III has $4$-order moment, the three proxies obtain comparable approximation performance.

\subsection{Study II: The Effectiveness of Our `Rounding' Process}
Here we empirically study the effectiveness of our rounding procedure. We show the average relaxed loss value ($l_r = f_r(\bar{X};C) + \beta g_r(\bar{X};C), \bar{X} \in [0,1]^n$), the average rounded loss value ($l_r' = f_r(\hat{X};C) + \beta g_r(\hat{X};C), \hat{X} \in \{0,1\}^n$ and the average true value of the rounded assignment ($l = f(\hat{X};C) + \beta g(\hat{X};C)$) on the testset of each problem in Table~\ref{tab:rounding}.
According to the table, we observe that both the methods that adopt entry-wise affine proxies and  entry-wise concave proxies are guaranteed to obtain a drop of the loss values after our rounding procedure. However, for the proxies that do not satisfy the constraints, the Na\"{i}ve method and the GS-Tr-R baseline could not always guarantee such a drop after the rounding process. In particular, the rounding in GS-Tr-R increases the loss in the application of A$\times$C circuit design, while the rounding in Na\"{i}ve increases the loss in the applications of 
edge covering and A$\times$C circuit design.

\subsection{Study III: The Random Seeds in the Experiments}
\label{sec:random_seed_apd}
\begin{wraptable}{r}{0.5 \textwidth}
\vspace{-0.5cm}
\resizebox{0.5\textwidth}{!}{\begin{tabular}{@{}ccccccc@{}}
\toprule
           & \multicolumn{3}{c}{AFF}        & \multicolumn{3}{c}{CON}        \\ \midrule
Threshold  & seed \#1 & seed \#2 & seed \#3 & seed \#1 & seed \#2 & seed \#3 \\
3 AC units & 3.098    & 3.117    & 3.091    & 3.180    & 3.219    & 3.154    \\
5 AC units & 5.423    & 5.358    & 5.377    & 5.168    & 5.109    & 5.132    \\
8 AC units & 10.064   & 10.035   & 10.022   & 10.185   & 10.150   & 10.178   \\ \bottomrule
\end{tabular}}
\caption{AFF and CON with three random seeds.}
\label{tab:random_seed_apd}
\end{wraptable}
We study the affect from random seeds on AFF and CON proxy in application III. We report the performance of the three random seeds in application III in Fig.~\ref{tab:random_seed_apd}. We observe no significant difference among the three seeds in our cases.

\begin{table}[]\scriptsize
\setlength\tabcolsep{2pt}
\begin{tabular}{@{}c|cccc|cccc|ccc|cccc@{}}
\toprule
      & \multicolumn{4}{c|}{Edge covering (App. I)} & \multicolumn{4}{c|}{Node matching (App. I)}      & \multicolumn{3}{c|}{Resource allocation (App. II)} & \multicolumn{4}{c}{A$\times$C circuit design (App. III)} \\ 
     \midrule
Proxy            & Na\"{i}ve  & GS-Tr-R  & CON    & AFF    & Na\"{i}ve   & GS-Tr-R    & CON     & AFF     & Na\"{i}ve      & GS-Tr-R      & CON        & Na\"{i}ve  & GS-Tr-R  & CON   & AFF  \\
$l_r$              & 62.58  & 78.56  & 80.91  & 46.37  & 5316.07 & 13103.39 & 5922.61 & 5389.95 & 5899.10    & 3485.60    & 2785.95    & 4.41   & 4.23   &6.75  & 6.67 \\
$l_r'$          & 70.20  & 51.04  & 52.09  & 45.73  & 442.15  & 442.03   & 430.11  & 432.63  & 3350.35    & 2741.32    & 2599.89    & 9.31   & 7.27   & 5.22  & 5.19 \\
$l$              & 68.52  & 46.91  & 52.04  & 44.55  & 429.12  & 429.39   & 422.47  & 418.96  & 2901.19    & 2749.08    & 2511.70    & 6.98   & 6.57   & 6.21  & 6.17 \\
OPT        & \multicolumn{4}{c|}{42.69}         & \multicolumn{4}{c|}{416.05}             & \multicolumn{3}{c|}{no OPT}           & \multicolumn{4}{c}{5.36}       \\ \bottomrule
\end{tabular}
\caption{The relaxed loss value $l_r$, the rounded loss value $l_r'$ and its true value $l$ of the methods.}
\label{tab:rounding}
\end{table}

\section{Experimental details}
\label{apd:implementation_details}
All of the experiments are carried out on the same server with $2$ Intel(R) Xeon(R) Gold 6248R CPUs, $1000$GB RAM in total. In each experiment we take $26$ processes of the CPU and run on one Quadro RTX 6000 GPU card. The maximum GRAM of the Quadro RTX 6000 GPU is 24GB. The proxies that satisfies our principle (AFF, CON) and GS-Tr run on PyTorch~\cite{paszke2019pytorch} frame with PyTorch geometric~\cite{fey2019fast}. The RL baseline follow the actor-critic technique in~\cite{konda1999actor}. ~\cite{wu2021ironman} also utilizes the same RL technique to solve the same problem.
The details of each dataset is displayed in Table.~\ref{tab:dataset_detail}. Adam~\cite{kingma2015adam} is used as the optimizer in all of the experiments. All the experiment results are conduct and averaged under three random seeds $12345$, $23456$ and $34567$.

To be fair, in the training process, we first train the baseline methods, such as the proxy without constraints and $\mathcal{A}_{\theta}$ based on the Gumbel-softmax trick. Then we train our entry-wise concave proxy and $\mathcal{A}_{\theta}$ with exactly the same hyper-parameters with the baselines except for the necessary changes to construct the entry-wise concavity.

\begin{table}[h]
\centering
\begin{tabular}{@{}ccccc@{}}
\toprule
Task             & Toy example & Application I & Application II & Application III \\ \midrule
$f_r$, $g_r$ training   & 95,000      & 95,000        & 7,200          & 95,000          \\
$f_r$, $g_r$ testing    & 5,000       & 5,000         & 800            & 5,000           \\
$A_{\theta}$ training       & 10,000      & 10,000        & 40             & 1,000           \\
$A_{\theta}$ testing        & 500         & 500           & 20             & 500             \\
\bottomrule
\end{tabular}
\caption{The number of instances in each dataset.}
\label{tab:dataset_detail}
\end{table}

\subsection{Other ways to construct the entry-wise concave proxy}
\label{apd:other-proxy}
Besides constructing the $2$-order entry-wise affine latent representation as introduced in Section~\ref{sec:proxy} Eq.\eqref{eq:latent} ($\phi(\bar{X};C) = W + \sum_{v \in V} U_{v} \bar{X}_v + \sum_{v,u\in V, (v,u)\in E} Q_{v,u} \bar{X}_{v} \bar{X}_{u}$), here we introduce another approach to obtain the latent representations with higher than $2$-order moments, which can also be conveniently implemented in the current platform.
We construct such latent representation as follows:
\begin{equation}
    \begin{aligned}
    \phi (\bar{X};C) = \sum_{v \in V} (U_{v}\bar{X}_v + U_{v}') \prod_{u:(u,v) \in E} (Q_{v,u}\bar{X}_u + Q_{v,u}').
    \end{aligned}
\end{equation}
where $U_v, U_v'$ are node representations (by breaking one whole node representation into two parts), and $Q_{v,u}$ and $Q_{v,u}'$ are edge representations (by breaking one whole edge representation into two parts) output by GNNs. The above form could be obtained by running one-layer message passing and the log-sum-exponential trick: 
$\sum_{v \in V} \exp [\log (U_v \bar{X}_v + U_v') + \sum_{u:(u,v) \in E} \log (Q_{v,u}\bar{X}_u + Q_{v,u}')]$. This alternative approach of the latent representation construction could generate the entry-wise affine latent representations with the order of the moments as high as the maximum degree of the graph. 

\subsection{The toy example}
The ground truth of the objectives is designed as follows.
\begin{equation}
\begin{aligned}
    f(X_1,X_2;C) = g_1(C)  X_1 + g_2(C)X_2 + g_3(C)  X_1  X_2 + g_4 (C),
\end{aligned}
\end{equation} where $C = [C_1,C_2]$ and 
\begin{equation}
\begin{aligned}
& g_1 = (580 - 10 C_1 - 3C_2) / 33, \\
& g_2 = (580 - 10 C_2 - 3C_1) / 33, \\
& g_3 = (3C_1 + 3 C_2 )/45, \\
& g_4 = - (5C_1 + 5C_2 )/33 + 60. \\
\end{aligned}
\end{equation}
The constants are set arbitrarily. To match our graph-based pipeline, in this toy example, we also build a single edge graph for each configuration where the two nodes are associated with the attributes $C_1$ and $C_2$, and the binary variables $X_1$ and $X_2$.

\textbf{The proxy without constraints:}
We use $3$ layers of GraphSAGE~\cite{hamilton2017inductive} convolutional layers that take both the node attibutes $C$ and the optimization variables $\bar{X}$ as inputs with leaky ReLU activation and batch normalization. Then, the structure is followed by a global mean pooling. After several MLP layers, the proxy outputs the cost. We use MSE loss to train this proxy. The learning rate is set as 1e-2. The batch size is set as $4096$. The reported performance is trained within $200$ epochs. 

\textbf{The entry-wise concave proxy:}
We use $3$ layers of GraphSAGE with the same hyper-parameters as the proxy without constraints but it only takes the configuration $C$ as inputs to encode the configuration. 
Then, the encoded configuration $H \in \mathbb{R}^{|V| \times d}$ ($d$ is the dimension of the latent node feature) is separated equally into two parts $U \in \mathbb{R}^{|V| \times \frac{d}{2}}$ and $W \in \mathbb{R}^{|V| \times \frac{d}{2}}$, we construct $U \bar{X} + W, (\bar{X} \in [0,1]^{|V|})$, then we multiply $U_1 \bar{X}_1 + W_1$ with $U_2 \bar{X}_2 + W_2$ to obtain the latent node representation $\phi(\bar{X};C)$, followed by a linear layer as the implementation of the AFF proxy $h_r^a(\bar{X};C)$ as introduced in Section~\ref{sec:proxy} Eq.~\ref{eq:con-proxy}. MSE loss is used as the criterion, the learning rate is 1e-2, the batch size is $4096$, the model is trained within $200$ epochs.


\textbf{$\mathcal{A}_{\theta}$ based on Gumbel-softmax tricks:}
When training $\mathcal{A}_{\theta}$, we also use $3$ GraphSAGE layers to encode the configuration $C$ with leaky Relu activation and batch normalization. Then the encoded latent feature is followed by fully connected layers to reduce the dimension and the Gumbel-softmax trick to sample a distribution from the soft probability predicted by the model. We use the soft Gumbel-softmax~\cite{maddison2016concrete,jang2016categorical} without the straight through trick. The learning rate is set as 1e-2, the batch size is $4096$, the model is trained for $200$ epochs.

\textbf{$\mathcal{A}_{\theta}$ in $\{$relaxation (Na\"{i}ve), relaxation with entry-wise concave proxy (Ours)$\}$:} 
In $\mathcal{A}_{\theta}$, the structure is the same as that based on Gumbel-softmax tricks except that the structure has no pooling layer and takes a Sigmoid layer. The learning rate is 1e-2, the batch size is $4096$, the model is trained for $200$ epochs.

\subsection{Application I - Edge Covering}
\label{apd:edge-proxy}
\begin{figure}[t]
    \centering
    \includegraphics[width = 4in]{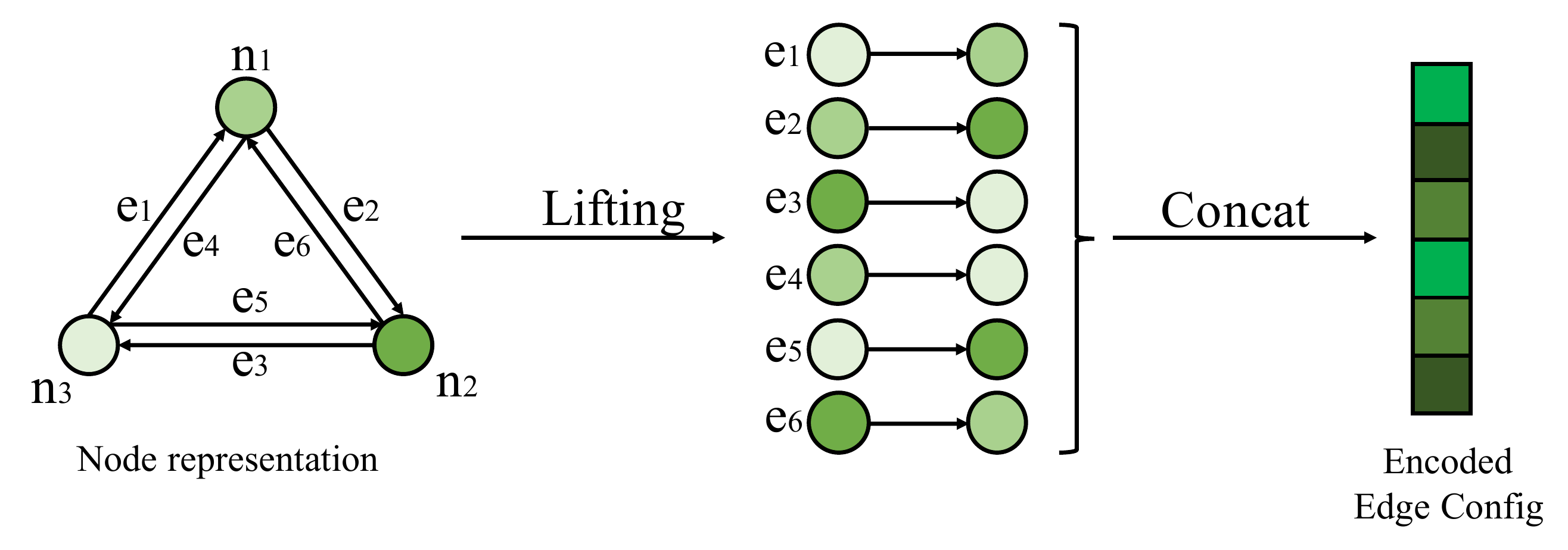}
    \caption{The generalization from node problems to edge problems.}
    \label{fig:edge_proxy_apd}
\end{figure}

\textbf{Dataset details:}
Each configuration $C$ is a $4 \times 4$ grid graph. Each node of the graph consists of two images randomly selected from the MNIST dataset~\cite{deng2012mnist} and thus represents a number between $00$ and $99$.

The ground truth of the objective are designed as follows:
\begin{equation}
    f(X;C) = \sum_{e \in E} w_e X_e,
\end{equation} where
\begin{equation}
\begin{aligned}
& w_e = (C_{v} + C_{u}) / 3 + (C_{v} C_{u}) / 100, \;\text{for}\; e=(u,v)\in E.
\end{aligned}
\end{equation}
\textbf{The proxy without constraints:}
We firstly utilize ResNet-50~\cite{he2016deep} to extract the latent fixed node feature and then send the feature into a GNN, the GNN is also based on $3$ MPNN~\cite{gilmer2017neural} layers, which involves the edge assignment $\bar{X}, (\bar{X}\in [0,1]^{|E|})$ in the message passing. Global mean pooling is used to generate the final predicted value $f_r(\bar{X};C)$. MSE loss is utilized as the criterion, the learning rate is 5e-3, the batch size is $160$.

\textbf{The entry-wise affine proxy:}
We also use ResNet-50 and $3$ layers of MPNN~\cite{gilmer2017neural} which take the output of ResNet-50 as inputs with leaky Relu activation and batch normalization to encode the configuration $C$ into the latent node representation $H' \in \mathbb{R}^{|V| \times d_1}$ ($d_1$ is the dimension of the node feature). After the encoding procedure, the encoded node features are lifted to each side of the edges according to the edge index, then these node features on two sides of the edges are concatenated together and sent into MLP layers to generate the latent edge representation $H'' \in \mathbb{R}^{|E| \times d_2}$ ($d_2$ is the dimension of the edge feature). Then $H''$ is separated into two parts $U \in \mathbb{R}^{|E| \times \frac{3d_2}{4}}$ and $W \in \mathbb{R}^{|E| \times \frac{d_2}{4}}$, and we calculate $U \bar{X} + W, (\bar{X} \in [0,1]^{|E|})$ to construct the latent representation $\phi(\bar{X};C)$. The structure is followed by mean pooling and linear layers to construct the AFF proxy $h_r^a(\bar{X};C)$. The whole procedure generalizes our framework from solving node problems to edge problems, as is shown in Fig.~\ref{fig:edge_proxy_apd}. We use MSE loss for training, the learning rate is set as 5e-3, the batch size is $160$.

\textbf{The entry-wise concave proxy:}
The network shares the same structure with the AFF proxy in the front part, while we utilize linear layers mixed with a negative $\text{Relu}$ function to construct the CON proxy $h_r^c(\bar{X};C) = \langle w^c, -\text{Relu}(\phi(\bar{X};C)) \rangle + b$, as introduced in Section~\ref{sec:proxy}. Note that we use $\text{torch.clamp()}$ function to control the entries in $w^c$ greater or equal to zero in each batch of data during the training  process. We use MSE loss for training and set the learning rate as 5e-3, the batch size is $160$.

\textbf{The RL baseline:}
We apply an actor critic model~\cite{konda1999actor}. This model consists of $4$ key components: 1)States, the states are formulated as every possible partially assigned grid graph; 2) Actions, given the current state and the currently candidate edges of the grid graph, the action is which new edge to pick. Note that the model is only allowed to pick from the edges which connect at least one node that has not been covered yet; 3) State transition, given a state and an action, the probability of the next states; 4)Reward, the reward is $0$ for all intermediate actions, in the last action the reward is the evaluation of the covering score predicted by the proxy without constraints.

In each state at a time step, we extract the features from the last layer of the proxy without constraints $f_r$. We utilize another ResNet-50 + GNN to encode the whole grid graph into a vector encoding. The features are further combined with the vector embedding as the state encoding. Then the state encoding is sent into the policy network that is made up of multiple MLP layers to output the critic value $c$ and the action $a$ which indicates the next edge to pick from. The loss for the actor is calculated by subtracting the reward by $c$, and we use Huber loss to make $c$ close to the reward. In each state, the model would only choose from the edges which connect at least one node that has not been covered yet. The reward is defined as the negative proxy prediction:
\begin{equation}
\begin{aligned}
 r_t=\left\{
\begin{aligned}
&-f_r(X;C),& s = T \\
&0, & 0<s<T,
\end{aligned}
\right.
\end{aligned}
\end{equation} where $T$ is the max step, and $s$ denotes the number of the step. The learning rate is set as 1e-2, the discount factor for the reward is set as $0.95$, we train the RL baseline for more than $20,000$ epochs to achieve the reported performance.

\textbf{The constraint $g_r(\bar{X};C)$:}
As to the penalty constraint $g_r(\bar{X};C) = \sum_{v \in V} \prod_{e:v\in e} (1 - \bar{X}_e)$ which naturally satisfies our definition of CO problems in Section~\ref{def_CO}, we apply the log-sum-exponential trick $\sum_{v\in V} \exp(\sum_{v \in e} \log (1-\bar{X}_e))$ to calculate it via message passing in PyTorch geometric.

\textbf{$\mathcal{A}_{\theta}$ based on Gumbel-softmax trick:}
We utilize $3$ GraphSAGE layers to encode the node feature, with leaky Relu activation function and batch normalization.
The encoded node features are lifted to each side of the edges, concatenated together and then sent into MLP layers to reduce the dimension and map to $\mathcal{A}_{\theta}(C)\in [0,1]^{|E|}$. Then the model is followed by the Gumbel-softmax trick to obtain the output $X \sim \text{Ber}(\mathcal{A}_\theta)$. We use the soft Gumbel-softmax~\cite{maddison2016concrete,jang2016categorical} without the straight through trick. The learning rate is set as 1e-3, the batch size is $60$.


\textbf{$\mathcal{A}_{\theta}$ in $\{$relaxation (Na\"{i}ve), relaxation with entry-wise concave proxy (Ours)$\}$:}
The model shares the same structure as that based on the Gumbel-softmax trick, except that the Gumbel-softmax trick is replaced by $\bar{X}\in [0,1]^{|E|}$ directly. The learning rate is set as 1e-3, the batch size is $60$.


\subsection{Application I - Node Matching}
\label{apd:match-proxy}
The ground truth of the objective are designed as follows:
\begin{equation}
    f(X;C) = \sum_{e \in E} w_e X_e,
\end{equation} where
\begin{equation}
\begin{aligned}
& w_e = C_{v} C_{u}, \;\text{for}\; e=(u,v)\in E.
\end{aligned}
\end{equation}
Every structure design keeps the same as Application I - Edge Covering except for the extra penalty constraint $\prod_{\substack{e_1,e_2:v\in e_1,e_2 \\ e_1\neq e_2} }\bar{X}_{e_1}\bar{X}_{e_2}$ in Eq.~\eqref{eq:node-matching} which naturally follows our request on $g_r$, and is also conducted via matrix operations on PyTorch geometric. As to the RL baseline, the structure design is basically the same as Application I, while in this problem, at each time step, the model is only allowed to pick an edge whose two nodes are both not covered. The reward is defined as follows:
\begin{equation}
\begin{aligned}
 r_t=\left\{
\begin{aligned}
&-f_r(X;C) & s = T\\
&- \beta & 0<s<T, \text{\ no options} \\
&0, & 0<s<T, \text{\ option exists},
\end{aligned}
\right.
\end{aligned}
\end{equation} where $\text{``no options''}$ means that there are some covered nodes whose neighboring nodes have been all covers, $\text{``option exists''}$ denotes the case when eligible edges still exist, $T$ is the max step, $\beta$ is a large hyper-parameter, and $s$ denotes the number of the step. The learning rate is set as 1e-2, the discount factor for the reward is set as $0.95$, we train the RL baseline for more than $20,000$ epochs to achieve the reported performance.

\subsection{Application II - Resource Binding Optimization}
\textbf{Dataset details}
\label{apd:application_2-proxy}
In this application, we focus on the resource binding problems in field-programmable gate array (FPGA) design. Each configuration $C$ in the dataset is a data flow graph (DFG) with more than $100$ nodes. Each node represents an arithmetic operation such as multiplication or addition. The operations need to be one-to-one mapped into a micro circuit to carry out the calculation. Given an assignment of the mapping, we run high-level synthesis (HLS) simulation tools to obtain the actual circuit resource usage under the assignment, which might take up to hours time. Note that different assignments of the mapping could result in vastly different actual resource usage. 

In this dataset, we focus on the resource balancing problems between digital signal processors (DSP) and look-up tables (LUT). Here DSP is a small processor that is able to quickly perform mathematical operation on streaming digital signals, LUT is the small memory that is used to store truth tables and perform logic functions. The optimization goal of the dataset is to allocate those nodes with pragma to either LUT or DSP, such that the actual usage amount of LUT could be minimized given a maximum usage amount of the DSP usage. We use $1$ (LUT) or $0$ (DSP) to assign each node's mapping. We encode the fixed node feature into a $10$-dimension embedding which contains the following information: 1) 4 digits to indicate the types of nodes in \{input, m-type, intermediate-type, output\}; 2) 5 digit binary encoding of the node's calculation precision, from $2$ bits to $32$ bits; 1 digit encoding that indicates whether the node requires pragma. For those nodes that do not require pragma, HLS tools have a set of heuristic assignments to the nodes during the simulation.

\textbf{The proxy without constraints:} We separately utilize two GraphSAGE GNN models~\cite{hamilton2017inductive} to predict the LUT usage $f_r(\bar{X};C)$ and the DSP usage $g_r(\bar{X};C)$, the structure of them are the same. We use $3$ layers of GraphSAGE~\cite{hamilton2017inductive} convolutional layers that take both the node attibutes $C$ and the optimization variables $\bar{X}$ as inputs with leaky ReLU activation and batch normalization. Then, the structure is followed by a global mean pooling. After several MLP layers, the proxy outputs the cost. We use MSE loss to train this proxy. The learning rate is set as 1e-3. The batch size is set as $256$.

\textbf{The entry-wise affine proxy:}
For both $f_r$ and $g_r$, we use $3$ layers of GraphSAGE to encode the configuration $C$ and the hyper-parameter $\alpha$ to control the DSP usage threshold with leaky Relu activation and batch normalization into the latent node representation $H \in \mathbb{R}^{|V| \times d}$ ($d$ is the node feature dimension). The hyper-parameters of the layers are exactly the same as the proxy without constraints. Then $H$ is separated equally into two parts $U \in \mathbb{R}^{|V| \times \frac{d}{2}}$ and $W \in \mathbb{R}^{|V| \times \frac{d}{2}}$, and we calculate $U \bar{X} + W$, after that we do the log-sum-exponential trick $\sum_{v \in V} \exp [ \log ( U_v \bar{X}_v + W_v) + \sum_{u:(u,v) \in E} \log ( U_u \bar{X}_u + W_u)]$ via message passing to generate the $2$-order moment entry-wise affine latent representation as introduced in Section~\ref{sec:proxy}.
Finally, the global mean pooling and a linear layer is used to obtain the output $f_r(\bar{X};C), g_r(\bar{X};C)$. Huber loss is used as the criterion, the learning rate is set as 5e-4 for $g_r$ and 1e-3 for $f_r$, the batch size is $256$.

\textbf{The entry-wise concave proxy:}
The models share basically the same structure as that in the AFF proxy except for the last layers. We use linear layers with a negative $\text{Relu}$ function to construct the CON proxy $h_r^c (\bar{X};C) = \langle w^c, -\text{Relu}(\phi(\bar{X};C)) \rangle + b$, as introduced in Section~\ref{sec:proxy}. We utilize $\text{torch.clamp()}$ function to control the entries in $w^c$ to be always greater or equal to zero in each batch of data processing during the training process.

\textbf{The simulated annealing baseline:}
We run the simulated annealing algorithm guided by the proxy without constraints. The initial temperature is set as $1000$, the cool down factor is $0.99$, the ending temperature is $699$. For each temperature, the number of jumps is $20$. And, we set the probability for mutation is $0.1$.

\textbf{The genetic algorithm baseline:}
We run the genetic algorithm guided by the proxy without constraints. The algorithm runs parallel on GPU. The max generation is set as $100$, the population of each generation is $256$, which is the same as the batch size that we used to train $\mathcal{A}_{\theta}$, the probability of crossover is $0.6$, the probability of gene mutation is $0.01$. The inference via proxy of each generation runs parallel on GPU.

\textbf{The RL baseline:}
We apply an actor critic model~\cite{konda1999actor}. This model consists of $4$ key components: 1)States, the states are formulated as every possible partially assigned DFG; 2) Actions, given the current state and the currently considered node of the DFG, the action is whether to assign the LUT to this node; 3) State transition, given a state and an action, the probability of the next states; 4)Reward, the reward is $0$ for all intermediate actions, in the last action the reward is the evaluation of the fully assigned DFG subject to the DSP usage threshold.

In each state at a time step, we extract the features from the last layer of the proxies without constraints $f_r, g_r$ and concatenate them together. We utilize another GNN to encode the whole DFG into a vector encoding. The concatenated features is further combined with the vector embedding as the state encoding. Then, the state encoding is sent into the policy network that is made up of multiple MLP layers to output the critic value $c$ and the action $a$ which indicates whether to assign LUT for the current multiplication node. The loss for the actor is calculated by subtracting the reward by the critic value $c$, we use Huber loss to make $c$ close to the reward. Note that the above scheme follows the original paper that studied the same application~\cite{wu2021ironman} while the status representation is based on an intermediate output given by the GNN in our proxy without constraints. The reward is defined as the negative weighted sum of LUT usage and the difference between the DSP usage and the DSP threshold:
\begin{equation}
\begin{aligned}
 r_t=\left\{
\begin{aligned}
&-\alpha f_r(X;C) - \beta \text{Relu}(t-g_r(X;C)),& s = T \\
&0, & 0<s<T,
\end{aligned}
\right.
\end{aligned}
\end{equation} where $T$ is the max step, $\alpha, \beta$ are hyper-parameters and set as $0.1,10$ respectively, $t$ is the DSP usage threshold and $s$ denotes the number of the step. The learning rate is set as 1e-2, the discount factor for the reward is set as $0.95$, we train the RL baseline for more than $9,000$ epochs to achieve the reported performance.

\textbf{The mapping of $g_r(\bar{X};C)$:} 
Here we introduce the mapping of the constraints in detail. 
The relaxed optimization goal could be written as follows:
\begin{equation}
\begin{aligned}
\min_{\theta} f_r(\bar{X};C), \ \ \text{s.t. }  g_r(\bar{X};C) < t,
\end{aligned}
\end{equation}where $t-1$ is the threshold for the DSP usage amount, $\bar{X} = \mathcal{A}_{\theta} \in [0,1]^{n}$.
As introduced in Section~\ref{def_CO}, we map the above constraints into the normalized constraint $g_r'(\bar{X};C)$ via the following normalization.
\begin{equation}
\begin{aligned}
g_r'(\bar{X};C) = \frac{g_r(\bar{X};C) - g_{\min}}{g_{\min}^+ - g_{\min}},
\end{aligned}
\end{equation}where $g_{\min}^+ = \min_{X \in \{0,1\}^{|V|} \backslash \Omega} g_r(X;C) = t$ and $g_{\min} = \min_{X\in \{0,1\}^{|V|}} g_r(X;C) = 0$ in this case. Thus, the normalized constraint could be written as:
\begin{equation}
\begin{aligned}
g_r'(\bar{X};C) = \frac{g_r(\bar{X};C)}{t}.
\end{aligned}
\end{equation}The constraint above could satisfy our definition of the CO problems as introduced in Section~\ref{def_CO}. The overall loss function could thus be written as follows:
\begin{equation}
\begin{aligned}
l_r(\bar{X};C) = f_r(\bar{X};C) + \beta \frac{g_r(\bar{X};C)}{t+1},
\end{aligned}
\end{equation}where $\beta > \max_{X \in \Omega} f(X;C)$.

In our implementation, we uniformly feed the network with different $\alpha = \frac{\beta}{t+1}$ for different $t$'s such that the model can be automatically suitable for different $\alpha$'s. Simultaneously, for different $t$, we expect the algorithm $\mathcal{A}_{\theta}$ to adapt such a constraint $t$, so we also use $t$ as an input, i.e., using $\mathcal{A}_{\theta}(\cdot; t)$. During testing, the obtained $\mathcal{A}_{\theta}(\cdot; t)$ outputs $\bar{X}$ that would satisfy different DSP usage thresholds by taking different $t$ as the input. By this, a single model could handle all ranges of DSP usage thresholds.

\textbf{$\mathcal{A}_{\theta}$ based on Gumbel-softmax trick:}
We also use $3$ GraphSAGE layers to encode the configuration $C$ into the latent features. Then, we use MLP layers to reduce the dimension and map to $\mathcal{A}_{\theta}(C)\in [0,1]^{n}$ and the Gumbel-softmax trick to sample $X \sim \text{Ber}(\mathcal{A}_\theta)$. We use the soft Gumbel-softmax~\cite{maddison2016concrete,jang2016categorical} without the straight through trick. The learning rate is set as 1e-3, the batch size is $256$.

\textbf{$\mathcal{A}_{\theta}$ in $\{$relaxation (Na\"{i}ve), relaxation with entry-wise concave proxy (Ours)$\}$:}
The model shares the same structure as that based on the Gumbel-softmax trick, except that the Gumbel-softmax trick is replaced by $\bar{X}\in [0,1]^{n}$ directly. The learning rate is 1e-3, the batch size is $256$.

\subsection{Application III - Circuit Design for Approximate Computing}
\label{apd:application_3-proxy}

\textbf{Dataset details:}
Each configuration $C$ in our approximating computing (A$\times$C) dataset is a computation graph whose nodes represent either multiplication or addition calculation. For each operand, we have two different calculators to carry out the calculation: one is the precise calculator which always output the precise result but requires high computational resource workload, the other is the A$\times$C unit which costs low computational resource but always randomly produces $10\%$ relative error of the actual result. To balance the computation precision and the resource workload, the optimization goal is to minimize the average relative error of the computation graph given the need to use at least a certain number $\theta$ of the A$\times$C units, where $\theta \in \{3, 5, 8\}$. For each instance in the dataset, we randomly take $1,000$ different inputs to calculate the average relative error. Each input consists of $16$ integer numbers that are uniformly sampled from $1$ to $100$. To simulate the locality of some data, some of the inputs only sample $14$ integers and randomly re-use two of them.

\textbf{The C-In, C-Out baselines:}
In the C-In (C-Out) baseline, as many A$\times$C units as the threshold requires are placed randomly near to the input (output) of the approximate computing circuit.

\textbf{The proxy without constraints:}
We utilize $4$ PNA~\cite{corso2020principal} layers as the GNN backbone to show that our method is not limited with certain GNN backbones. The PNA layers take both the configuration $C$ and the optimization variable $\bar{X}$ as inputs with leaky Relu activation and batch normalization. Then the structure is followed by global mean pooling with MLP layers to output $f_r(\bar{X};C)$. Huber loss is used as the criterion, the learning rate is 1e-3, and the batch size is $2048$.

\textbf{The entry-wise affine proxy:}
We also use $4$ PNA layers but only take the configuration $C$ as input to generate the latent node features $H \in \mathbb{R}^{|V| \times d}$ ($d$ is the dimension of the node features). The hyper-parameters are exactly the same as the proxy without constraints. Then $H$ is separated equally into two parts: $U \in \mathbb{R}^{|V| \times \frac{d}{2}}$ and $W \in \mathbb{R}^{|V| \times \frac{d}{2}}$, and we calculate $U \bar{X} + W$, after that we do the log-sum-exponential trick $\sum_{v \in V} \exp [ \log ( U_v \bar{X}_v + W_v) + \sum_{u:(u,v) \in E} \log ( U_u \bar{X}_u + W_u)]$ via message passing to generate the $2$-order moment entry-wise affine latent representation as introduced in Section~\ref{sec:proxy}. Huber loss is used as the criterion, the learning rate is set as 1e-3, the batch size is $2048$.

\textbf{The entry-wise concave proxy:}
The model shares basically the same structure as that in the AFF proxy except for the last layers. We utilize linear layers mixed with a $-\text{Relu}$ function to construct the CON proxy $h_r^c(\bar{X};C) = \langle w^c, -\text{Relu}(\phi(\bar{X};C)) \rangle + b$, as introduced in Section~\ref{sec:proxy}. We use $\text{torch.clamp()}$ function to control the entries in $w^c$ to be always greater or equal to zero in each batch of data processing during the training process.

\textbf{The RL baseline:}
We apply an actor critic model~\cite{konda1999actor}. This model consists of $4$ key components: 1)States, the states are formulated as every possible partially assigned A$\times$C computation graph; 2) Actions, given the current state and the currently considered node of the A$\times$C circuit, the action is the next node to assign with the A$\times$C unit; 3) State transition, given a state and an action, the probability of the next states; 4)Reward, the reward is $0$ for all intermediate actions, in the last action the reward is the evaluation of the fully assigned A$\times$C computation graph.

In each state at a time step, we extract the features from the last layer of the proxy without constraints $f_r$. We utilize another GNN to encode the whole computation graph into a vector encoding. The features are further combined with the vector embedding as the state encoding. Then the state encoding is sent into the policy network that is made up of multiple MLP layers to output the critic value $c$ and the action $a$ which indicates the next node to assign with an A$\times$C unit. The loss for the actor is calculated by subtracting the reward by $c$, and we use Huber loss to make $c$ close to the reward. Note that the state stops if the model has already assigned with as many A$\times$C units as the threshold requires. The reward is defined as the negative proxy prediction:
\begin{equation}
\begin{aligned}
 r_t=\left\{
\begin{aligned}
&- f_r(X;C),& s = T \\
&0, & 0<s<T,
\end{aligned}
\right.
\end{aligned}
\end{equation} where $T$ is the max step, and $s$ denotes the number of the step. The learning rate is set as 1e-2, the discount factor for the reward is set as $0.95$, we train the RL baseline for more than $9,000$ epochs to achieve the reported performance.

\textbf{The mapping of $g_r(\bar{X};C)$:}
The relaxed optimization goal could be written as follows:
\begin{equation}
\begin{aligned}
\min_{\theta} f_r(\bar{X};C), \ \ \text{s.t. } \sum_{i=1}^n \bar{X}_i > t,
\end{aligned}
\end{equation}where $t+1$ is the A$\times$C unit usage threshold, $X_v = 1$ denotes the usage of an A$\times$C unit. The relaxation of the above constraint could be written as $g_r'(\bar{X};C) = n - \sum_{i=1}^n \bar{X}_i \in [0,n-t)$. With the method introduced in Section~\ref{def_CO}, we could normalize it as follows:
\begin{equation}
\begin{aligned}
g_r(\bar{X};C) = \frac{g_r'(\bar{X};C) - g_{\min}}{g_{\min}^+ - g_{\min}},
\end{aligned}
\end{equation}where $g_{\min}^+ = \min_{X\in \{0,1\}^{n} \backslash \Omega} g_r'(X;C) = n - t$ and $g_{\min} = \min_{X\in \{0,1\}^{|V|}} g_r'(X;C) = 0$ in this case. Thus, the normalized constraint could be written as:
\begin{equation}
\begin{aligned}
g_r(\bar{X};C) = \frac{n - \sum_{i=1}^n \bar{X}_i}{n - t}.
\end{aligned}
\end{equation}
The constraint above could satisfy our definition of the CO problems as introduced in Section~\ref{def_CO}, the overall loss function could thus be written as:
\begin{equation}
\begin{aligned}
l_r(\bar{X};C) = f_r(\bar{X};C) + \beta \frac{n - \sum_{i=1}^n \bar{X}_i}{n-t},
\end{aligned}
\end{equation}where $\beta > \max_{X \in \Omega} f(X;C)$.

In our implementation, we uniformly feed the network with different $\alpha = \frac{\beta}{n-t}$ for different $t$'s such that the model can be automatically suitable for different $\alpha$'s. Simultaneously, for different $t$, we expect the algorithm $\mathcal{A}_{\theta}$ to adapt such a constraint $t$, so we also use $t$ as an input, i.e., using $\mathcal{A}_{\theta}(\cdot; t)$. During testing, the obtained $\mathcal{A}_{\theta}(\cdot; t)$ outputs $\bar{X}$ that would satisfy different A$\times$C unit usage thresholds by taking different $t$ as the input. By this, a single model could handle all ranges of A$\times$C unit usage thresholds.

\textbf{$\mathcal{A}_{\theta}$ based on Gumbel-softmax trick:}
We also use $3$ GraphSAGE layers with leaky Relu activation functions and batch normalization to encode the configuration $C$ into the latent features. Then, we use MLP layers to reduce the dimension and map to $\mathcal{A}_{\theta}(C)\in [0,1]^{n}$ and the Gumbel-softmax trick to sample $X \sim \text{Ber}(\mathcal{A}_\theta)$. We use the soft Gumbel-softmax~\cite{maddison2016concrete,jang2016categorical} without the straight through trick. The learning rate is set as 1e-3, the batch size is $2048$.

\textbf{$\mathcal{A}_{\theta}$ in $\{$relaxation (Na\"{i}ve), relaxation with entry-wise concave proxy (Ours)$\}$:}
The model shares the same structure as that based on the Gumbel-softmax trick, except that the Gumbel-softmax trick is replaced by $\bar{X}\in [0,1]^{n}$ directly. The learning rate is 1e-3, the batch size is $2048$.

\section{Broader Impact}
\label{sec:broader_impact}
In this paper, we introduce a general unsupervised framework to resolve LCO problems. The broader impact of this paper is discussed from the following aspects:

\textit{1) Who may benefit from this research.}
The researchers, companies and organizations who utilize our optimization framework to solve CO, LCO or PCO problems might benefit from this research, because our framework reduces the cost of data labeling and improves the performance of the optimization. In addition, more broader people might also benefit from this research, because the unsupervised framework and the standardized low-cost training in comparison with the current methods mean lower energy cost and less pollution, which might do good to the whole society.

\textit{2) Who may be put at a risk from this research.}
Although our method guarantees the quality of the obtained solution when the loss is low, how much gap between the obtained solution and the optimal solution is still unclear. There might be still some gaps to fill in before our method gets deployed in the scenarios where rigorous approximation guarantee of the solutions is requested.  

\textit{3) What are the sequences of the failure of the system.}
A failure of our approach will fail to give a relatively good enough solution to the CO problem.

\section{Licenses}
\label{sec:licenses}
We use the following datasets in our research, their licenses are listed as follows:
\begin{itemize}
    \item The feature based edge covering and node matching problem dataset in application I is generated and proposed by us. It is inspired by~\cite{poganvcic2019differentiation} and utilizes the images from MNIST~\cite{deng2012mnist}, which is under the Creative Commons Attribution-Share Alike 3.0 license. The dataset is publicly available.
    \item The resource binding problem dataset in application II is from~\cite{wu2021ironman} and is publicly available. Please cite their paper in the new publications.
    \item The imprecise functional unit assignment problem dataset in application III is from~\cite{ma2021workload} and is publicly available. Please cite their paper in the new publications.
\end{itemize}

All the datasets and code bases are publicly available. They contain no human information or offensive content.






















\appendix





\end{document}